%% file: manuscript-grusdt-lassiter-franke.tex
\documentclass{sp}

\pdfauthor{Author Full Name(s)}
\pdftitle{Full title}
\pdfkeywords{indicative conditionals, Rational-Speech-Act model, computational pragmatics, causal Bayes nets, inferentialism}

\title[A probabilistic model of communication with conditionals]{Probabilistic modeling of rational communication with conditionals 
 \thanks{We would like to thank three anonymous reviewers for their highly appreciated and very valuable feedback which helped to improve the manuscript substantially.}
}

\author[B. Grusdt, D. Lassiter, M. Franke]{
  \spauthor{Britta Grusdt \\ \institute{Institut f\"ur Kognitionswissenschaften \\
    Universit\"at Osnabr\"uck}} \AND
  \spauthor{Daniel Lassiter \\ \institute{School of Philosophy, Psychology and Language Sciences \\
  The University of Edinburgh}}
  \spauthor{Michael Franke \\ \institute{Allg. Sprachwissenschaft \& Pragmatik \\
  Eberhard Karls Universit\"at T\"ubingen}} \AND
}
\usepackage{enumitem}
\usepackage{multirow}
\usepackage{multicol}
\usepackage{array}
\usepackage{tikz}
\usetikzlibrary{fit,positioning}
\usetikzlibrary{bayesnet}
\tikzset {
 every label/.style = {font=\footnotesize,label distance=0.35cm}
}
\usepackage[titletoc]{appendix}
\usepackage{cleveref}
\usepackage{mathtools}
\newtagform{brackets}{[}{]}
\usetagform{brackets}
\newcommand{\indep}{\rotatebox[origin=c]{90}{$\models$}}
\usepackage{pbox}
\usepackage{nicefrac}
\usepackage{booktabs}
\usepackage{moreenum}
\usepackage{diagbox}
\usepackage{bm}
\usepackage{enumitem}
\usepackage{bibentry}
\definecolor{firebrick}{RGB}{178,34,34}

\newcommand{\den}[1]{\ensuremath{[\![ #1 ]\!]}}

\usepackage{gb4e-emulate}
\noautomath
\newcommand{\pref}[1]{(\ref{#1})}

\newcommand{\ac}[1]{\ensuremath{A\overset{\scriptscriptstyle{#1}}{\rightsquigarrow} C}}
\newcommand{\ca}[1]{\ensuremath{C\overset{\scriptscriptstyle{#1}}{\rightsquigarrow} A}}

\begin{document}

\maketitle

\begin{abstract}
While a large body of work has scrutinized the meaning of conditional sentences, considerably less attention has been paid to formal models of their pragmatic use and interpretation.
Here, we take a probabilistic approach to pragmatic reasoning about indicative conditionals which flexibly integrates gradient beliefs about richly structured world states.
We model listeners' update of their prior beliefs about the causal structure of the world and the joint probabilities of the consequent and antecedent based on assumptions about the speaker's utterance production protocol.
We show that, when supplied with natural contextual assumptions, our model uniformly explains a number of inferences attested in the literature, including epistemic inferences, conditional perfection and the dependency between antecedent and consequent of a conditional.
We argue that this approach also helps explain three puzzles introduced by \citet{Douven2012b} about updating with conditionals: depending on the utterance context, the listener's belief in the antecedent may increase, decrease or remain unchanged.
\end{abstract}


\section{Introduction}

Despite the long history of research on conditionals, no consensus has been reached on crucial questions concerning their semantics and pragmatics, let alone on fundamental questions such as whether conditionals have truth conditions at all.
What is inarguably true is the diversity of how conditionals are interpreted, yet the circumstances that lead to different interpretations of conditionals depending, among others, on the utterance context are not well understood.
An utterance of the conditional in \pref{itm:if-pandemic} might be disillusioning when given by a professional epidemiologist, as one would naturally infer that the speaker is ignorant about the truth of the antecedent and the truth of the consequent.
But how does this inference about the speaker's uncertainty come about? 

\begin{exe}
	\ex\label{itm:if-pandemic} If the pandemic is over next year,  the conference will take place offline again.
\end{exe}

\noindent
The conditional in \pref{itm:if-pandemic} further suggests that the pandemic, mentioned in the antecedent, is related to the form of the conference, mentioned in the consequent.
Does this inferred relationship originate from pragmatic reasoning or do we draw this inference because it is inherently part of the semantics of conditionals?
The latter is advanced by prominent recent accounts of the meaning of conditionals, so-called \emph{inferentialism} \cite[e.g.,][]{Douven2017, Douven2018}.

One long-standing problem that has particularly gained attention in relation with inferentialism is the problem of missing-link conditionals.
Missing-link conditionals are \dash pragmatically infelicitous \dash conditionals with no discernible causal or conditional relationship between antecedent and consequent, as in example \pref{itm:missing-link}.

\begin{exe}
	\ex \label{itm:missing-link} If Jo doesn't pass the exam, it will rain tomorrow.
\end{exe}

\noindent
Contrary to inferentialist accounts, we argue that it is possible to explain the observation that missing-link conditionals are infelicitous solely on the basis of pragmatics, making use of a notion of \emph{listener surprise} about the speaker's utterance choice.

In this paper, we explore a formalization of interlocutors' pragmatic reasoning about the use and interpretation of conditionals within the Rational-Speech-Act (RSA) modeling framework \citep{Frank2012a,Franke2016,Goodman2016}.\footnote{When we write `conditional',  we refer to indicative conditionals whose antecedent and consequent are simple propositions,  to which we restrict the considered kind of conditionals in this paper. }
We aim to contribute to the understanding of how the inferences commonly associated with conditionals arise and how different utterance contexts influence the interpretation of conditionals.
In a nutshell, a probabilistic pragmatics approach enables us to combine two aspects, which we here argue are key to obtaining a more realistic picture of the pragmatics of conditionals: pragmatic reasoning in the vein of \citet{Grice1975} and richly structured, probabilistic world knowledge, particularly about \emph{a priori} plausible causal relationships between antecedent and consequent.
 
To illustrate this, consider example \pref{itm:if-pandemic} again.
We argue that the inferences outlined above can be explained by pragmatic reasoning about the speaker's utterance choice: a cooperative speaker who is in the position to assert the antecedent and/or the consequent alone, should make this stronger claim in order to be as informative as required \citep{quine65,Grice1975,griceIC}.
Given that the speaker did not claim the truth of the antecedent nor the consequent, a listener would plausibly infer uncertainty on the side of the speaker about whether the antecedent and the consequent are true. It is in this sense that we expect Gricean pragmatic reasoning about the speaker's protocol of choosing utterances (e.g., governed by \emph{Maxims of Conversation} or similar) to matter for the interpretation of conditionals.

As to the effects of prior world knowledge, the conditional in \pref{itm:if-pandemic} also illustrates that the listener's prior beliefs may influence the strength of the beliefs the listener holds after receiving information in form of a conditional.
When an assertion of \pref{itm:if-pandemic} is accepted, the listener's degree of belief in the conference taking place offline conditional on the pandemic being over will be high, simply as a result of the conditional's semantic meaning. 
But the degree to which the listener believes in the conference \emph{not} taking place offline conditional on the pandemic \emph{not} being over (arguably one possible measure of a Condtional Perfection reading, see below) crucially depends on the listener's further beliefs about the world. 
The important, though perhaps uncontroversial, point to notice in connection with this example is that prior world knowledge is likely to impact the beliefs the listener holds after accepting and interpreting an utterance of a conditional sentence.

At large,  it  is challenging to account for how people update beliefs in light of new conditional information whereas there is a widely adopted general account of how people update beliefs in light of new factual information, \emph{probabilistic conditioning}.
In this approach, factual information is represented by a probability distribution over a set of possible worlds. 
When a new piece of information $E$ is received, an updated belief state is formed by assuming that the information is true and adjusting the distribution accordingly. 
The consequence is that $P(A)$, the prior belief distribution on proposition $A$, is updated to a posterior distribution $P(A \mid E)$ so that:
$$
P(A \mid E) = \frac{P(A \wedge E)}{P(E)}\,.
$$ 
This idea lies at the heart of Bayesian models in epistemology and decision theory (e.g., \citealt{earman92}) as well as computational cognitive science \citep{tenenbaumetal11}. 
Conditional sentences present a challenge to the generality of the Bayesian approach to learning, because there is no consensus about how the update should work for conditionals, or whether conditioning is even applicable \cite[e.g.,][]{Grove1997, Douven2011a}.
Many authors have argued that conditionals cannot be regarded as having truth conditions at all \citep[for a recent review see][]{Rothschild2015}, but in its standard form, Bayesian update cannot be applied to sentences without truth-conditions.
Even if conditionals do have truth-conditions, it remains unclear how to implement a plausible update operation for conditional information.

In an influential paper, \cite{Douven2012b} argues that there is no uniform way to capture the update effect of conditionals within Bayesian theory, regardless of the assumed semantics. The argument hinges on the relationship between the listener's prior and posterior beliefs about the antecedent. 
Douven shows that \dash depending on the context in which a conditional is uttered \dash the inferred probability of the antecedent may either decrease, increase or remain unchanged compared to the listener's beliefs prior to the uptake of the conditional. Surveying a variety of possible semantics for conditionals, he shows that all of them predict that Bayesian update should have a uniform effect on the probability of the antecedent.
We argue that it is possible to explain the non-uniform update effect of conditionals, that we will refer to as \emph{Douven's puzzle}, by standard Bayesian belief revision if we also take into account pragmatic reasoning and world knowledge. 
Specifically, prior knowledge about how the relevant propositions are likely to relate to each other influence assumptions about what a rational speaker would say under various circumstances. 
Assumptions about a speaker's production protocol in turn influence the listener's interpretation of an utterance of a conditional.
From this perspective, Douven's examples make a strong case for the importance of pragmatic and contextual information in the interpretation of conditionals, and provide a useful test suite for our model which attempts to formalize these factors.\footnote{Like-minded proposals to account for Douven's puzzle have been advanced recently \cite[e.g.,][]{Gunther2018, Eva2019, Vandenburgh2020}, making use of the different intuitive causal structures of Douven's scenarios. However, none of these proposals considers the crucial role of pragmatic reasoning.}

We argue that prior knowledge about the likely causal relationship between antecedent and consequent is not only important to account for Douven's puzzle, but generally mediates a listener's interpretation of a conditional. 
For instance, in \pref{itm:if-pandemic} our background knowledge supports the intuitive inference that there is a causal and probabilistic relationship between the form of a conference and the state of the pandemic.
However, no such relationship seems reasonable in missing-link conditionals like \pref{itm:missing-link}.
But why does background knowledge affect our interpretation in this way, leading to a sense of infelicity when a dependency between consequent and antecedent is implausible?
We argue that assumptions about the causal relationship between antecedent and consequent have an (implicit) influence on the speaker's utterance choice and thereby also on the listener's interpretation, assuming a Gricean pragmatic interpretation of conditionals uttered by a cooperative speaker. 
As we will show,  Gricean pragmatic reasoning \dash formalized in the Rational-Speech-Act model and supplemented by structured representations of causal world knowledge \dash is better able to account for missing-link conditionals than inferentialists have claimed.
Overall, the key to the success of our model stems from leveraging the natural stochastic association between (i) whether antecedent and consequent are causally related and (ii) whether ``If \emph{A}, \emph{C}'' (henceforth $A \rightarrow C$) is the most informative true utterance the speaker can make.

The paper is structured as follows.
\Cref{sec:rsa-conditionals} introduces the vanilla RSA model and details our implementation of the model for reasoning about the use of conditionals.
\Cref{sec:results} and \ref{sec:douvens-puzzle} describe how this model of pragmatic reasoning helps explain key phenomena of interest, on the basis of unspecific as well as contextualized, concrete examples, including Douven's puzzles, conditional perfection readings, and general inferences about the causal relationship between antecedent and consequent.
\Cref{sec:missing-links-bcs} discusses how the presented approach positions itself with respect to missing-link and biscuit conditionals.
\Cref{sec:conclusion} reflects critically and concludes.

\section{A Rational-Speech-Act model for communication with conditionals}
\label{sec:rsa-conditionals}

To provide a principled and uniform formal treatment of pragmatic reasoning which explains the observations raised in the introduction, we turn to probabilistic pragmatic modeling.
The Rational-Speech-Act (RSA) model is a prominent instance of a formalization of Gricean pragmatic reasoning using tools from probability calculus, decision and game theory \citep[see][for an overview]{Franke2016,Goodman2016}.
RSA models are probabilistic, data-driven and based on the assumption that linguistic behavior is goal-oriented and in this regard (approximately) optimal.
A noteworthy benefit of a probabilistic and data-driven approach is that the predictions by the models can often be directly compared to quantitative aspects of experimental data, thereby allowing the statistical comparison of theoretically relevant models based on empirical data \citep[e.g.][]{Qing2015,Degen2020,Franke2020}.
Moreover, as probabilistic modeling is prominent in other areas in the cognitive sciences, it becomes relatively easy to integrate insights from these other areas as well, such as belief formation \citep{Goodman2013,Herbstritt2019}, sequential adaptation \citep{Schuster}, or learning biases \citep{Brochhagen2018}. 
Since it was first introduced by \citet{Frank2012a} in the context of a language reference game, the number of RSA models and modeled phenomena has been growing; scalar implicatures \citep{Goodman2013}, hyperboles \citep{Kao2014a} as well as more complex phenomena such as the interpretation of vague or polite language \citep{lassitergoodman17, Yoon2016}, projective content \citep{Qing2016}, metaphors \citep{Kao2014} or social meaning \citep{Burnett2019} have been investigated by means of RSA.\footnote{A hands-on introduction to RSA modeling, using the probabilistic programming language WebPPL \citep{Goodman2014a}, that we also use to implement our model, is provided by \citet{Scontras2017}.}

There are several reasons why RSA also seems promising for modeling the interpretation of conditionals.
First of all, we expect an influence of the availability of non-conditional utterances that the speaker might have chosen instead of the conditional. In this way,  RSA formalizes a Gricean account of listeners' interpretations and the speakers' utterance choices.
Since RSA is able to flexibly integrate contextual knowledge, we can model quite different utterance contexts, including situations in which richly structured world knowledge is relevant. 
This will be crucial in our account of the effects of causal knowledge on the interpretation of conditionals.

In the following, \Cref{sec:rsa-vanilla} gives a general introduction to the vanilla version of the RSA model.
Subsequent sections then elaborate on the specific adaptations necessary in order to capture reasoning about the use of conditionals against the background of richly structured causal world knowledge. 
We use \Cref{sec:model-example} to demonstrate the main underlying ideas of the model by means of a toy example before we lay out the assumed priors on the generalized set of states in the sections thereafter. 

\subsection{The vanilla Rational-Speech-Act model}
\label{sec:rsa-vanilla}

At the heart of the vanilla RSA model lies the formalization of a cooperative Gricean speaker who, when trying to communicate a state $s$, probabilistically selects an utterance $u$ by preferably choosing utterances that are not only true, but also maximize the amount of relevant information conveyed to a literal listener. The pragmatic listener is modeled as a rational interpreter who combines prior beliefs with the speaker's protocol of choosing utterances by using Bayes' rule.

In order to capture the pragmatic speaker's behavior, in particular, in order to ground out a notion of truth and informativity, the RSA model considers a literal listener first, whose interpretation behavior is defined as a conditional probability distribution $P_{\text{lit}}(s \mid u)$ obtained by updating any prior beliefs $P _{\text{prior}}(s)$ about likely world states $s$ with the set of states $\den{u}$ where utterance $u$ is permissible.\footnote{The formula in Equation \eqref{eq:ll}, and in Equations \eqref{eq:speaker} and \eqref{eq:pl} for other model components, gives the probability up to proportionality ($\propto$), leaving the normalizing constant of the probability distribution implicit.
If $F(x) \ge 0$ is a non-normalized score for any $x \in X$ with $X$ a finite set, the notation $P(x) \propto F(x)$ is shorthand for $P(x) = \frac{F(x)}{\sum_{x'}F(x')}$. Moreover, $\delta_{s \in \sv{u}}$ is the Kronecker delta function, which returns 1 when its Boolean argument \dash the denotation function of utterance $u$ applied to state $s$ \dash is true,  otherwise 0.}
\begin{align}
  & P_{\text{lit}}(s\mid u) \propto \delta_{s \in \sv{u}} \cdot P_{\text{prior}}(s)  \label{eq:ll}
\end{align}
The pragmatic speaker is then defined in terms of a notion of \emph{utterance utility} $U(u;s)$, reflecting how informative $u$ is for communicating $s$.\footnote{A common modification of the basic definition of utterance utility in terms of informativity further accounts for differences among utterances by using utterance cost regarding, for instance, their complexity,  saliency or social compatibility \citep[e.g., see][]{Qing2016, Gates2018}.  We do not include utterance costs in our model, but they could easily be integrated.}
\begin{align}
    & U(u;s) = \text{log } P_{\text{lit}}(s\mid u) \label{eq:utility}
\end{align}
\noindent 
The informativity of an utterance $u$ as description of a state $s$ is defined as the log-likelihood (negative surprisal) of the literal listener's beliefs for state $s$ after hearing utterance $u$.
The probability that a speaker in state $s$ will choose utterance $u$ is then defined as a soft-max operation on utility scores:
\begin{align}
    & P_{\text{S}}(u\mid s) \propto \text{exp}(\alpha \cdot U(u;s)) \label{eq:speaker}
\end{align}
\noindent $\alpha$ is a free model parameter governing how closely the speaker approximates utility maximization: the higher $\alpha$, the more likely the speaker is to choose the utility-maximizing utterance.
At the extremes, a hyperrational speaker ($\alpha \rightarrow \infty$) would only choose utterances that maximize utility, while a randomizing speaker ($\alpha=0$) would choose randomly among true utterances.

Finally, the pragmatic listener's interpretation is captured by a conditional probability distribution, $P_{\text{PL}}(s\mid u)$, which represents the listener's \emph{a posteriori} beliefs (after having heard utterance \emph{u}) about the probability of state $s$, taking the priors over states and a Gricean speaker's utterance-choice behavior into account.
\begin{align}
    & P_{\text{PL}}(s\mid u) \propto P_S(u\mid s) \cdot P_{\text{prior}}(s) \label{eq:pl}
\end{align}

\subsection{World states, utterances \& assertability}
\label{sec:struct-world-stat}

\paragraph{World states.}
Conditionals like $A \rightarrow C$ are often associated with the speaker's uncertainty about whether $A$ and/or $C$ are true.
Therefore, to model pragmatic reasoning about conditionals, we include potentially uncertain speakers into our modeling.
Concretely, we look at the partition of possible worlds into the four types of worlds which agree on the truth values of $A$ and $C$: $W = \{ w_{\emptyset}, w_{A}, w_{C}, w_{AC}\}$. 
The concrete set of states used are probability distributions over worlds $w \in W$.

There are at least two prominent possibilities of how to interpret probability distributions: as precise \emph{objective} chances or as \emph{subjective} beliefs.
Consequently, world states in our model can be interpreted in different ways, too.
For one, we can think of $s \in S$  as the true beliefs of a maximally competent speaker about the objective chance of each type of world.
In this case, the conversational goal, implicitly defined in the vanilla RSA model, is to communicate the true (objective, but intrinsically stochastic) world state known to the speaker.
For another, we can conceptualize world states as a representation of an uncertain speaker's subjective beliefs $s \in S$ about the true (non-probabilistic) state of the world $w \in W$.
Under this interpretation, the vanilla RSA model implicitly treats the conversational goal as that of communicating the speaker's belief state \citep[see][]{Aloni_2007:FC_BiOT,Franke2011:Quantity-Implic}.
The interpretation of world states, as either objective or subjective, may matter for the interpretation of the assertability conditions presented below.
If not stated otherwise, we refer to the subjective interpretation of model states.

\paragraph{Utterance alternatives.}
Predictions of Gricean pragmatic reasoning strongly depend on the assumed set of alternative utterances.
There has been much discussion of alternative sets for scalar items like \emph{some}, \emph{warm} and \emph{or} \citep[e.g.][]{Matsumoto1995,Katzir2007}, but much less for pragmatic reasoning about conditionals \citep[for some discussion see][]{VanDerAuwera1997,VonFintel2001}. 
The selection of alternative utterances  considered in our model is largely governed by the desire to present a balanced set of alternative utterances sufficient to describe the most salient differences in the set of world states.
Utterances are compositionally built from literals, possibly negated.
They may be combined with \emph{and} to form a conjunction, with \emph{if} to form a conditional or with the word \emph{likely}.
Table \ref{tbl:utts-assertability} lays out all alternative utterances together with the rule used to compute the update effects of each \dash its ``assertability condition'', to be introduced next.

\begin{table}
\centering
\begin{small}
\begin{tabular}{llll}
\toprule
\multirow{2}{*}{utterance type}  & \multirow{2}{*}{assertability in state $s$}							&   \multicolumn{2}{l}{example:}	 \\ 
												&																							&	utterance $u$							& assertability $u$ in state $s$\\ \midrule
conjunction 				   				& $P^{(s)}(\phi,\psi) \geqslant \theta$  								   	& $A\wedge \neg C$ 				& $P^{(s)}(A=a,  C=\neg c) \geqslant \theta$  \\ 
literal  										& $P^{(s)}(\phi) \geqslant \theta$ 										   	& $A$										& $P^{(s)}(A=a) \geqslant \theta$ \\ 
\multirow{1}{*}{conditional} 		& \multirow{1}{*}{$P^{(s)}(\psi \mid \phi) \geqslant \theta$}	& $A\rightarrow \neg C$			& $P^{(s)}(C=\neg c\mid A=a) \geqslant \theta$ \\
likely + literal   						    	& $P^{(s)}(\phi) > 0.5$ 											   				& $likely\; \neg C$						& $P^{(s)}(C=\neg c) > 0.5$ \\ \bottomrule
\end{tabular}
\caption{Types of utterances with corresponding assertability conditions and an example,  ordered from most informative utterance on top to least informative at the bottom.  For conditionals and conjunctions, $\phi \neq \psi$. }
\label{tbl:utts-assertability}
\end{small}
\end{table}

\paragraph{Assertability conditions.}
Given a world state $s \in S$ and an utterance $u \in U$,  the definition of the semantics $\den{u} \subseteq S$ serves as the anchoring of pragmatic reasoning in literal interpretation,  defined in Equation~\eqref{eq:ll}.
As especially the semantic meaning of conditionals is highly controversial, we would like to stay as uncommitted and encompassing as possible.
This is possible, to a certain extent, if we focus not on the nature of the denotation function $\den{u}$ but rather at the functional role it plays in the architecture of the pragmatic reasoning model.
In particular, since the utility function in Equation~\eqref{eq:utility} and the speaker rule in Equation~\eqref{eq:speaker} entail that whenever $s \not \in \den{u}$, the speaker will not choose $u$ when in state $s$, the main effect of $\den{u}$ is to give \emph{assertability conditions} and \dash as a side-effect \dash information about how informative each utterance is.\footnote{The informativity of the modeled utterances generally follows the order  shown in  \Cref{tbl:utts-assertability}.  As we were pointed at by an anonymous reviewer,  depending on the chosen prior over states, it is, however,  possible for a conditional (e.g.,  $A\rightarrow C$) to be \emph{literally} more informative than a literal (e.g., \emph{C}) since the assertability of a literal does not per se entail the assertability of a conditional,  e.g.~$P^{(s)}(c) \geqslant \theta \not\Rightarrow P^{(s)}(c\mid a) \geqslant \theta$. This is an interesting observation, in particular in the context of conditionals whose consequent is independent of the antecedent as for instance in concessive conditionals (e.g., ``\emph{Even if \dots}'') which is certainly worth looking at in future work, but beyond the scope of  what we can cover in this paper.}
Consequently, our model lays out a general method of computing update effects of conditionals at the level of a literal interpreter with the ulterior goal of defining reasonable speaker behavior, while avoiding as much as possible concrete commitment to a specific semantic interpretation.

We treat utterances of literals like ``$A$'' as conveying that the target state $s$ makes the probability that $A$ is true (i.e., $A=a$) high enough for conversational purposes; the corresponding probability, e.g., $P^{(s)}(A=a)$ for literal ``A'',  must exceed a certain threshold $\theta$ for the respective utterance to be assertable.\footnote{We write $P^{(s)}$ to refer to probabilities \emph{within} states to distinguish them from probabilities \emph{across} states; $P^{(s)}(X)$ denotes the probability assigned to any event $X$ that may be inferred from state $s$, e.g., $P^{(s)}(A = a)$ is the probability of $A$ to be true in $s$.} 
In a model of objective chance, as we assume on an objective interpretation of the probability distributions building up the set of world states,  determinate truth corresponds to probability 1. 
That is,  a speaker will treat $A$ as true just in case $A$ is determinately true in $s$ \dash when $P^{(s)}(A=a) = 1$.
The assumption that speakers sometimes assert things that are not certain, but very likely true, yet justifies a threshold below 1 as assertability condition.
We also make use of this assumption on a subjective interpretation of world states; a factual sentence $A$ is thus assertable as long as the speaker's subjective belief in $A$ is sufficiently large ($P^{(s)}(A=a) \geqslant \theta$, respectively $P^{(s)}(A=\neg a) \geqslant \theta$ when $u=\neg A$).

Similarly, ``\emph{likely A}'' directly conveys that the subjective probability of $A$ in $s$ is greater than $0.5$ as we assume that ``\emph{likely A}'' is assertable in $s$ iff $P^{(s)}(A=a) > 0.5$.
This aspect of our account is reminiscent of expressivist accounts of probability language \citep{yalcin12a,moss15,swanson15}. 
On the objective interpretation, this means that \emph{likely} expresses high objective chance, consistent with the empirical findings of \citet{fox15} and \citet{lassiter17}.
Note, however, that these authors show that \emph{likely} can also express subjective uncertainty. 

In parallel fashion, we render the assertability conditions of an indicative conditional $A \rightarrow C$ as $P^{(s)}(C=c \mid A=a) \geqslant \theta$, corresponding either to a high objective chance of $C$ to be true when $A$ is true or to a strong belief in this conditional probability (on part of the speaker), assuming an objective and subjective interpretation respectively.
Since we do not want or need to commit to a specific semantic theory of conditionals here, we settle for motivating this condition as a plausible minimal bound on assertability that should be acceptable from a wide range of theoretical perspectives. For theories that are able to support the equation $P(A \rightarrow C) = P(C \mid A)$ while avoiding triviality results (e.g., \citealt{lewis76,hajek89}), the derivation is strictly parallel to the factual case above. This includes
non-propositional theories \citep{edgington95,bennett03}, 
trivalent theories \citep{definetti36,milne97,lassiter19}
and various others (\citealt{vanfraassen76,stalnakerjeffrey94,kaufmann04,khoo16b}, among others). 
Our assertability condition is also particularly natural for theories that render the truth-condition of $A\rightarrow C$ as $P(C\mid A) = 1$ \citep[e.g.,][]{moss15}.

The status of our assertability condition for conditionals is somewhat murkier from the perspective of other prominent theories such as \citet{stalnaker68} and \citet{kratzer91}, as well as strict conditional theories. Because of the complexity of the way that they assign truth-values to epistemically possible worlds where the antecedent $A$ is false, these theories can make $P(A \rightarrow C)$ high even while $P(C \mid A)$ is low. As a result, our assertability condition is stronger than these accounts would predict. However, we note that there is by now an enormous body of empirical evidence supporting the equation between the probability of a conditional and the corresponding conditional probability (\citealt{hadj,evansover04,Douven2010}, among many others). This evidence problematizes a key prediction made by the latter group of theories: that a conditional can be judged highly probable simply because of the likely falsehood of its antecedent. Instead, situations where the antecedent is false are generally judged irrelevant to the probability of a conditional, in a probabilistic analogue of the classic paradoxes of the material conditional \citep{edgington95}. We do not doubt that the theories under consideration have theoretical resources available that may allow them to avoid this problem---for example, by using pragmatic reasoning to explain why false-antecedent cases are not considered relevant in assertion \citep{lewis76,griceIC}. But doing so would be tantamount to adopting our assertability condition or something quite close to it. Therefore, we believe that our results should be relevant to theorists with a wide variety of semantic commitments, including those for whom probabilistic reasoning has not previously played a major theoretical role.

\subsection{Toy example}
\label{sec:model-example}

Let us consider a toy example to illustrate model predictions, and to motivate further generalizations to be introduced hereafter.
Concretely, we will consider a case with just three world states, all equally likely and just constructed for the sake of illustration.
The topic of conversation in this example is whether Alex and Chris are likely to go to a party.
The states we look at differ in the probability they assign to all four logical possibilities of Alex and/or Chris going to the party.
\begin{enumerate}[label=($s_\arabic*$)]
\item Alex and Chris both really like to go out.  Both are seen at most parties, but whether either comes is unrelated to whether the other comes.
\item Alex and Chris go slightly more often than not, but usually not without each other (e.g., they are a couple, best friends, etc.).
\item Alex and Chris each go out more often than not, but have no connection with each other.
\end{enumerate}

\noindent
These three scenarios are translated into the probability distributions $P(A,C)$ shown at the top of   \Cref{tbl:states-toy-example} where  $A$ ($C$) denotes whether Alex (Chris) comes to the party.

With this context in mind, imagine the following utterance Bigi directs at Wobo:

\begin{exe}
\ex \label{bsp:out-of-the-blue-C}
	Bigi: \emph{Chris will come to the party}. \hfill \textcolor{gray}{[``\emph{C}.'']}
\end{exe}

\noindent Given that Bigi only says what she believes to be true and $s_1, s_2$ and $s_3$ describe all possible scenarios, Wobo should infer that Bigi describes the situation modeled in $s_1$ since this utterance (\emph{C}) is neither assertable in $s_2$ nor $s_3$.

Now, imagine Bigi to utter the following conditional instead:
\begin{exe}
\ex \label{bsp:out-of-the-blue-ifac}
  Bigi: If Alex comes to the party, Chris comes too. \hfill \textcolor{gray}{[``If $A$, $C$.'']} \\
\end{exe}

\begin{table}
\caption{Model predictions for the scenario from \Cref{sec:model-example} where $\alpha=1, \theta=0.9$. (a) States $s_1, s_2, s_3$. (b) Assertability of utterances given states with respectively most informative utterances in bold. (c) Literal interpretation. (d) Speaker production likelihoods. (e) Pragmatic interpretation.}
\label{tbl:states-toy-example}
\begin{center}
\begin{footnotesize}
\begin{tabular}{lllccllccclcc}\toprule
& 												&	$s_1$		& $c$ 	& $\neg c$ 	&& $s_2$		& $c$ 	& $\neg c$ 	&& $s_3$		& $c$ 	& $\neg c$ \\ \midrule
\multirow{2}{0.25cm}{(a)} & \multirow{2}{2cm}{$P(A,C)$} 	& $a$			& 0.81 	& 0.09 			&& $a$ 		& 0.6 	& 0.05 			&& $a$ 		& 0.36	& 0.24 \\
&												& $\neg a$ 	& 0.09 	& 0.01 	  		&& $\neg a$ & 0.05 	& 0.3 			&& $\neg a$ 	& 0.24 	& 0.16  \\ \midrule
\multirow{4}{0.25cm}{(b)} & $u=\text{likely}\;  C$ 					& \multicolumn{3}{c}{1}  				& 			& \multicolumn{3}{c}{1} 			&& \multicolumn{3}{c}{$\bm{1}$}\\
& $u=A\rightarrow C$ 					& \multicolumn{3}{c}{1}					& 			& \multicolumn{3}{c}{$\bm{1}$}	&& \multicolumn{3}{c}{0}\\
& $u=C$										& \multicolumn{3}{c}{$\bm{1}$} 	& 			& \multicolumn{3}{c}{0} 			&&\multicolumn{3}{c}{0}\\
& $u=A \wedge C$ 						& \multicolumn{3}{c}{0} 				& 			& \multicolumn{3}{c}{0} 			&&\multicolumn{3}{c}{0}\\ \midrule
\multirow{4}{0.25cm}{(c)} & \multicolumn{1}{l}{$P_{\text{lit}}(s_i\mid u=likely\; C)$} & \multicolumn{3}{c}{$\nicefrac{1}{3}$}		&&  \multicolumn{3}{c}{$\nicefrac{1}{3}$} 	&& \multicolumn{3}{c}{$\nicefrac{1}{3}$} \\ 
&																				   \multicolumn{1}{l}{$P_{\text{lit}}(s_i\mid u=A\rightarrow C)$} 			& \multicolumn{3}{c}{$\nicefrac{1}{2}$}		&&  \multicolumn{3}{c}{$\nicefrac{1}{2}$} 	&& \multicolumn{3}{c}{0} \\ 
&																			   \multicolumn{1}{l}{$P_{\text{lit}}(s_i\mid u=C)$} 								& \multicolumn{3}{c}{1}								&&  \multicolumn{3}{c}{0} 						&& \multicolumn{3}{c}{0} \\ 
&																				   \multicolumn{1}{l}{$\sum_{u\prime} P_{\text{lit}}(s_i\mid u\prime)$} & \multicolumn{3}{c}{$\nicefrac{11}{6}$}	&&  \multicolumn{3}{c}{$\nicefrac{5}{6}$} && \multicolumn{3}{c}{$\nicefrac{1}{3}$} \\ \midrule

\multirow{3}{0.25cm}{(d)} & \multicolumn{1}{l}{$P_{\text{S}}(u=likely \; C\mid s_i)$} 				& \multicolumn{3}{c}{$\nicefrac{\frac{1}{3}}{\frac{11}{6}}=\nicefrac{2}{11}$}	&&  \multicolumn{3}{c}{$\nicefrac{\frac{1}{3}}{\frac{5}{6}}=\nicefrac{2}{5}$}  && \multicolumn{3}{c}{$\nicefrac{\frac{1}{3}}{\frac{1}{3}}=1$} \\ 
& \multicolumn{1}{l}{$P_{\text{S}}(u=A\rightarrow C\mid s_i)$}		& \multicolumn{3}{c}{$\nicefrac{\frac{1}{2}}{\frac{11}{6}}=\nicefrac{3}{11}$}	&&  \multicolumn{3}{c}{$\nicefrac{\frac{1}{2}}{\frac{5}{6}}=\nicefrac{3}{5}$}  && \multicolumn{3}{c}{0} \\ 
& \multicolumn{1}{l}{$P_{\text{S}}(u=C\mid s_i)$} 					 		& \multicolumn{3}{c}{$\nicefrac{1}{\frac{11}{6}}=\nicefrac{6}{11}$}					&&  \multicolumn{3}{c}{0}  																			&& \multicolumn{3}{c}{0} \\  \midrule

\multirow{3}{0.25cm}{(e)} & \multicolumn{1}{l}{$P_{\text{PL}}(s_i\mid u=likely \; C)$} 			& \multicolumn{3}{c}{$\nicefrac{\frac{2}{11}}{(\frac{2}{11} + \frac{2}{5} + 1)}=\nicefrac{10}{87}$}	&&  \multicolumn{3}{c}{$\nicefrac{\frac{2}{5}}{(\frac{2}{11} + \frac{2}{5} + 1)} = \nicefrac{22}{87}$} 	&& \multicolumn{3}{c}{$\nicefrac{1}{(\frac{2}{11} + \frac{2}{5} + 1)}=\nicefrac{55}{87}$}\\ 
& \multicolumn{1}{l}{$P_{\text{PL}}(s_i\mid u=A\rightarrow C)$} 	& \multicolumn{3}{c}{$\nicefrac{\frac{3}{11}}{(\frac{3}{11} + \frac{3}{5})}=\nicefrac{5}{16}$}	&&  \multicolumn{3}{c}{$\nicefrac{\frac{3}{5}}{(\frac{3}{11} + \frac{3}{5})}=\nicefrac{11}{16}$} 	&& \multicolumn{3}{c}{0}\\ 
& \multicolumn{1}{l}{$P_{\text{PL}}(s_i\mid u=C)$} 						& \multicolumn{3}{c}{$\nicefrac{\frac{6}{11}}{\frac{6}{11}}=1$}						&&  \multicolumn{3}{c}{0} 																			&& \multicolumn{3}{c}{0}\\ \bottomrule
\end{tabular}
\end{footnotesize}
\end{center}
\end{table}

\noindent
As noted in \Cref{tbl:states-toy-example}, $A\rightarrow C$ is assertable in states $s_1$ and $s_2$.
Therefore, under a literal interpretation of Bigi's utterance,  $s_1$ and $s_2$ are considered equally likely ($P_{\text{lit}}(s_1\mid u=A\rightarrow C) = P_{\text{lit}}(s_2\mid u=A\rightarrow C) = 0.5$).
However,  under a pragmatic interpretation of $A\rightarrow C$, $s_2$ is judged as more likely than $s_1$ ($\nicefrac{11}{16}$ vs.  $\nicefrac{5}{16}$).
This is because Bigi could have chosen a more informative utterance to communicate $s_1$.

\subsection{Inferring latent causal relations}
\label{sec:inferring-causal-relations}

The example in Table~\ref{tbl:states-toy-example} shows how the model introduced so far yields the inference that $A \rightarrow C$ is most likely associated with $s_{2}$, by rather straight-forward Gricean reasoning.
It also demonstrates how, on top of inferring a state $s$, Wobo might draw inferences about the likely causal relation between $A$ and $C$ which may have led to Bigi's beliefs as captured in $s_{1}$, $s_{2}$ or $s_{3}$.
By motivation of the example, the coming of Alex and Chris was assumed to be \emph{independent} in states $s_{1}$ and $s_{3}$, while the very reason for writing a table like in $s_{2}$ was because we assumed that there was a stochastic relationship between Alex's and Chris' coming to the party.
Suppose that there are two equally likely relations $r \in \{\text{independent}, \text{dependent}\}$, meaning that $A$ and $C$ are either independent or dependent.
Intuitively,  $P(S = s_{2} \mid r = \text{independent}) $ is much smaller than $P(S = s_{1} \mid r = \text{independent})$ or $P(S = s_{3} \mid r = \text{independent})$, and also $P(S = s_{2} \mid r = \text{dependent}) $ is much higher than $P(S = s_{1} \mid r = \text{dependent})$ or $P(S = s_{3} \mid r = \text{dependent})$.
In this way, by Bayesian inference, we can obtain an indirect inference of a likely causal/stochastic relation between $A$ and $C$ just from probabilistic pragmatic reasoning \emph{and} natural assumptions about the differential likelihood between different causal/stochastic relations $r$ and states $s$.
Note the importance of the pragmatic reasoning for drawing an inference about the likely relation in this example: under a literal interpretation of the conditional $A\rightarrow C$,  Wobo would not show a preference between $s_1$, where $A$ and $C$ are likely dependent, and $s_2$, where they are likely independent.

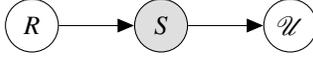
\begin{figure}
\centering
\begin{tikzpicture}
\node[latent] (r) [] {$R$};
\node[obs] (s) [right=of r] {$S$};
\node[latent] (u) [right=of s] {$\mathcal{U}$};
\edge {r} {s};
\edge {s} {u};
\end{tikzpicture}
\caption{Through world knowledge, beliefs about the causal relation $r \in R$ inform probabilities $s\in S$ which are, in turn, considered as known by the speaker. 
Probabilities in $s$ then directly influence the speaker's utterance choice $u\in \mathcal{U}$.}
\label{fig:schematic-inferences}
\end{figure}

We can think of this model as a sequence of inferences: beliefs about $r$ stochastically inform $s$ via world knowledge or intuitions about  dependence/independence and $s$, in turn, stochastically informs the speaker's utterance choice $u$ due to pragmatic constraints on what counts as a good utterance.
Schematically: $P(r) \Rightarrow P(s \mid r) \Rightarrow P(u \mid s)$ which is depicted in \Cref{fig:schematic-inferences}.
We can then derive $P(r \mid u)$ via Bayes' rule.
In the following we specify and motivate a concrete prior structure, in particular for the $P(s \mid r)$ part, so as to be able to derive general predictions from this model for what we may consider a \emph{default context}, where no specific world knowledge is assumed to be available regarding antecedent and consequent.

\subsection{Prior over world states in default context}
\label{sec:prior-states}

\begin{table}
\centering
\begin{tabular}{p{1.75cm}p{2.5cm}l}  \toprule
causal relation ($R$)		& instance causal relation ($r$) & interpretation \\ \midrule
\multirow{2}{*}{$A\rightsquigarrow C$} 	& \ac{++}									& Truth of $A$ increases probability for truth of $C$\\
													& \ac{-+} 									& Falsity of $A$ increases probability for truth of $C$\\ \midrule
\multirow{2}{*}{$C\rightsquigarrow A$} 	& \ca{++} 									& Truth of $C$ increases probability for truth of $A$\\
													& \ca{-+} 									& Falsity of $C$ increases  probability for truth of $A$\\ \bottomrule
\end{tabular}
\caption{Notation for dependent causal relations (types and instances).  The instance of the causal relation provides information about the associated joint probability tables, spelled out in column `interpretation'. }
\label{tbl:notation-dep-tables}
\end{table}

A state $s$ where variables $A$ and $C$ are assumed to be independent ($r=A\indep C$) is represented by probabilistically independent distributions where $P^{(s)}(A,C)=P^{(s)}(A)\cdot P^{(s)}(C)$.
Since in the default context, we do not make any specific assumptions, $A$ and $C$ are both assigned a uniform prior probability over the interval $[0, 1]$:
\begin{equation}
P^{(s)}(A=a), \; P^{(s)}(C=c) \sim \text{Uniform}(0,  1)
\end{equation}
\noindent
Together with the assumption of independence, this yields the following probability distribution over partitions of possible worlds, representing a single state $s$ when $r=A\indep C$:
\begin{align*}
& P(w_{AC}) = P^{(s)}(A=a)\cdot P^{(s)}(C=c), & & P(w_{A}) = P^{(s)}(A=a) - P(w_{AC})  \\
& P(w_{C}) = P^{(s)}(C=c) - P(w_{AC}), & & P(w_{\emptyset}) = 1 - (P(w_{AC}) + P(w_{A}) + P(w_{C}))
\end{align*}

To derive the probability distributions over partitions of possible worlds that represent states where variables $A$ and $C$ are assumed to be dependent,  we distinguish between two possible \emph{types} of causal relation: either $A$ has causal power to provoke $C$ ($R=A\rightsquigarrow C$) or vice versa ($R=C \rightsquigarrow A$).\footnote{
The choice to model only a single relevant cause is corroborated by many empirical studies that have shown people's tendency to neglect alternative causes \citep[e.g., see][]{Fernbach2010, Fernbach2011, Fernbach2013, Hagmayer, Krynski2007}; also,  in a theoretical model from \citet{Icard2015}, the loss of information resulting from neglecting alternative causes was on average too small as to justify their consideration.
}
While the causal relation, $R$, merely provides information about the causal direction, the concrete \emph{instances} of a causal relation, denoted as $r$, are distinguished based on how exactly they affect the conditional probabilities, as shown in  \Cref{tbl:notation-dep-tables}.
$R=A\rightsquigarrow C$, for instance, tells us that the outcome of variable $A$ has direct influence on the outcome of variable $C$, and the instance of the causal relation, $r$, further tells us \emph{how} the outcome of $A$ influences the outcome of $C$.
For example, we write $r=$\ac{++} to pick out the class of probability distributions where the truth of $A$ increases the probability that $C$ is true, whereas $r=$\ac{-+} picks out probability distributions where the \emph{falsity} of $A$ increases the probability that $C$ is true. 
Putting this differently,  when $r=$\ac{++},  $w_{AC}$ will be considerably more likely than $w_{A}$ ($P(w_{AC} \gg w_A$), and when $r=$\ac{-+},  $P(w_C) \gg P(w_{\emptyset})$.

\begin{figure}
\begin{minipage}{0.3\textwidth}
\begin{small}
\begin{tikzpicture}
\node (C) {$C$};
\node (A) [above left=of C] {$A$};
\node (B) [above right=of C,label={}] {$B$};
\draw[->] (A) -- (C) node[midway,left] {$\tau$};
\draw[->] (B) -- (C) node[midway,right] {$\beta$};
\end{tikzpicture}
\end{small}
\end{minipage}
\begin{minipage}{0.7\textwidth}
\begin{align*}
& P(C=c \mid A=a) = 1 -  (1-\tau) \cdot (1-\beta) = \tau + \beta - \tau \cdot \beta \\
& P(C=c \mid A=\neg a) = \beta
\end{align*}
\end{minipage}
\caption{Graphical representation of a leaky noisy-or model (left) with a single explicitly modeled cause $A$ and the corresponding conditional probabilities of $C$ when $A$ is true or false (right). 
Variable $B$ summarizes all potential other causes of $C$; $\tau$ and $\beta$ denote the causal power of $A$, respectively $B$, to induce the truth of $C$.}
\label{fig:noisy-or}
\end{figure}
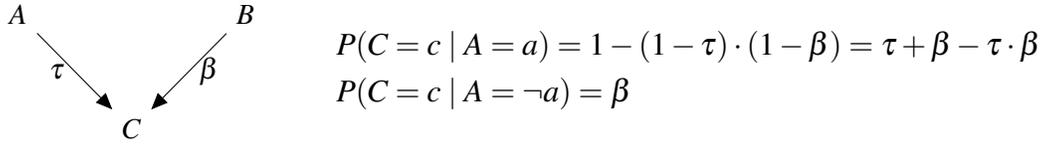

Formally,  the probability distributions of the dependent states can be described as leaky-noisy or model with binary variables allowing for positive as well as negative causes. 
Noisy-or models \citep{pearl88}  describe the relationship between an effect variable and its cause variables where each cause is capable of producing the effect independently of all other causes; corresponding to a logical OR-function where the effect is only true if at least one of its causes is true.  
Leaky noisy-or models \citep{diezParameterAdjustmentBayes1993a} comprise ``background noise'' represented by an additional cause variable that is always present and summarizes all potential causes of the effect that are not explicitly modeled.
Figure \ref{fig:noisy-or} shows a graphical representation of the leaky noisy-or model underlying the dependent states in  our model with the corresponding conditional probabilities of the effect $C$ to be true when $A$ is true or false respectively,  assuming non-deterministic relations where variables $A$ and $B$ have causal power $\tau$, respectively $\beta$, to provoke $C$.
Since we want to stick to the simplest set of induced probability distributions,  but still need to cover all possible stochastic relations between the two explicitly modeled variables $A$ and $C$ to have a balanced set of states that preserves the informativity of utterances, we use a generalization of the classical noisy-or model which further allows negative causes: not only the truth, but also the falsity, of a cause can have an influence on the truth of the effect \citep[e.g., see][]{hyttinenNoisyORModelsLatent2011}.\footnote{If we only used positive causes, states where $P(w_{AC})$ and $P(w_{\emptyset})$ tend to be low would be very rare and thus, conditionals like $A\rightarrow \neg C$ would only be assertable in very few states which would, in turn, render this utterance very informative.  As a result, this particular conditional would become more informative than for instance literals, which does not seem reasonable.  Using states with positive and negative causes results in a balanced set of states such that any conjunction is more informative than any literal and any literal is, in turn,  more informative than any conditional. }
Therefore,  $C$ is likely true when $A$ is false and a negative cause ($r=$\ac{-+}) or when $A$ is true and a positive cause ($r=$\ac{++}); in both cases the truth of $C$ may independently be due to background noise $B$.

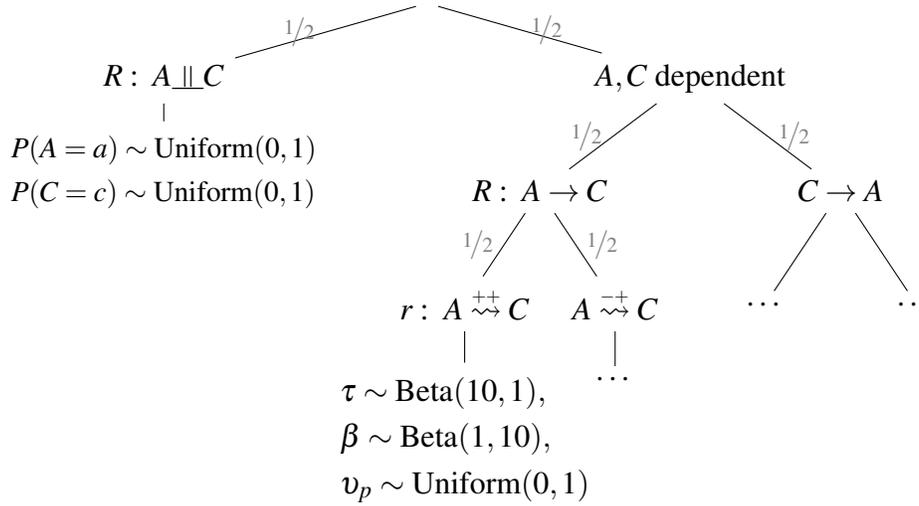
\begin{figure}
\centering
\begin{tikzpicture}
\node {} [sibling distance = 7cm, level distance = 1cm]
    child [] {node {$R:\; A\indep C$} 
    		child [level distance = 1.25cm] { node {
    		\small{$\begin{aligned}
    		& P(A=a)\sim  \text{Uniform}(0,1)\\
    		& P(C=c) \sim \text{Uniform}(0,1)
    		\end{aligned}$
    		}}		 			 			
 		edge from parent [] node [left, gray] {}
 		} 	    
    edge from parent [] node [left, gray] {\small{$\nicefrac{1}{2}$}}
    } 
    child {node (dep) [] {$A,C$ dependent}
    		child [level distance = 1.5cm, sibling distance = 4cm] {node {$R:\;A\rightarrow C$}
 			child [level distance = 1.5cm, sibling distance = 2cm] {node {$r:\;$ \ac{++}} 
 				child [level distance = 1.8cm] { node {
					$\begin{aligned} 
						& \tau \sim \text{Beta}(10,1), \\
 						& \beta \sim \text{Beta}(1,10), \\
 						& \upsilon_p \sim \text{Uniform}(0,1)
					\end{aligned}$ 	 				
 				}		 			 			
 			 	edge from parent [] node [left, gray] {}
 			 	}
 			edge from parent [] node [left, gray] {\small{$\nicefrac{1}{2}$}}
 			} 			    
 			child [level distance = 1.5cm, sibling distance = 2cm] {node {\ac{-+}} 
 				child [level distance = 1cm] { node {\dots}}
 			edge from parent [] node [right, gray] {\small{$\nicefrac{1}{2}$}}}
 			edge from parent [] node [left, gray] {\small{$\nicefrac{1}{2}$}}    				
    		}
    		child  [level distance = 1.5cm, sibling distance = 4cm] {node  {$C\rightarrow A$}
 			child [level distance = 1.5cm, sibling distance = 2cm] {node {\dots} [sibling distance = 0.5cm] edge from parent [] node [left] {}}    
 			child [level distance = 1.5cm, sibling distance = 2cm] {node {\dots} [sibling distance = 0.5cm] edge from parent [] node [right] {}}
 			edge from parent [] node [right, gray] {\small{$\nicefrac{1}{2}$}}   		
    		}
    		edge from parent [] node [right, gray] {\small{$\nicefrac{1}{2}$}}
    	};
\end{tikzpicture}
\caption{Graphical representation of the procedure for sampling a state $s$ from the prior in the default context.}
\label{fig:sampling-procedure}
\end{figure}

To instantiate the set of probability distributions for the respective dependent causal relations, we specify the prior distributions over the respective causal power ($\tau$), noise ($\beta$) and prior probability of the parent variable ($\upsilon_p$) as shown in \Cref{fig:sampling-procedure}.\footnote{
Note that even though the notion of \emph{causal} relations is eventually not important as the modeling  hinges on certain probabilistic dependencies between events (depending on the relation),  it is important for the choice of our prior distributions which is motivated by relations that are causal by nature. 
}
The values of the hyperparameters for the beta distributions of the causal power and noise are chosen such that the mean of the distribution of the causal power $\tau$ exceeds the assertability threshold, set to 0.9 in all simulations below; for the causal power of the background noise, the parameters are simply reversed such that the prior distribution $P(\beta)$ is skewed towards 0.
As we consider a default context here,  where no further information is available, the prior probability of the parent variable ($\upsilon_p$) is sampled from a uniform distribution over the interval $[0,1]$. 
The joint probability distributions over $A,C$ are then derived based on $\tau, \beta$ and $\upsilon_p$.  
Depending on the relation $R$ and the cause variable being a positive or a negative cause, $\beta$,  $\upsilon_p$ and $\upsilon_c$ correspond to different (conditional) probabilities, as listed in  \Cref{tbl:probs-dep-tables} where $\upsilon_c$ denotes the conditional probability of the effect to be true when the cause variable is true: $\upsilon_c = \tau + \beta - \tau \cdot \beta$ (see  \Cref{fig:noisy-or}).

The joint probability distributions over partitions of possible worlds are then computed as follows when $R=A\rightsquigarrow C$ (similarly for $R = C\rightsquigarrow A$):
\begin{align*}
& P(w_{AC}) = P(c\mid a)\cdot P(a), & & P(w_{A}) = P(\neg c\mid a)\cdot (P(a) \\
& P(w_{C}) = P(c\mid \neg a) \cdot P(\neg a), & & P(w_{\emptyset}) = P(\neg c\mid \neg a) \cdot P(\neg a)
\end{align*}

\begin{table}
\centering
\begin{tabular}{p{2.5cm}lll} \toprule
instance causal relation ($r$)														& \multicolumn{1}{c}{$\upsilon_p$} 	& \multicolumn{1}{c}{$\upsilon_c$} & \multicolumn{1}{c}{$\beta$} \\\midrule
\ac{++}	& $P^{(s)}(A=a)$ 															& $P^{(s)}(C=c\mid A=a)$ 					& $P^{(s)}(C=c \mid A=\neg a)$\\
\ac{-+}	& $P^{(s)}(A=\neg a)$ 													& $P^{(s)}(C=c \mid A=\neg a)$ 			& $P^{(s)}(C=c\mid A=a)$\\ \midrule

\ca{++}	& $P^{(s)}(C=c)$ 															& $P^{(s)}(A=a\mid C=c)$ 					& $P^{(s)}(A=a \mid C=\neg c)$\\
\ca{-+}	& $P^{(s)}(C=\neg c)$ 													& $P^{(s)}(A=a\mid C=\neg c)$ 			& $P^{(s)}(A=a \mid C=c)$\\ \bottomrule
\end{tabular}
\caption{Probabilities ($\upsilon_p, \upsilon_c, \beta$) that define the joint probability distribution of a state $s$, $P^{(s)}(A,C)$, for each instance of a dependent causal relation.  $\upsilon_p$ is the prior probability of the cause, $\upsilon_c$ is the conditional probability of the effect to be true when the cause is true  and $\beta$ is the power of the unmodeled variables to provoke the effect corresponding to the conditional probability of the effect to be true when the explicitly modeled cause is false.}
\label{tbl:probs-dep-tables}
\end{table}

\begin{figure}
\centering
\includegraphics[width=0.31\textwidth]{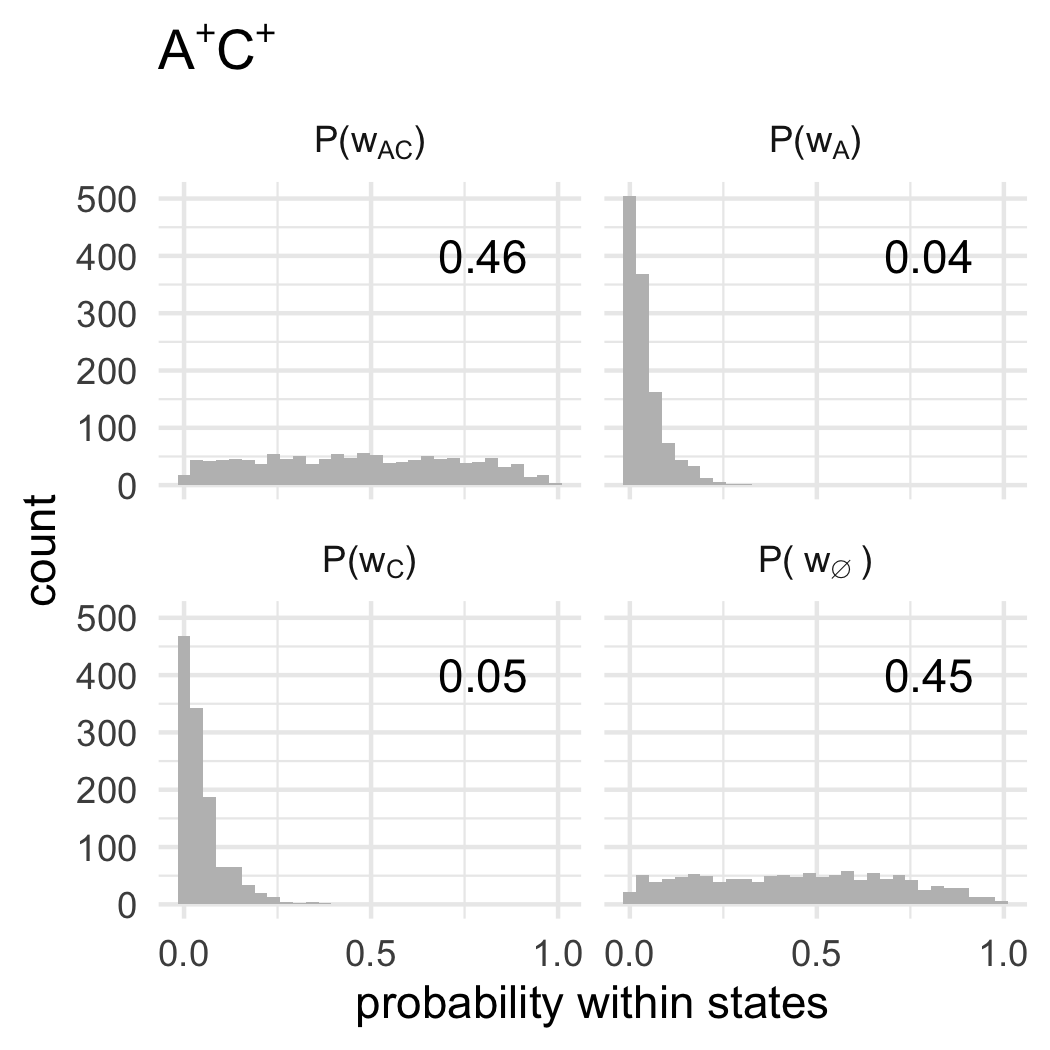}\quad
\includegraphics[width=0.31\textwidth]{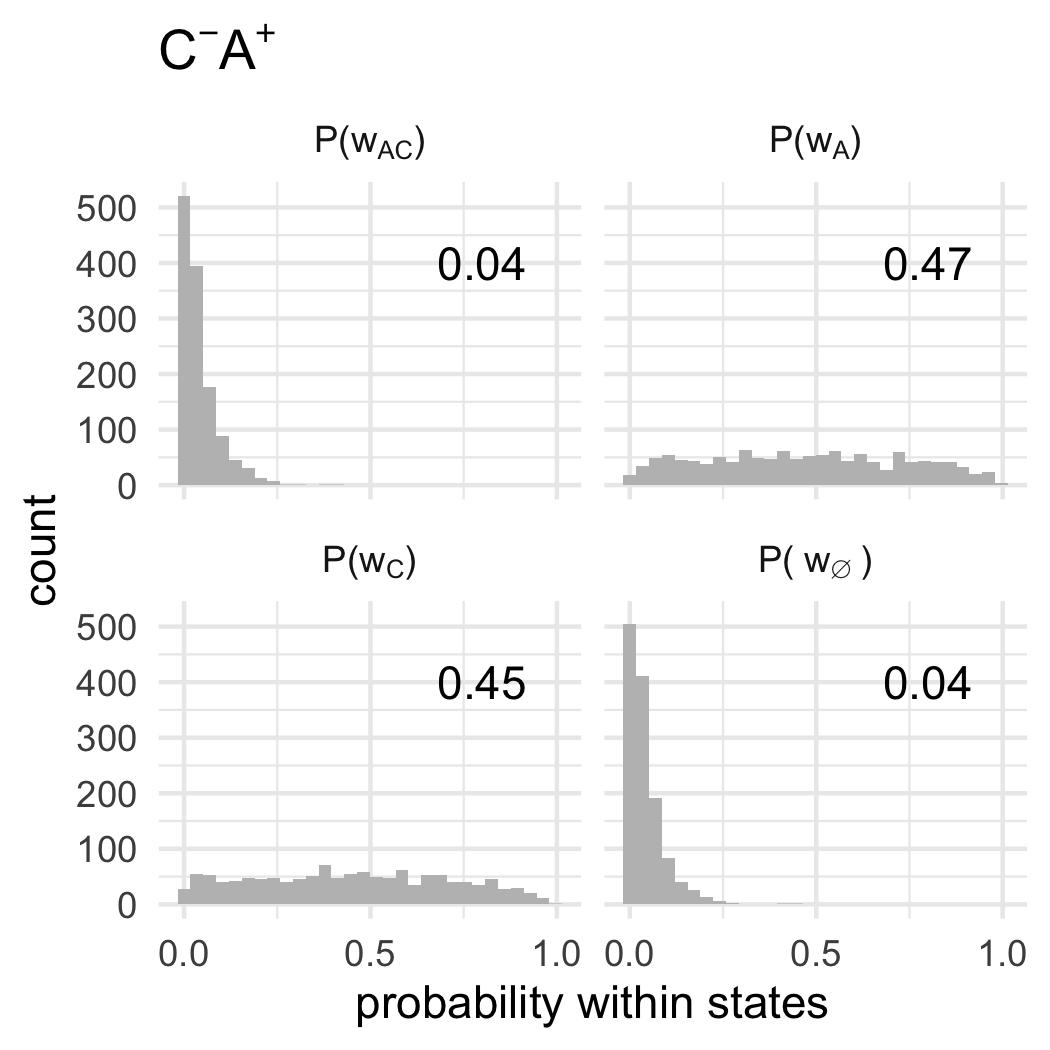}\quad
\includegraphics[width=0.31\textwidth]{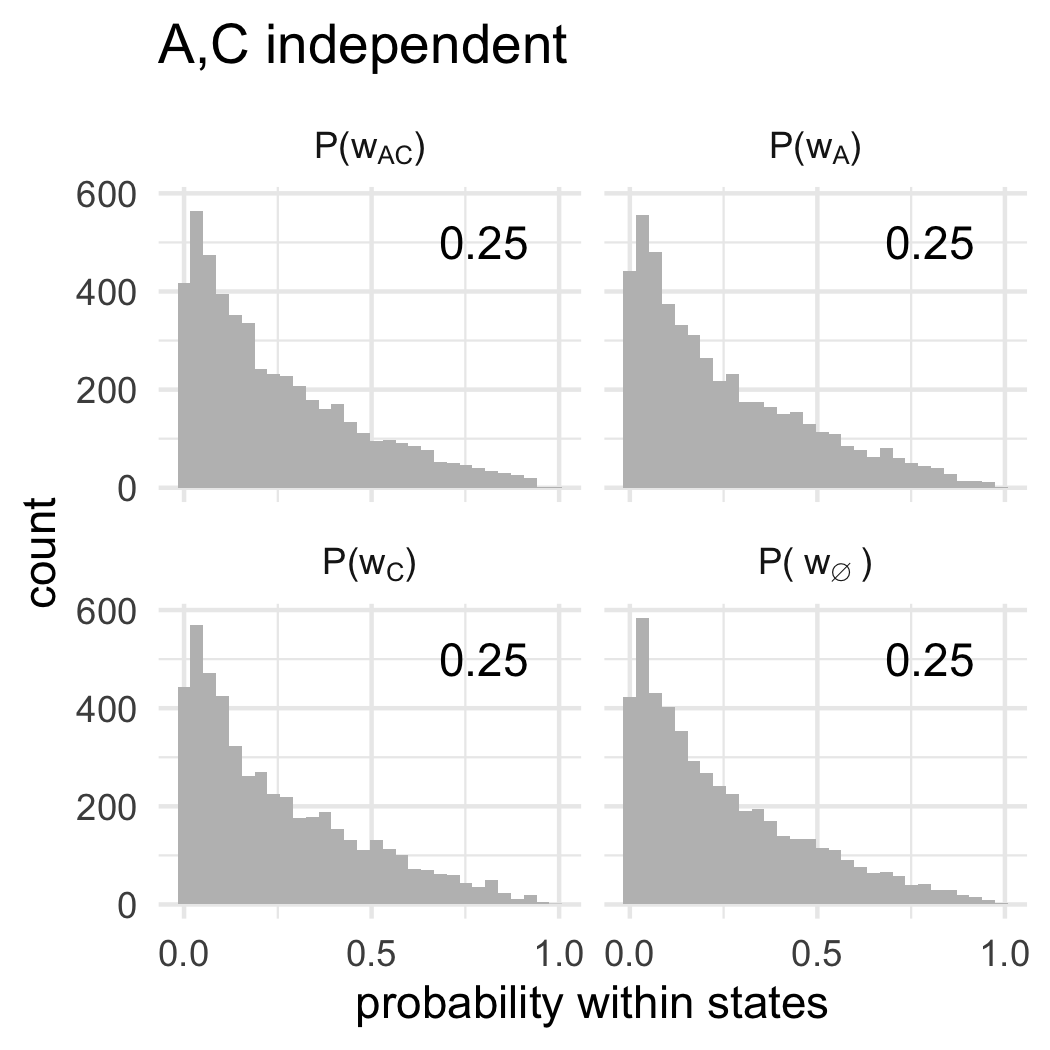}
\caption{Histograms of the probabilities of the four possible worlds, $w_{AC}, w_{A}, w_{C}$ and $w_{\emptyset}$ of all sampled probability tables $P^{(s)}(A,C)$ with $r=$ \ac{++} (left),  $r=$ \ca{-+} (middle) and $r=A \indep C$ (right).  Numbers in the upper right corners are the expected values for the respective worlds.
}
\label{fig:table-distributions}
\end{figure}

In total, we sample 10,000 probability distributions that comprise the set of world states used by our model.
More concretely, we first sample a causal relation $r$ from its prior distribution given in Equation~\eqref{eq:prior-cn}.
Our choice to put a prior on the causal relation,  which determines the shape of the associated sampled probability distributions, is primarily based on work from  \citet{tenenbaum_theory-based_2003, griffiths_structure_2005}; they were the first who put priors on the causal structure itself to predict human causal judgments instead of focusing on learning the causal strength of a,  possibly non-existent, causal link \citep[e.g., see][]{Cheng1997}.

\begin{equation}\label{eq:prior-cn}
P(r) = \begin{cases} \nicefrac{1}{2} & \text{if } r=A\indep C \\
\nicefrac{1}{8} &  \text{if } r \in \{ \ac{++}, \ac{-+} \}\\
\nicefrac{1}{8} &  \text{if } r \in \{ \ca{++}, \ca{-+} \}
\end{cases}
\end{equation}

Based on the causal relation $r$, we then sample probability distributions according to the procedure described above.
A visualization of the sampled states is given in \Cref{fig:table-distributions} which shows histograms of the probabilities for each of the four possible worlds across all sampled states for three selected causal relations.\footnote{The source code and all modeling results are publicly available: \url{https://osf.io/6bshq/?view_only=1703a3417a1343a5a66b78ac8ce206c2}.}

\subsection{Communicating causal information implicitly via conditionals}
\label{sec:comm-caus-inform}

So far, we have specified how the joint probability distributions are derived under the assumption of particular causal relations.  
That is,  strictly speaking, our world states have two components, a probability distribution $s$ and a causal relation $r$.
Independently of the causal relation from which $s$ originates, the choice probabilities of a speaker who aims to communicate her beliefs about the world can then be written as:

\begin{align}\label{eq:speaker-long}
    & P_{\text{S}}(u\mid r,  s) =  P_{\text{S}}(u\mid s) \propto \text{exp}(\alpha \cdot (\text{log P}_{\text{lit}}(s\mid u)))   \\
& \text{where } P_{\text{lit}}(s\mid u) = \sum_{r^\prime} P_{\text{lit}}(r^\prime, s \mid u) \nonumber
\end{align}
The speaker's goal of communication, when using conditional sentences, that we assume in this paper, is first and foremost to convey their beliefs about the antecedent and the consequent. 
In other words, we start by exploring a probabilistic model of communication with conditionals from the most austere assumption, namely that not only the assertability of a conditional does \emph{not} hinge on any putative causal relation necessarily, but that also the purpose of communication itself is \emph{not} to directly communicate information about the causal relation.\footnote{
There are circumstances where the goal of the communication reasonably includes the causal relation which we leave for future work here. 
}

\begin{figure}
\centering
\begin{minipage}{0.15\textwidth}
\begin{tikzpicture}
\node[latent] (A) [] {$A$};
\node[latent] (C) [below=of A] {$C$};
\edge {A} {C};
\end{tikzpicture}
\end{minipage}
\begin{minipage}{0.35\textwidth}
\begin{align*}
& P(A=a) = 0.65 \\
& P(C=c\mid A=a) =  \nicefrac{12}{13} \\
& P(C=c\mid A=\neg a) =  \nicefrac{1}{7}
\end{align*}
\end{minipage}
\caption{Bayes net representing $s_2$  from \Cref{tbl:states-toy-example} consisting of a graphical representation (left) and a set of associated (conditional) probabilities (right) that define $P^{(s_2)}(A,C)$.}
\label{fig:graph-s2}
\end{figure}
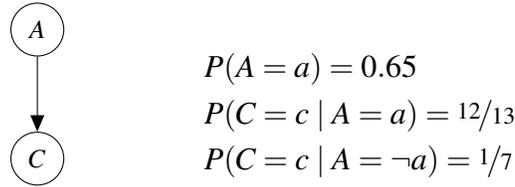

Before we discuss the results of our simulations, we would like to add a final note on the interpretation of the world states in our model.
Instead of considering them as simple pairs consisting of a probability distribution and a causal relation,  we can also think of them as \emph{causal Bayesian networks}  \citep{pearl88,Pearl2009, Pearl2014}, henceforth abbreviated as `Bayes nets'.
They have, similar to the closely related formalism of Structural Causal Models, a rich tradition of supporting semantic theories of conditionals already, \citep[e.g.][]{hiddleston05,Pearl2009,Pearl2013,Rips2010,briggs12,kaufmann13,Lucas2015,santorio16,Lassiter2017}.
Bayes nets represent sets of variables (here binary variables) and the dependencies among them; they consist of a directed, acyclic graph that defines the relations among variables, and a set of conditional probabilities for each variable given all possible instantiations of all its \emph{direct} parent nodes, which simplify to unconditional probabilities for variables without parent nodes.
This set of (conditional) probabilities is sufficient to define the joint probability distribution over all variables represented in the graph; see \Cref{fig:graph-s2} for a Bayes net that represents the joint probability distribution $s_2$ from \Cref{tbl:states-toy-example}.

In sum, the model we explore here can be seen as capturing the implicit communication of causal information by (i) treating world states as causal Bayes nets, (ii) identifying the purpose of utterance (the \emph{question under discussion} \citep{Roberts2012:Information-Str} or the relevance projection \citep{Kao2014a}) to be the precise communication of the probability table $s$ (the speaker's beliefs about joint truth of $A$ and $C$) and (iii) a pragmatic process of utterance generation favoring true and informative utterances.

\section{Pragmatic inferences from conditionals in default contexts}
\label{sec:results}

In this section, we explore model predictions in default contexts.
We can think of these as the model's predictions generalized over a wide range of more specific contexts, or, relatedly, as the model's predictions for the interpretation of utterances of conditionals in unbiased, out-of-the-blue contexts.
We will particularly look at the listener's inferences from an utterance of a conditional about the speaker's uncertainty about $A$ and $C$,  the speaker's beliefs about any systematic relation between $A$ and $C$, and the strength of a conditional perfection reading.
Doing so, we show how informativity-driven pragmatic choice of utterances leads to \emph{de facto} assertability conditions as postulated by inferentialist accounts without having to stipulate these directly.\footnote{All results reported below were obtained with the literal meaning threshold $\theta = 0.9$, and the rationality parameter $\alpha=3$. Qualitatively identical results were obtained for parameters from a grid with $\theta \in [0.9, 0.95,  0.975], \alpha \in [1, 3,5, 10]$.}

\paragraph{Setting the scene.} 
Let us reconsider example (\ref{bsp:out-of-the-blue-ifac}), repeated here from above,  now uttered in an out-of-the-blue context.

\begin{exe}
\exr{bsp:out-of-the-blue-ifac}
  Bigi: If Alex comes to the party, Chris comes too. \hfill \textcolor{gray}{[``If $A$, $C$.'']} \\
  Wobo: Who are these guys? What party are you talking about?
\end{exe}

\noindent 
Even without any strong prior convictions, Wobo is likely to draw pragmatic inferences from Bigi's utterance.
Intuitively, Wobo would infer that Bigi is uncertain about whether Alex comes to the party, similarly for Chris, and that Chris' coming to the party is not entirely unrelated to Alex's coming.
Moreover, the implicit dependency between antecedent and consequent may be interpreted to be so strong that the conditional is understood as biconditional; it does not seem unnatural for Wobo to infer that, according to Bigi, Chris comes to the party \emph{only if} Alex comes.

To analyse the listener's \emph{a posteriori} beliefs and the speaker's utterance choices we make use of the following definitions concerning the modeled states.
The set of states in which the speaker is \emph{uncertain about event $X$} contains all and only states $s$ such that the probability of $X$ in $s$, $P^{(s)}(X)$, is neither too low nor too high:
\begin{equation} \label{eq:speaker-uncertainty}
\text{Uncertain}(X) = \{ s \mid 1-\theta \leqslant P^{(s)}(X) \leqslant \theta \}
\end{equation}
A similar construction captures the speaker's \emph{certainty about whether $X$ is true}:
\begin{equation}\label{eq:speaker-certainty}
\text{Certain}(X) = \{ s \mid P^{(s)}(X) >  \theta \} \cup  \{ s \mid P^{(s)}(X) < 1 - \theta \}
\end{equation}
The uncertainty / certainty about $X$ given $Y$ is captured analogously. 

\paragraph{Hyperrational utterance choices in default contexts.}

\begin{figure}
\centering
\scalebox{0.65}{\input{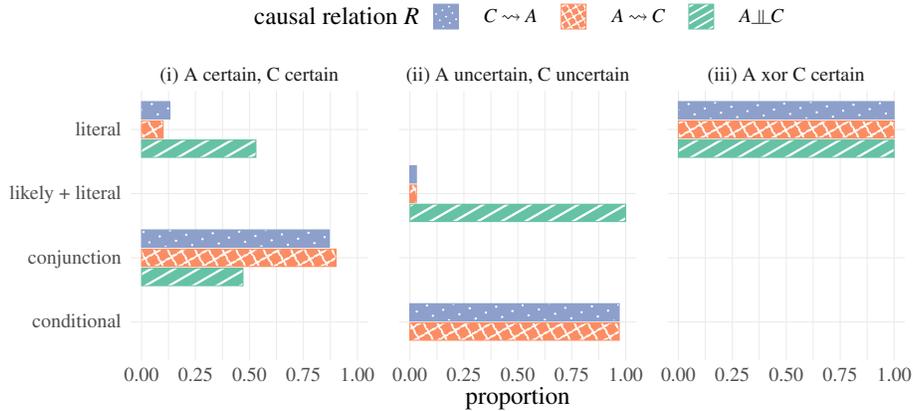}}
\caption{Relative frequency of how often each utterance type is the speaker's \emph{best} choice  for a set $S$ of 10,000 states sampled from the prior (default context), given that the speaker is (i) certain or (ii) uncertain about $A$ and about $C$, i.e., $\forall s \in S: s\in \text{Certain}(A)\wedge s\in \text{Certain}(C)$, respectively $\forall s \in S: s\in \text{Uncertain}(A)\wedge s\in \text{Uncertain}(C)$, or (iii) the speaker is uncertain about the truth of one proposition but certain about the truth of the other, e.g., $\forall s \in S: s\in \text{Uncertain}(A) \wedge s\in \text{Certain}(C)$.}
\label{fig:speaker-uncertainty-best}
\end{figure}

According to the model presented here, the listener's inferences about the speaker's epistemic state
can be put down, at least in part, to a Q-implicature \citep[][]{Atlas1981a,Horn1984}: from the fact that the speaker decided \emph{not} to utter a more specific and thus more informative utterance than $A\rightarrow C$, the pragmatic listener should deem it unlikely that the speaker refers to a state $s$ in which a more informative utterance would also apply.
To see whether such an alternatives-based explanation is endorsed by the present modeling setup, we first look at a speaker who always chooses the utterance with the highest utility (hyperrational speaker where rationality parameter $\alpha \rightarrow \infty$).
Figure~\ref{fig:speaker-uncertainty-best} shows model predictions from the speaker's point of view for different probabilistic beliefs with respect to propositions $A$ and $C$ given the causal relation between antecedent and consequent.
It clearly shows that a (hyperrational) speaker chooses utterances in dependence of her belief state: when the speaker is uncertain about both propositions, the best utterance is either a conditional or ``\emph{likely} $\Phi$'' which are the two least informative utterance types.\footnote{For simplicity, we discuss all results assuming an epistemic interpretation of the probabilities in our world states,  but an interpretation based on objective chance is equally applicable.}
On the other hand, the speaker's best utterance is either a conjunction or a literal when she is certain about both propositions, that is, when she is certain that $A$ is true (or false) and certain that $C$ is true (or false), but she is not necessarily certain that a conjunctive event occurs (e.g. $A=a \wedge C=\neg c$). 
This is also the reason, why conjunctions do not seem to be preferred over literals when  $A$ and $C$ are independent  and the speaker is certain about $A$ and certain about $C$ (\Cref{fig:speaker-uncertainty-best} (i)). 
In this case, when the speaker's best utterance is a literal, there simply is no conjunction that truthfully describes the given state (e.g., $P^{(s)}(a,c)=0.82,  P^{(s)}(a, \neg c) = 0.08,  P^{(s)}(\neg a,c)=0.08,  P^{(s)}(\neg a, \neg c)=0.02$).
When the speaker is only certain about one proposition, her best utterance is a literal (\Cref{fig:speaker-uncertainty-best} (iii)).  

\Cref{fig:speaker-uncertainty-best} (ii) further reveals that, when the speaker is uncertain about both $A$ and $C$,  her utterance choice is strongly influenced by their causal relation: for these states the speaker's best utterance will always be ``\emph{likely} $\Phi$'' when $A$ and $C$ are independent,  whereas when  $A$ and $C$ are dependent, it will almost certainly be a conditional.

\paragraph{Inferences about causal dependency.}

Since pragmatic interpretation is here modeled as backwards-inference based on the speaker's utterance choice protocol, we can already anticipate from the above results of hyperrational speakers that pragmatic interpreters may draw rather specific inferences about the causal relation between $A$ and $C$ from an utterance of $A \rightarrow C$.
In what follows,  we look at inferences about the causal relationship in more detail, assuming ``normal speakers'' ($\alpha = 3$).

\begin{figure}
\centering
\scalebox{0.75}{\input{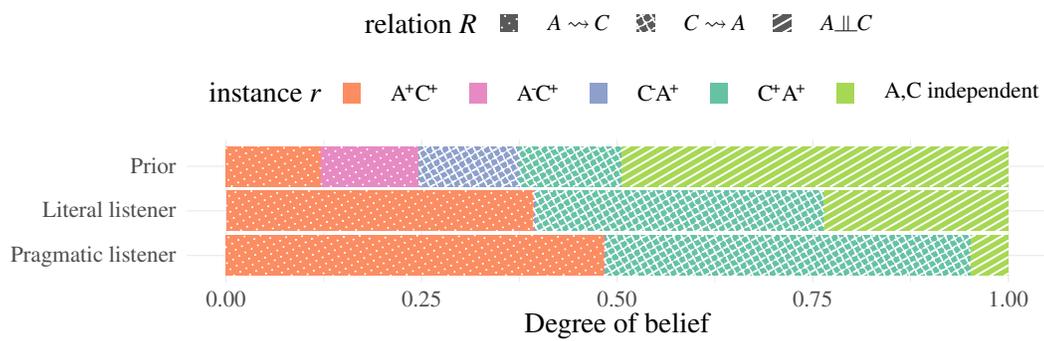}}
\caption{Degree of belief in each causal relation (A$^+$C$^+$ is shorthand for \ac{++},  analogous for other relations) at all three levels of interpretation; prior to uptake of the conditional $A\rightarrow C$ and \emph{a posteriori} given a literal / pragmatic listener.}
\label{fig:evs-relations}
\end{figure}

Figure~\ref{fig:evs-relations} shows the causal inferences that listeners draw about $A$ and $C$ when the speaker utters the conditional $A\rightarrow C$.
We observe that \emph{a posteriori} the listener (literal and pragmatic) assigns very low probability to states where $r=$ \ac{-+} or $r=$ \ca{-+}, as these are very unlikely to give rise to a probability table in which the conditional $A \rightarrow C$ is assertable.
Interestingly, the listener is not committed to a single underlying causal relation, but instead merely infers that there is a positive dependency between antecedent and consequent: $A$ tends to be true (false) when $C$ is true (false) and vice versa. 
The pragmatic listener assigns almost the entire probability mass to the corresponding causal relations (\ac{++}, \ca{++}), the literal listener approximately 75\%.

This result suggests that also under a pragmatic interpretation, the listener needs further knowledge to disambiguate the underlying causal structure since, in this most general context, it is not possible to infer that the antecedent is a cause of the consequent or that, as under a diagnostic reading of the conditional $A\rightarrow C$,  the consequent is a cause of the antecedent.

Another interesting result concerns the states where $A$ and $C$ are independent:  even though the literal listener largely diminishes her beliefs in the independence of $A$ and $C$ as compared to her beliefs prior to the speaker's utterance of the conditional $A\rightarrow C$, the pragmatic listener assigns a still smaller probability to states where antecedent and consequent are causally independent.
This is an important and interesting result that bears emphasis.
Since the pragmatic listener combines a rich representation of the speaker's beliefs about truth of propositions and their causal relation with Gricean pragmatic reasoning, the pragmatic listener concludes more about the (speaker's beliefs about the) causal structure of the world than is entailed by the semantics of a conditional.
This additional causal-pragmatic inference essentially rides piggyback on standard Gricean Quantity reasoning.

To see this, let us consider the perspective of the hyperrational speaker again (Figure~\ref{fig:speaker-uncertainty-best}).
Those states where $A$ and $C$ are independent  are  exclusively  states for which an informative Gricean speaker will either prefer to utter a bare proposition (conjunction or literal) which is more informative than a conditional or ``\emph{likely} $\Phi$''.
As a result, the pragmatic listener infers a causal relation from entirely unbiased assertability conditions for conditionals and standard Gricean Quantity reasoning.
Causal inference comes up as a pragmatic inference without having to stipulate an additional pragmatic constraint concerning a (causal) relation between propositions, let alone hard-coding such a requirement in the semantic meaning.
\begin{figure}
\centering
\scalebox{.8}{\input{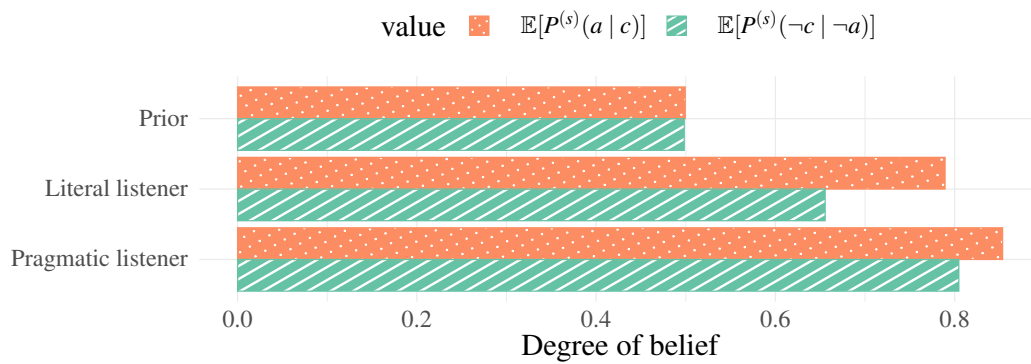}}
\caption{Degree of belief in the two CP-related conditional probabilities $P^{(s)}(\neg c\mid \neg a)$ and $P^{(s)}(a \mid c)$; prior to the uptake of the conditional $A\rightarrow C$ and \emph{a posteriori}, given a literal or pragmatic listener.}
\label{fig:cp-evs-probs}
\end{figure}
\paragraph{The strength of conditional perfection readings.}
In the example given in the beginning of this section, it does not seem astounding to interpret the ``if'' in ``\emph{If Alex comes to the party, Chris comes too}'' as  ``\emph{if and only if}''.
It might even seem to be a quite acceptable, if not natural, inference although we did not specify any further context and although, from a logical point of view, this inference is \emph{not} valid.

The phenomenon that conditionals are sometimes interpreted as biconditionals,  known as \emph{conditional perfection} (CP),  has caught much attention in the literature, especially since \citet{Geis1971}. 
CP remains  a topic of ongoing debate; no consensus has for instance been found concerning its prevalence or the circumstances that trigger a CP reading \citep[e.g.~see][]{newstead_conditional_1997,  VonFintel2001,  oberauer_meanings_2003, moldovanDenyingAntecedentConditional2013}.

Neither exists a standard measure that quantifies the degree to which a conditional receives a CP reading. 
Here we will refer to two inferences that are prominently considered in the literature on conditional reasoning, `\emph{Denying the antecedent}' (DA) and `\emph{Affirming the consequent}' (AC), shown below.  
The endorsement of DA or AC inferences suggests that participants interpret the conditional as biconditional since only then these are logically valid inferences \citep[e.g.,][]{newstead1997conditional, evansThinkingConditionalsStudy2007}.
\begin{itemize}
\item [DA:]  \ \ \ \ $A \rightarrow C.\ \  \neg A.\ \  \therefore \ \  \neg C$.
\item [AC:] \ \ \ $A \rightarrow C. \ \ C.\ \  \therefore \ \ A$.
\end{itemize}
To learn how strongly $A\rightarrow C$ is interpreted as biconditional, we will therefore look at the listener's expected beliefs (about the speaker's beliefs) $P^{(s)}(\neg c\mid \neg a)$ and $P^{(s)}(a\mid c)$.
Only when ``\emph{if}'' is interpreted as ``\emph{iff}'' the two considered quantities should be large. 

As can be seen in Figure~\ref{fig:cp-evs-probs}, without any further contextual assumptions our model predicts a quite strong CP-reading for the interpretation of conditionals in the default context.
The speaker's utterance of $A\rightarrow C$ elicits an increase in the listener's beliefs (about the speaker's beliefs) in the conditional probabilities $P^{(s)}(\neg c \mid \neg a)$ and $P^{(s)}(a \mid c)$ as compared to her prior beliefs.
This is true for the literal and the pragmatic interpretation, yet the pattern is more pronounced in the latter.
This result is to a great extent due to the representation of dependent world states as noisy-or models (see \Cref{sec:prior-states}).
In this way, this interpretation is explained as something akin to an I-implicature \citep{Atlas1981a,Levinson2000}, as suggested by \citet{Horn2000}; CP-readings are supported in large part by what is arguably a cognitively economic, perhaps stereotypical representation format of causal dependency between events.
However, we also see that pragmatic reasoning about alternatives further strengthens the CP-reading quantitatively, similar to the accounts of \citet{VanDerAuwera1997} and \citet{VonFintel2001} which share with our model the feature that CP is a result of a listener reasoning about the speaker's production protocol \dash given \posscitet{VonFintel2001} account, including, crucially, the possibility that the speaker could simply have asserted the consequent.

\paragraph{Deriving inferentialist assertability conditions.}
So far, we showed that the current modeling setup predicts that speakers will use conditional sentences predominantly in cases where there is a (causal/inferential) relationship between antecedent and consequent and that listeners, therefore, infer such a relationship from an utterance of a conditional.\footnote{
Our model predicts the dependency relation to be a defeasible inference, but it is not predicted to arise in any context (e.g., see the discussion on missing-link and biscuit conditionals in  \Cref{sec:missing-links-bcs}). Therefore, it is also compatible with \posscitet{lassiter_decomposing_nodate} account of when the dependency between antecedent and consequent will (or will not) arise based on discourse coherence.}
These predictions particularly challenge the idea advanced by advocates of inferentialism that a (causal/inferential) relation between antecedent and consequent is part, in whatever form, of the core semantics of conditionals.
We here argue that an austere assertability condition for conditionals in combination with rich representations of contextual (causal) world knowledge and pragmatic reasoning is sufficient to derive the kind of assertability conditions postulated by inferentialist accounts.
In particular, we will consider the assertability condition formulated in Equation \eqref{eq:delta-p-star} which was proposed by \citet{VanRooij2019} improving on like-minded work of \cite{Douven2008}.

\begin{equation}  \label{eq:delta-p-star}
 A \rightarrow C \text{ is acceptable/assertable only if } \Delta^* P = \frac{P(c \mid a) - P(c \mid \neg a)}{1-P(c \mid \neg a)} \geqslant \theta \text{.}
\end{equation}

\noindent
The general idea behind assertability criteria of this kind is that an utterance of a conditional $A \rightarrow C$ is acceptable only if $P(c \mid a)$ is high (like we assume as well here) and, in addition, $P(c \mid \neg a)$ is low,  that is,  $C$ being true is only likely when $A$ is true.

\begin{figure}
\centering
\includegraphics[width=\textwidth]{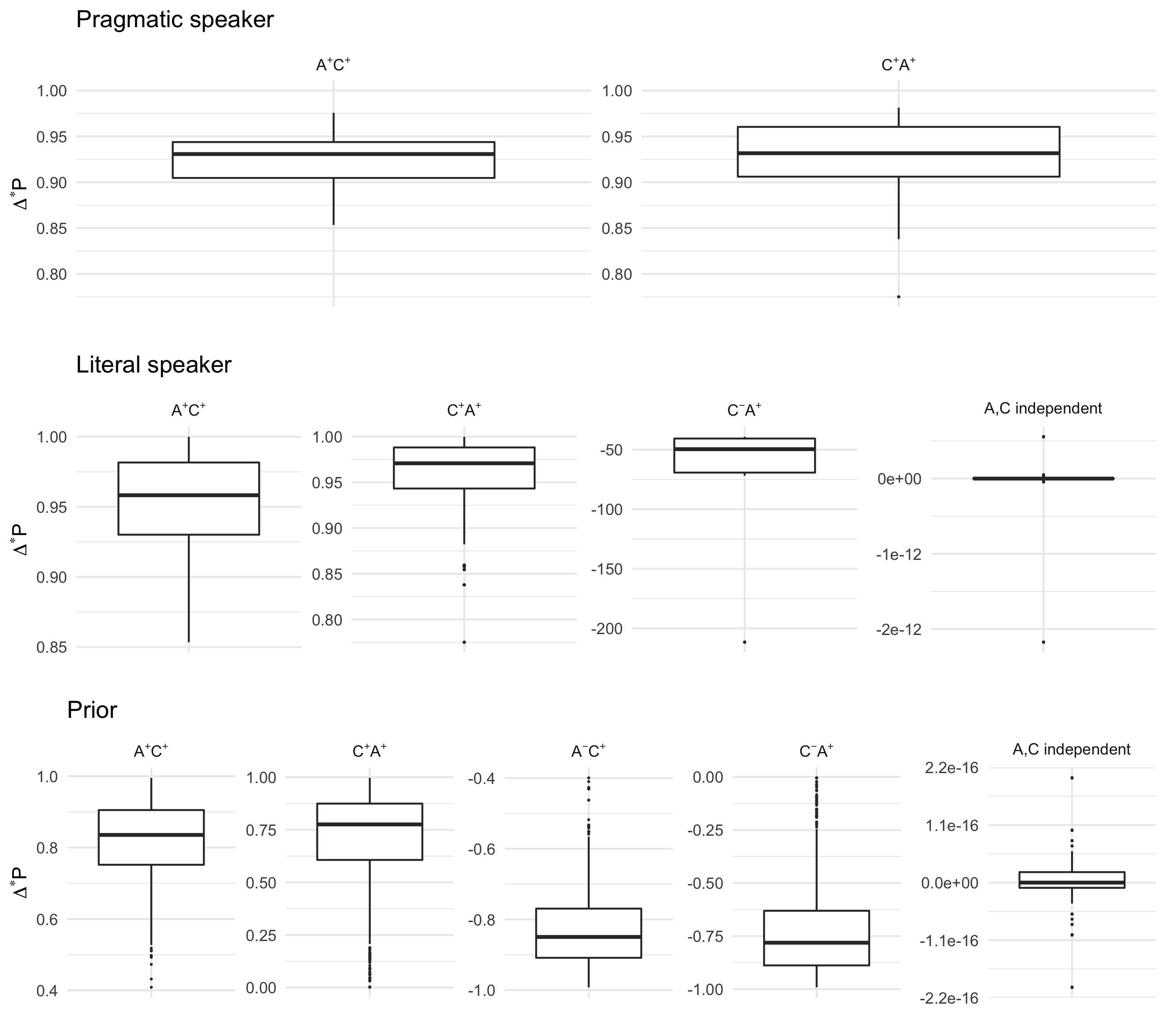}
\caption{Distribution of $\Delta^* P$ values for 10,000 states randomly sampled from (i) the prior (bottom),  (ii) from the prior given $A \rightarrow C$ is assertable (middle), and (iii) for states from (ii) where additionally $A \rightarrow C$ is the pragmatic speaker's best choice (top).  Causal relations are abbreviated, e.g., A$^+$C$^+$ is shorthand for \ac{++}.}

\label{fig:accept_conditions}
\end{figure}

We do not argue that Equation \eqref{eq:delta-p-star} necessarily holds, empirical research is needed to find out whether the acceptance/assertability of conditionals is related to such criteria. 
\footnote{While some of the empirical studies that have been conducted to date suggest that the relationship between antecedent and consequent has an influence on the assertability of conditionals \citep[e.g.,see][]{douven_indicatives_2012, Skovgaard-Olsen2016}, others found no such effect \citep[e.g., see][]{singmann_new_2014, oberauer_what_2007}.}
Instead, we aim to investigate how the model that we propose here relates to accounts that propose this kind of assertability conditions for conditionals, possibly bringing Inferentialists' ideas and pragmatic reasoning closer together.

We find that our pragmatic model for the use of conditionals derives the criterion from Equation \eqref{eq:delta-p-star},  in the sense that whenever the model predicts a (hyperrational) speaker to use a conditional in some state $s$, the value $\Delta^{*}P$ calculated from the probability table entailed by $s$ is indeed very high.
To see this, Figure~\ref{fig:accept_conditions} shows the distribution of $\Delta^{*}P$ measures for three sets of states: (i) sampled from the prior (default contexts), (ii) sampled from the prior conditioned on $A \rightarrow C$ being assertable (literal speaker condition), and (iii) the subset of states sampled in the literal speaker condition, in which a hyperrational speaker would utter $A \rightarrow C$ (pragmatic speaker condition).
Figure~\ref{fig:accept_conditions} reveals, reassuringly, that $\Delta^{*}P$ is not always high for any state sampled from the prior.
It also shows that just from our austere assertability condition alone, namely $P^{(s)}(c \mid a) \geqslant \theta$, the average associated $\Delta^{*}P$ increases, even if there are still quite a number of cases where what we may call a ``literal speaker'' might use $A \rightarrow C$ while the measure $\Delta^{*}P$ is quite low.
This clearly shows that, despite biases for ``simple situations'' introduced by \emph{noisy-or} parameterization of state priors, the assertability condition that $P^{(s)}(c \mid a) \geqslant \theta$ alone does not guarantee that $\Delta^{*}P$ is high.
But for the pragmatic speaker, we see that the $\Delta^{*}P$ measure is exclusively very high, thus lending support to the idea that a straightforward model of Gricean pragmatic use of conditionals explains an otherwise stipulative assertability condition for conditionals.

\begin{figure}
\centering
\scalebox{0.8}{\input{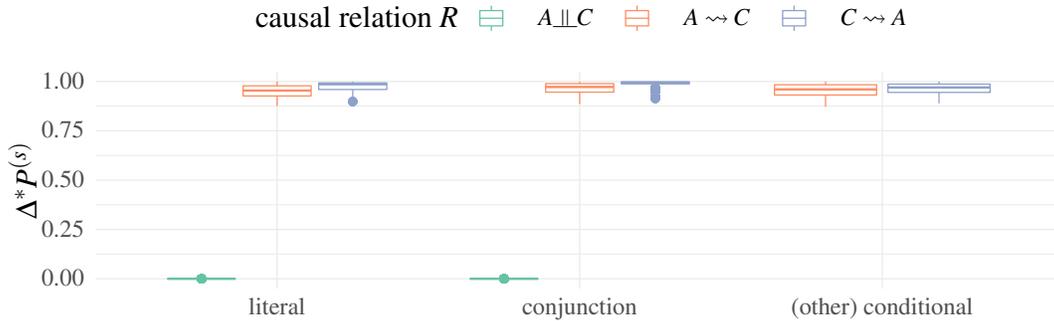}}
\caption{$\Delta^* P^{(s)}$ values, zoomed into the range where  $0 \leqslant \Delta^* P^{(s)} \leqslant 1$ for $s$ from a set of 10,000 states sampled from the prior given $A\rightarrow C$ is assertable (literal speaker condition in \Cref{fig:accept_conditions}) where $A\rightarrow C$ is \emph{not} the best choice of a hyperrational speaker ($\alpha=\infty$).}
\label{fig:speaker-p-rooij-large-best-not-ac}
\end{figure}

We saw that when $A\rightarrow C$ is asserted by our pragmatic speaker,  $\Delta^{*}P$ is large,  but is it also the case that a large value of $\Delta^{*}P$ is sufficient for a hyperrational pragmatic speaker to utter the conditional $A\rightarrow C$? 
If it was a sufficient condition,  our pragmatic speaker should prefer this conditional whenever $\Delta^{*}P$ is large.
\Cref{fig:speaker-p-rooij-large-best-not-ac} shows boxplots in the range $0 \leqslant \Delta^{*}P \leqslant 1$ of the $\Delta^{*}P$-values of states where $A\rightarrow C$ is assertable, but \emph{not} the most likely utterance for a hyperrational speaker.\footnote{For states where $R = C \rightsquigarrow A$,  $\Delta^*P$  has a minimum value of -212 when the hyperrational speaker's best utterance is a literal,  and a minimum value of -69.3 when it is a conjunction.  When $R = A \indep C$, $\Delta^{*}P^{(s)}$ clusters closely around 0.}
The utterance type of the hyperrational speaker's best choice for the respective states \dash a literal, a conjunction or a conditional other than $A\rightarrow C$ \dash is shown on the x-axis; color codes represent the causal relation of the states.
Clearly, there are states where $\Delta^{*}P^{(s)}$ is large and the speaker chooses a different utterance than $A\rightarrow C$. 
Particularly interesting for us are those states where the speaker does not choose a different conditional but an utterance that is more informative than a conditional, that is, a conjunction or a literal.
These  are situations where the predictions of our model diverge from the predictions of accounts arguing that a conditional is a assertable/acceptable when $\Delta^{*}P$ is large: although $A\rightarrow C$ is (literally) assertable, the speaker may have good reasons to choose a different utterance (to be maximally informative) and so,  the conditional $A\rightarrow C$ might still be rejected.
This explains why a criterion like the one defined in Equation \eqref{eq:delta-p-star} might be a necessary, but not a sufficient condition for the assertability of a conditional.\newline

Overall the predictions from our model corroborate the hypothesis that the dependency relation  does not need to be incorporated into the semantics of conditionals.
The model shows that pragmatic reasoning about the speaker's conditional utterance $A\rightarrow C$ is a possible way to derive the dependency relation \dash given an appropriate representation of  the modeled variables, in particular their underlying causal structure.

\section{Interpretation in concrete contexts: Douven's puzzle}
\label{sec:douvens-puzzle}

So far we showed that an RSA model combined with unbiased default priors is able to explain general pragmatic inferences associated with utterances of conditionals.
In the following we will turn towards predictions for utterances of conditionals in rather specific and explicitly given contexts of use.
In particular, we investigate how the present setup helps explain a puzzle put forward by \citet{Douven2012b}.
\citeauthor{Douven2012b} contrasts three cases of conditionals uttered in concrete contexts, that we will fully cite below.
In these contexts, learning a conditional $A \rightarrow C$ either leads to an increase (the Skiing Example given in \pref{itm:skiing}, discussed in \Cref{sec:skiing}) or a decrease (the Garden Party Example given in \pref{itm:garden-party}, discussed in Section~\ref{sec:garden-party}) in the listener's degree of belief in the truth of $A$ or the listener's degree of belief in the antecedent $A$ does not change at all (the Sundowners Example given in \pref{itm:sundowners}, see Section~\ref{sec:sundowners}).

We argue here that an explicit representation of contextually-grounded world knowledge that does not only comprise plausible prior beliefs about the probability of the represented variables, but also caters for their causal structure, is sufficient to explain how pragmatic listeners adjust their beliefs about the antecedent in one way or another after receiving information in form of a conditional.
The key to the explanation we propose here is that, in each case, the listener learns a piece of causal information, that is, the listener learns that propositions are causally related where this was previously not expected or deemed rather unlikely.\footnote{A perhaps more realistic picture would be to model a listener as completely \emph{unaware} of the causal relation in question; that agent's stock of explicitly entertained alternatives would just not represent that contingency. 
Since adding agent's unawareness to a model of pragmatic reasoning is possible but technically rather involved \citep[e.g.][]{HeifetzMeier2006:Interactive-Una,FrankeJagerde-Jager2010:Now-that-you-me,Franke2012:Pragmatic-Reaso}, this paper makes the simplifying assumption that the listener is aware of the possible causal connection but deems it very unlikely to begin with.}

\subsection{The skiing case}
\label{sec:skiing}

The Skiing Example is a case where, intuitively,  the listener's degree of belief in the antecedent increases.

\par
\begingroup
\leftskip2em
\rightskip\leftskip
\noindent
\paragraph{The Skiing Example \citep{Douven2012b}.}
Harry sees his friend Sue buying a skiing outfit. This surprises him a bit, because he did not know of any plans of hers to go on a skiing trip. He knows that she recently had an exam and thinks it unlikely that she passed. Then he meets Tom, another friend of Sue’s, who is just on his way to Sue to hear whether she passed the exam, and who tells him:
\begin{exe}
  \ex \label{itm:skiing} If Sue passed the exam, her father will take her on a skiing vacation.\\
        $\leadsto$ listener belief in antecedent \emph{increases}
\end{exe}
\par
\endgroup

\noindent
In this example, there are three relevant propositions: $E$ (Sue passed the exam), $S$ (Sue goes skiing) and $C$ (Sue buys skiing clothes).
The listener Harry has observed $C$, so his beliefs in $C$ are high, possibly 1.
The speaker Tom utters the conditional $E \rightarrow S$.
We want to explain how this can lead to an increase in Harry's probabilistic beliefs about $E$.

Our explanation hinges on assuming that Harry has certain plausible beliefs about the propositions involved and their causal relationship.
It proceeds in three steps:
\begin{enumerate}
  \item From pragmatic reasoning, the listener infers from the utterance of $E \rightarrow S$ that the speaker likely believes in a causal relation $E \overset{\scriptscriptstyle{++}}{\rightsquigarrow} S$: Sue passing the exam increases her chance to go on a skiing trip.
  \item The listener takes the speaker to be an authority on the matter and, at least to a certain extent, also increases degrees of beliefs in the causal relation $E \overset{\scriptscriptstyle{++}}{\rightsquigarrow} S$.
  \item Since the listener also has a high degree of belief in $C$, and given that it is plausible to assume that in general a relation of the kind $S \overset{\scriptscriptstyle{++}}{\rightsquigarrow} C$ holds, the listener ends up with a higher degree of belief in $E$ after processing the utterance.
\end{enumerate}

\begin{figure}
\subfigure[$s_{dep}$: $E, S$ dependent \label{fig:graph_ski_dep}]{\begin{tikzpicture}
\node[latent, label={right: $P(e)=0.2$}] (E) [] {$E$};
\node[latent, label={right: $\begin{aligned}
& P(s\mid e)=1 \\
& P(s\mid \neg e) = 0
\end{aligned}$
}] (S) [below=of E] {$S$};
\edge {E} {S}; 
\end{tikzpicture}}\qquad
\subfigure[$s_{ind}$: $E, S$ independent \label{fig:graph_ski_ind}]{
\begin{tikzpicture}
\node[latent, label = {right: $P(e)=0.2$}] (E) [] {$E$};
\node[latent, label = {right: $P(s)=1$}] (S) [below=of E] {$S$}; 
\end{tikzpicture}}
\subfigure[Listener's degree of belief in the antecedent\label{fig:results_ski_pragmatic}]{\scalebox{0.55}{\input{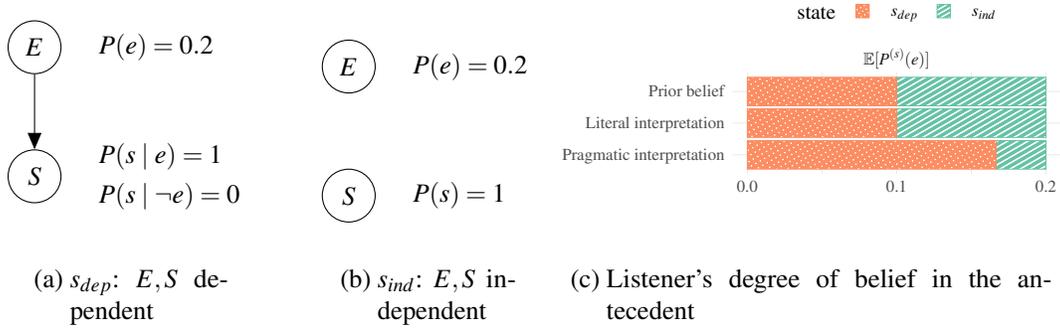}}}
\caption{Bayes nets and results for the pragmatic reasoning part in the Skiing Example with $\mathcal{U}=\{S, \text{likely } S,  E\rightarrow S\}$, $\alpha=1$, $E$: pass exam, $S$: go skiing.  Both states, $s_{dep}$ and $s_{ind}$, are assigned equal prior probability.}
\label{fig:ski_pragmatic}
\end{figure}

To illustrate this reasoning schema, we offer one concrete context model for the Skiing Example in Figure~\ref{fig:ski_pragmatic}.\footnote{A more realistic choice than using a single independent state with $P^{(s_{ind})}(s)>\theta$, would be to include several independent states, e.g. with $P^{(s)}(s)\in [0, 0.5, 1]$ which would, however, require a larger set of utterances such that for every state there is at least one utterance assertable.  
For the sake of a simpler discussion, here and in the following examples, we give just one set of concrete numbers of the relevant (conditional) probabilities. }
There are two Bayes nets that are in line with the speaker's utterance $E\rightarrow S$, one in which passing the exam stands in a direct causal relation to going on a skiing trip (\Cref{fig:graph_ski_dep}),  and one in which it does not (\Cref{fig:graph_ski_ind}).\footnote{Even though it seems less probable, there is the possibility that the causal relation between $E$ and $S$ is reversed (i.e., $S\rightarrow E$ instead of $E\rightarrow S$); Sue may for instance study extra hard \emph{because} her father invites her to go on a skiing trip.  Note that in this case, the listener's observation of Sue buying skiing clothes would, however, not increase the listener's degree of belief in Sue passing the exam. }

Their entailed joint probability distributions are spelled out in \Cref{tbl:joint-prob-skiing}.
The situation model in Figure~\ref{fig:ski_pragmatic} gives the listener's (Harry's) beliefs about what Tom might believe about the relation between $E$ and $S$.
Given the context-story, there is no indication that Harry believes that Tom believes that the dependent Bayes net is any more or less likely than the independent one.
Therefore, we assign equal prior probability to both Bayes nets, even though this is not crucial for the case at hand.

\begin{table}
\scriptsize
\begin{center}
\subfigure[$P^{(s_{dep})}(E,S)$	]{\begin{tabular}{lll}
\toprule
					& $s$ 												& $\neg s$ \\ \midrule
$e$ 			&	$P(e)\cdot P(s\mid e)=0.2$ 				& $P(e) \cdot P(\neg s\mid e)=0$\\ \midrule
$\neg e$	& $P(\neg e) \cdot P(s\mid \neg e)=0$ 	& $P(\neg e) \cdot P(\neg s\mid \neg e)=0.8$ \\ \bottomrule
\end{tabular}
}
\subfigure[$P^{(s_{ind})}(E,S)$]{\begin{tabular}{lll}
\toprule
					& $s$ 										& $\neg s$ \\ \midrule
$e$ 			&	$P(e)\cdot P(s)=0.182$ 			& $P(e) \cdot P(\neg s)=0.018$\\ \midrule
$\neg e$ 	& $P(\neg e) \cdot P(s) = 0.728$	& $P(\neg e) \cdot P(\neg s)=0.072$ \\  \bottomrule
\end{tabular}}
\end{center}
\caption{Entailed joint probability distributions of variables $E,S$ of the two Bayes net $s_{ind}, s_{dep}$ shown in \Cref{fig:ski_pragmatic}; $E:$`pass exam', $S$: `go on skiing trip'. }
\label{tbl:joint-prob-skiing}
\end{table}

With the set of alternative utterances $\mathcal{U}=\{S, \text{likely } S,  E\rightarrow S\}$, $\alpha=1$ and the probabilities and causal relations as shown in \Cref{fig:ski_pragmatic}, we get the following probability distribution for the pragmatic listener when $u=E\rightarrow S$: $P_{\text{PL}}(s=s_{\text{dep}}\mid u = E\rightarrow S) = \nicefrac{5}{6}$ and $P_{\text{PL}}(s=s_{\text{ind}}\mid u = E\rightarrow S) = \nicefrac{1}{6}$; see \Cref{tbl:skiing} for the respective values of other model components.
The dependent Bayes net becomes more likely under a pragmatic interpretation only: by taking into account the fact that the speaker could have chosen a more informative utterance (e.g.~``Sue goes on a skiing trip'' (S)) to refer to the independent Bayes net (see Table \ref{tbl:skiing}),  the listener learns that, most likely, the speaker believes in a connection between Sue going on a skiing trip and her passing the exam.
Contrary to that, under a literal interpretation both Bayes nets remain equally likely as the chosen utterance, $E\rightarrow S$, is literally true in both states.

The predicted interpretation of Tom's utterance, $E\rightarrow S$, with respect to the probability of the antecedent is shown in \Cref{fig:results_ski_pragmatic}: the listener's degree of belief about the speaker's beliefs in the antecedent remains at an expected value of 0.2,  only the degree of belief related to the causal relation between $E$ and $S$ is influenced by the speaker's utterance of the conditional.
Crucially, since we so far only modelled the listener's pragmatic inferences about the speaker's beliefs, the listener's observation ($C=c$) is not considered yet.

How should a listener change their beliefs in the light of a (pragmatically derived) belief about the speaker's beliefs?
That depends on the more general assumptions the listener makes about the speaker: is she trustworthy and well-informed on the subject matter at hand?
Since nothing in the scenario described in Example~(\ref{itm:skiing}) gives us reason to expect otherwise, we may follow the usual assumptions in Gricean belief-based reasoning that the speaker is cooperative and knowledgeable \citep[e.g.][]{Geurts2010:Quantity-Implic}.
Even if the precise effect of integrating beliefs of a cooperative and knowledgeable agent into one's own beliefs are elusive, the effect in a scenario like the one at hand is most likely that the listener increases their own beliefs in the relation $E \overset{\scriptscriptstyle{++}}{\rightsquigarrow} S$.
To keep matters simple for a fully fleshed out numerical example, we just assume that the listener adopts exactly the same probabilistic beliefs as the inferred speaker beliefs.

\begin{small}
\begin{table}
\begin{center}
\caption{Distributions for the Skiing Example with two Bayes nets $s_{\text{dep}}$, $s_{\text{ind}}$,  utterances $\mathcal{U}=\{E\rightarrow S, S, \text{likely } S\}$ and $\alpha=1$. }
\label{tbl:skiing}
\begin{tabular}{lllllllll}\toprule
							& \multicolumn{3}{l}{Literal meaning $\delta_{s \in \sv{u}}$}  											& & & \multicolumn{3}{l}{Speaker $P_{\text{S}}(u\mid s)$} \\\midrule
$u$						& $E\rightarrow S$ 	& $S$ 	& $\text{likely } S$ 	&& $u$ 					&  $E\rightarrow S$ & $S$ & $\text{likely } S$  \\\midrule
$s_{\text{dep}}$ 	& 1 						  	& 0 		& 0							&& $s_{\text{dep}}$ & 1 & 0 & 0  \\
$s_{\text{ind}}$   	& 1 							& 1 		& 1 							&& $s_{\text{ind}}$ 	& $\nicefrac{1}{5}$ & $\nicefrac{2}{5}$ & $\nicefrac{2}{5}$\\ \midrule
\multicolumn{9}{c}{} \\
& \multicolumn{3}{l}{Literal listener $P_{\text{LL}}(s\mid u)$} 	&& 								& \multicolumn{3}{l}{Pragmatic listener $P_{\text{PL}}(s\mid u)$} \\ \midrule
$u$						& $E\rightarrow S$ & $S$ & $\text{likely } S$ && $u$ 						& $E\rightarrow S$ & $S$ & $\text{likely } S$  \\\midrule
$s_{\text{dep}}$ 	& $\nicefrac{1}{2}$ & 0 		& 0 						&& $s_{\text{dep}}$ 	& $\nicefrac{5}{6}$ & 0 & 0\\
$s_{\text{ind}}$   	& $\nicefrac{1}{2}$ & 1 		& 1 						&& $s_{\text{ind}}$ 	& $\nicefrac{1}{6}$ & 1 &1 \\ \bottomrule
\end{tabular}
\end{center}
\end{table}
\end{small}

\begin{figure}
\centering
\subfigure[$s_{wk}$: $S, C$ dependent\label{fig:graph_ski_obs}]{\begin{tikzpicture}
\node[latent, label= {right: $P(s)$}] (S) [] {$S$};
\node[latent,  label={right: $\begin{aligned}
& P(c\mid s)=0.5 \\
& P(c\mid \neg s) = 0
\end{aligned}$}] (C) [below=of S] {$C$};
\edge {S} {C};
\end{tikzpicture}}\qquad\qquad\qquad
\subfigure[Expected value of the listener's degree of belief in the antecedent\label{fig:results_ski_obs}]{\scalebox{0.6}{\input{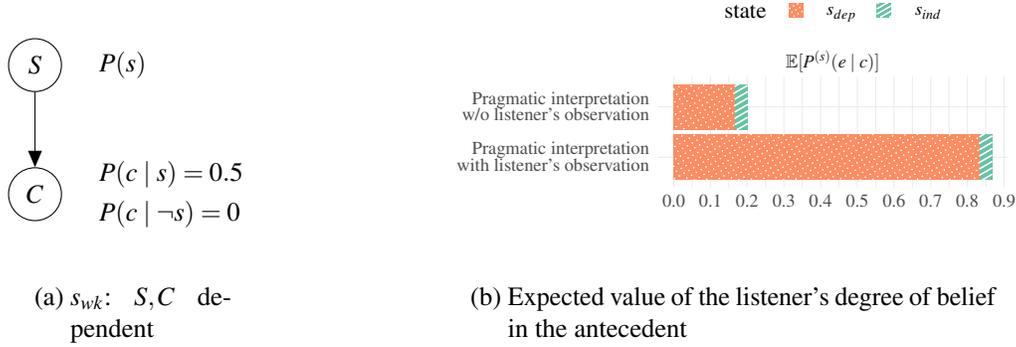}}}
\caption{Bayes net  for the assumed world knowledge in the Skiing Example (left) and results for the the listener's inference about the antecedent (right); $E$: pass exam, $S$: go skiing, $C$: buy skiing clothes.}
\label{fig:ski_observation}
\end{figure}

This takes us to the last step of the explanation, where we look at the effect of the listener's private knowledge that $C$ is the case.
Put differently,  we look at the listener’s inference of the speaker’s beliefs about Sue passing the exam under the assumption that the speaker makes the same observation as the listener.
Similar to \cite{Douven2012b},  we draw on world knowledge about the relation between $C$ and $S$,  where the former is highly unlikely when one does not go on a skiing trip.\footnote{We argue that the relation between $C$ and $S$ is essential for the desired interpretation of the conditional. Imagine that Harry, the listener,  had no idea what skiing is. In this case,  both states should be modeled without the link between $S$ and $C$ making Harry's observation irrelevant with respect to his belief about Sue passing the exam ($E$). Therefore,  his belief in the antecedent would remain unchanged. See \citet{Gunther2018} who also makes use of causal Bayes nets but argues that the intuitive interpretation of the Skiing example is independent of whether or not the relation between variables $S$ and $C$ is modeled.}
This is reflected in the model by setting $P(c\mid \neg s)$ to 0  and $P(c\mid s)$ to 0.5 (see \Cref{fig:graph_ski_obs}) as the context story does not provide any information concerning Sue's shopping behavior.

Based on this (world) knowledge about $S$ and $C$ (Bayes net $s_{wk}$), we can compute the listener's updated belief in the probability that Sue goes on a skiing trip given the listener's observation of Sue buying skiing clothes:
\begin{equation}\label{eq:ski_psc}
P^{s_{wk}}(s\mid c) = \frac{P^{s_{wk}}(c\mid s) \cdot P^{s_{wk}}(s)}{P^{s_{wk}}(c)} = \frac{0.5 \cdot P^{s_{wk}}(s)}{0.5 \cdot P^{s_{wk}}(s) + 0 \cdot P^{s_{wk}}(\neg s)} = 1
\end{equation}
Assuming the updated probability of Sue going on a skiing trip, $P^{s_{wk}}(s\mid c)$ given in Equation \eqref{eq:ski_psc}, the updated probability of Sue passing the exam ($E=e$) for the two modeled states $s_{dep}$ and $s_{ind}$ is given in Equations \eqref{eq:ski_sdep_pe} and \eqref{eq:ski_sind_pe}:
\begin{equation}\label{eq:ski_sdep_pe}
\begin{split}
P^{(s_{dep})}(e\mid c) &= P^{(s_{dep})}(s,e \mid c) + P^{(s_{dep})}(\neg s, e \mid c) \\
						   			&= P^{(s_{dep})}(e\mid s) \cdot P^{(s_{wk})}(s\mid c) + P^{(s_{dep})}(e\mid \neg s) \cdot P^{(s_{wk})}(\neg s\mid c) \\
						   			& = 1 \cdot 1 + 0 \cdot 0 = 1
\end{split}
\end{equation}
\begin{equation}\label{eq:ski_sind_pe}
\begin{split}
P^{(s_{ind})}(e\mid c) 	&= P^{(s_{ind})}(s,e\mid c) + P^{(s_{ind})}(\neg s, e\mid c) \\
							&= P^{(s_{ind})}(e) \cdot P^{(s_{wk})}(s\mid c) + P^{(s_{ind})}(e) \cdot P^{(s_{wk})}(\neg s\mid c)\\
						 	&= 0.2 \cdot 1 + 0.2 \cdot 0 = 0.2 
\end{split}
\end{equation}
These values emphasize the importance of taking into account the causal relations among variables: only for $s_{dep}$ the degree of belief in $E$ is influenced by the listener's observation of $c$.
Given $s_{ind}$, the probability of Sue going on a skiing trip remains the same as without the listener's observation.
The listener's updated belief in the antecedent, given the listener's independent observation of Sue buying skiing clothes, is then equal to the expected value of $P^{(s)}(e\mid c)$ ($s\sim P_{\text{PL}}(s\mid u=E\rightarrow S), s\in [s_{dep}, s_{ind}]$) which is approximately 0.87, as spelled out in Equation \eqref{eq:skiing-update}.
Remember that $P_{\text{PL}(s\mid u=E\rightarrow S)}$ describes the listener's beliefs after the uptake of the conditional,  that is, the listener is approximately $83\%$ confident that the speaker's beliefs correspond to $s_{dep}$ and approximately $17\%$ confident that they correspond to $s_{ind}$. 
\Cref{fig:results_ski_obs} shows the predictions for the listener's posterior beliefs in the antecedent, with and without consideration of the listener's observation. 

\begin{equation}\label{eq:skiing-update}
\begin{split}
E[P^{(s)}(e\mid c)] &= \sum_{s_i \in \{s_{\text{dep}}, s_{sind}\}} P^{(s_i)}(e|c) \cdot P_{\text{PL}}(s_i\mid u = E\rightarrow S) =\\
				 & P^{(s_\text{dep})}(e\mid c) \cdot P_{\text{PL}}(s_{\text{dep}} \mid u=E\rightarrow S) \; + \\
& P^{(s_\text{ind})}(e\mid c) \cdot P_{\text{PL}}(s_{\text{ind}} \mid u=E\rightarrow S) = 1  \cdot \nicefrac{5}{6} + 0.2 \cdot \nicefrac{1}{6} \approx 0.87
\end{split}
\end{equation}

Conceptually,  the expected value may be interpreted as integration of the listener's own observation with the information received from the speaker about the speaker's beliefs about the world that the listener takes over.

\subsection{The Garden Party Example}
\label{sec:garden-party}
Despite its structural similarity to the Skiing Example, for completeness, let us also briefly consider the Garden Party Example where, intuitively,  the listener's degree of belief in the antecedent decreases.
\par
\begingroup
\leftskip2em
\rightskip\leftskip
\noindent
\paragraph{The Garden Party Example \citep{Douven2012b}.}
Betty knows that Kevin, the son of her neighbors, was to take his driving test yesterday. She has no
idea whether or not Kevin is a good driver; she deems it about as likely as not that Kevin passed the
test. Betty notices that her neighbors have started to spade their garden. Then her mother, who is
friends with Kevin’s parents, calls her and tells her the following:
\begin{exe}
\ex \label{itm:garden-party} If Kevin passed the driving test, his parents will throw a garden party.\\
        $\leadsto$ listener belief in antecedent \emph{decreases}
\end{exe}
\par
\endgroup

\begin{figure}
\subfigure[$s_{dep}$: $D, G$ dependent\label{fig:graph_gp_dep}]{
\begin{tikzpicture}
\node[latent, label={right:$\begin{aligned}
& P(g\mid  d)=1 \\
& P(g\mid  \neg d) = 0.5
\end{aligned}$}] (G) [] {$G$};
\node[latent, label={right:$P(d)=0.5$}] (D) [above=of G] {$D$};
\edge {D} {G};
\end{tikzpicture}
}
\subfigure[$s_{ind}$: $D, G$ independent\label{fig:graph_gp_ind}]{
\begin{tikzpicture}
\node[latent, label={right: $P(g) = 0.95$}] (G) [] {$G$};
\node[latent, label={right: $P(d)=0.5$}] (D) [above=of G] {$D$};
\end{tikzpicture}
}%
 \resizebox{0.5\textwidth}{!}{%
\subfigure[Expected value of the listener's degree of belief in the antecedent \label{fig:results_gp_pragmatic}]{\scalebox{0.6}{\input{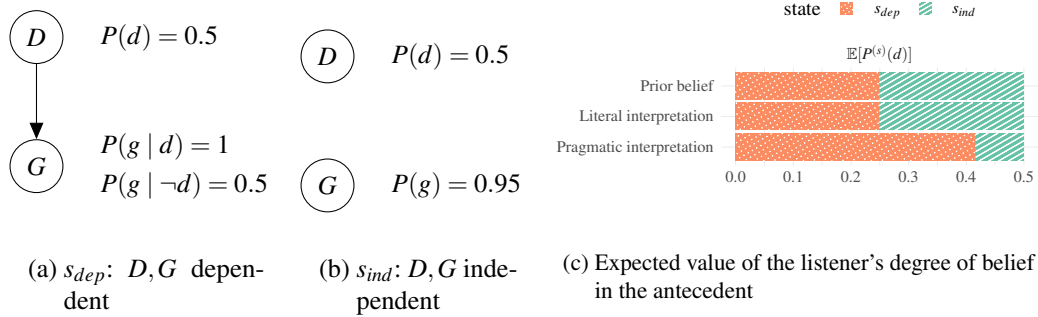}}
}
}%
\caption{Bayes nets and results for Garden Party Example with $\mathcal{U}=\{G, D\rightarrow G, \text{likely } G \}$ and $\alpha=3$, $D$: pass driving exam, $G$: throw garden party.  Both states, $s_{dep}$ and $s_{ind}$, are assigned equal prior probability.}
\label{fig:gp_pragmatic}
\end{figure}

\noindent
The difference between the Bayes nets shown in \Cref{fig:gp_pragmatic} to the Bayes nets used in the Skiing example only lies in the intuitive instantiation of the prior probabilities. 
The results for the interpretation of the conditional in the Garden Party Example (\Cref{fig:results_gp_pragmatic}) also falls in with the results in the Skiing Example: only under a pragmatic interpretation, the listener increases her belief in the Bayes net where both variables are connected ($s_{dep}$), but the listener's belief in the probability of the antecedent remains as prior to the speaker's utterance. 

\begin{figure}
\centering
\subfigure[$s_{wk}$: $S, G$ dependent\label{fig:graph_gp_obs}]{\begin{tikzpicture}
\node[latent, label= {right: $P(g)$}] (G) [] {$G$};
\node[latent,  label={right: $\begin{aligned}
& P(s\mid g)=0 \\
& P(s\mid \neg g) = 0.5
\end{aligned}$}] (S) [below=of G] {$S$};
\edge {G} {S};
\end{tikzpicture}}\qquad\qquad\qquad
\subfigure[Expected value of the listener's degree of belief in the antecedent\label{fig:results_gp_obs}]{\scalebox{0.6}{\input{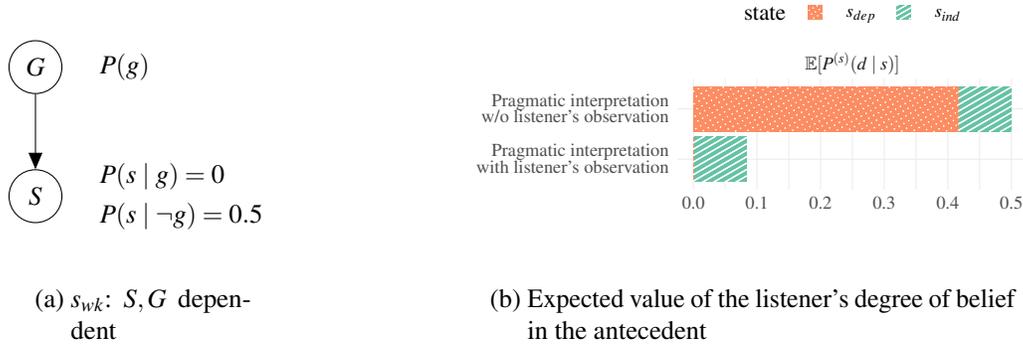}}}
\caption{Bayes net  for the assumed world knowledge in the Garden Party Example (left) and results for the the listener's inference about the antecedent (right); $D$: pass driving exam, $G$: throw garden party, $S$: spade garden.}
\label{fig:gp_observation}
\end{figure}

Again, to consider the listener's own beliefs, we assume that the listener simply takes over the speaker's beliefs communicated by the utterance of the conditional.
The world knowledge that we draw on in this example, shown in \Cref{fig:graph_gp_obs}, concerns  the fact that throwing a garden party is incompatible with spading the garden ($S=s$). 
Therefore,  $P(s\mid g)$ is set to 0 and, due to the lack of more concrete information, $P(s\mid \neg g)$ is set to 0.5.
Contrary to the Skiing Example, the listener's observation in this example is  not evidence, but counterevidence for the antecedent: Betty observes her neighbors spading the garden, thus, the probability for a garden party decreases,  $P^{(s_{wk})}(g\mid s) = 0$.
While in $s_{ind}$, the degree of belief in $D$ is not influenced by the truth or falsity of $G$,  $P^{(s_{dep})}(d\mid \neg g)=0$ (as opposed to $P^{(s_{dep})}(d\mid g)=\nicefrac{2}{3}$). 
The combination of world knowledge about the connection between $S$ and $G$, the speaker's utterance ($D\rightarrow G$),  communicating a likely connection between $D$ and $G$,  and the listener's observation related to $G$ therefore make Betty decrease her belief in the antecedent (Kevin passing the driving test), shown in \Cref{fig:results_gp_obs}.

\subsection{The Sundowners Example}
\label{sec:sundowners}
The Sundowners Example is a case where, intuitively, the listener's degree of belief in the antecedent does not change much, if at all.

\par
\begingroup
\leftskip2em
\rightskip\leftskip
\noindent
\paragraph{The Sundowners Example \citep{Douven2011a}.}
Sarah and Marian have arranged to go for sundowners at the
Westcliff hotel tomorrow. Sarah feels there is some chance that
it will rain, but thinks they can always enjoy the view from
inside. To make sure, Marian consults the staff at the Westcliff
hotel and finds out that in the event of rain, the inside area will
be occupied by a wedding party. So she tells Sarah:
\begin{exe}
\ex \label{itm:sundowners} If it rains tomorrow, we cannot have sundowners at the Westcliff.\\
        $\leadsto$ listener belief in antecedent remains \emph{unchanged}
\end{exe}
\par
\endgroup

\begin{figure}
\subfigure[$s_{dep}$: $R, S$ dependent ($R\rightarrow S$)\label{fig:graph_sun_dep}]{
\begin{tikzpicture}
\node[latent, label={right:$\begin{aligned}
& P(s\mid  r) = 0\\
& P(s\mid  \neg r) = 1
\end{aligned}$}] (S) [] {$S$};
\node[latent, label={right:$P(r)=0.5$}] (R) [above=of S] {$R$};
\edge {R} {S};
\end{tikzpicture}
}
\subfigure[$R, S$ independent  ($R\indep S$) with $P^{(s_{ind\_low})}(s)=0.05$ and $P^{(s_{ind\_high})}(s)=0.95$\label{fig:graph_sun_ind}]{
\begin{tikzpicture}
\node[latent, label={right: $P(r) = 0.5$}] (R) [] {$R$};
\node[latent, label={right: $P(s) \in [0.05,  0.95]$}] (S) [below=of S] {$S$};
\end{tikzpicture}
}%
\subfigure[Examplary Bayes net where $R$ and $S$ are related via a mediating variable ($W$) \label{fig:graph_sun_obs}]{\begin{tikzpicture}
\node[latent, node distance=1.3cm, label= {below: $P(r)=0.5$}] (R) [] {$R$};
\node[latent, node distance=1.3cm,label={below: $\begin{aligned}
& P(w\mid r)=1 \\
& P(w\mid \neg r) = 0
\end{aligned}$}] (W) [right=of R] {$W$};
\node[latent,  node distance=1.3cm,label={below: $\begin{aligned}
& P(s\mid w)=0 \\
& P(s\mid \neg w) = 1
\end{aligned}$}] (S) [right=of W] {$S$};
\edge {R} {W};
\edge {W} {S};
\end{tikzpicture}}\\
\begin{center}
\subfigure[Expected value of the listener's degree of belief in the causal relation between $R$ and $S$ (left), the probability of the antecedent (middle) and the joint event of antecedent and consequent (right), given the Bayes nets from \Cref{fig:graph_sun_dep} and \Cref{fig:graph_sun_ind}.\label{fig:results_sun_pragmatic}]{\scalebox{0.9}{\input{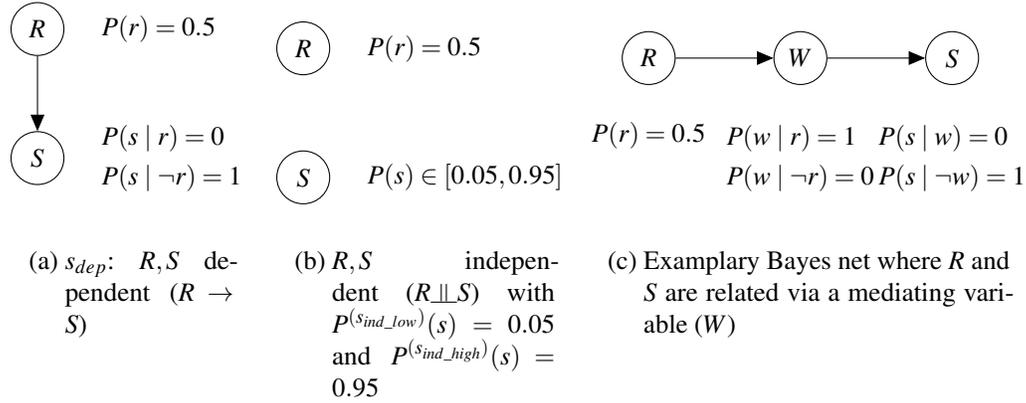}
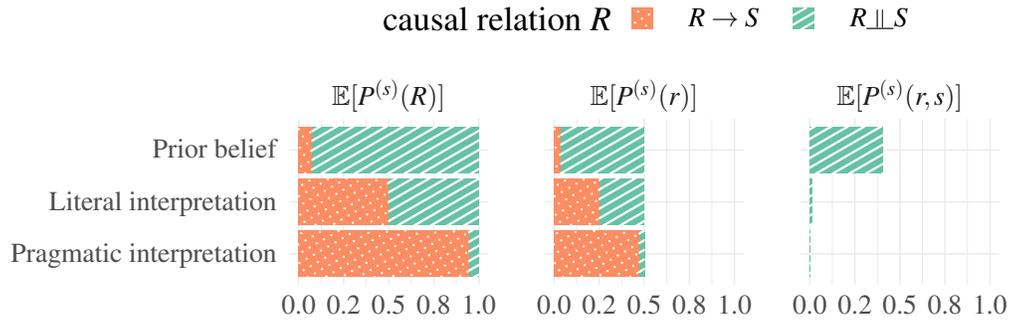}}
\end{center}
\caption{Bayes nets and results for Sundowners Example with $\mathcal{U}=\{R\rightarrow \neg S, \text{likely} S, \text{likely} \neg S, S, \neg S \}$, $\alpha=3$ and a prior probability of 0.85 for Bayes net $s_{ind\_high}$, which is \emph{a priori} most likely according to the context story; $s_{ind\_low}$ and $s_{dep}$ are both assigned a prior probability of 0.075.}
\label{fig:sun_pragmatic}
\end{figure}

\noindent
The intuition behind the formal treatment of the Sundowners Example, given below, is as follows.
Even though the speaker's utterance of the conditional does not seem infelicitous, it seems less natural than the conditionals in the previous two examples.
This oddness is reflected in a response from the listener that is not far to seek: why should rain prevent them from having sundowners?
Put differently, we expect the listener, who has a strong prior belief in the independence of $R$ and $S$, to be somewhat surprised by the speaker's utterance.
The most rational explanation for the speaker's utterance choice of the conditional $R\rightarrow \neg S$ is to give up the assumption of independence \dash at least if the integrity of the speaker is taken for granted.
The listener's surprise may be resolved by accommodating a third, latent variable which is neither observed nor part of the speaker's utterance and acts as mediator between `rain' and `having sundowners'. 
This would justify the speaker's choice to utter the conditional $R\rightarrow \neg S$ which strongly suggests that (the speaker knows that) there is a (causal) connection between `rain' and `having sundowners'.
In other words, we propose that, as opposed to the previous examples, the speaker's utterance in the Sundowners Example forces the listener to accommodate a mitigating variable, which forms a ``causal bridge'' between antecedent and consequent, so as to rationalize the speaker's utterance.
As nothing further is known about this newly introduced ``causal bridge'' or any other relevant variable, the result is that the listener's beliefs about the variable `rain' remains largely the same.

A formally explicit treatment of these ideas makes use of the Bayes nets shown in \Cref{fig:graph_sun_dep}--\ref{fig:graph_sun_obs}.
As for the Skiing and the Garden Party Example, the speaker's utterance of the conditional provokes a shift from the Bayes net where $R$ and $S$ are independent (\Cref{fig:graph_sun_ind}) towards the Bayes net where they are causally related (\Cref{fig:graph_sun_dep}). 
Yet,  as can be seen in the leftmost panel of \Cref{fig:results_sun_pragmatic}, in this example, the listener has a strong  prior belief in the former which explains the listener's surprise resulting from the speaker's utterance choice.
Assuming independence of $R$ and $S$, the listener would rather expect the speaker to choose a more informative utterance assertable than the conditional (e.g. ,$\neg S$).

The RSA model of pragmatic language production entails a notion of \emph{surprising utterances}.
Given prior beliefs about states $P_{\text{prior}}(s)$ and the speaker's assumed production probabilities $P_{\text{S}}(u \mid s)$, the pragmatic listener expects utterance $u$ with probability:
\begin{align*}
  P_{\text{PL}}(u) = \sum_{s} P_{\text{prior}}(s) \ P_{\text{S}}(u \mid s)
\end{align*}
If the pragmatic listener only considers the three states represented in Figures~\ref{fig:graph_sun_dep} and \ref{fig:graph_sun_ind}, with a prior probability of 0.85 for $s_{ind\_high}$  and a prior probability of respectively 0.075 for $s_{ind\_low}$ and $s_{dep}$,  the conditional $R \rightarrow \neg S$ is highly surprising in the sense that its expected occurrence probability is very low ($\approx 0.08$ when $\alpha = 3$).\footnote{$P_{\text{S}}(u = R\rightarrow \neg S \mid s_{ind\_low}) \approx 0.06, P_{\text{S}}(u = R\rightarrow \neg S \mid s_{dep}) = 1$}
This notion of listener surprise in the light of an unexpected utterances helps explain the intuition that an utterance of $R \rightarrow \neg S$ may trigger the desire to go look for an additional explanation which may rationalize the observed utterance.

One possibility of how the listener may rationalize a surprising utterance through a mitigating variable is shown in \Cref{fig:graph_sun_obs} where $W$ denotes the event of `a wedding taking place inside the hotel' that represents any event which may possibly prevent the interlocutors from having sundowners at the hotel.
Rather than denying the speaker's integrity due to the speaker's somewhat puzzling utterance,  pragmatic reasoning, eminently causal-pragmatic reasoning,  therefore allows the listener to infer a previously unexpected relation between $R$ and $S$,  that is able to explain  the speaker's utterance.
Nevertheless,  the conditional still seems to be  incomplete as one could expect the speaker to be more informative about the reasons why the event of rain may prevent them from having sundowners.

Contrary to the Skiing and the Garden Party Example, in the Sundowners Example,  the listener does not intuitively update her belief in the probability of the antecedent. 
\Cref{fig:results_sun_pragmatic} (middle) shows that the listener's predicted degree of belief in the antecedent remains at its prior value of 0.5 after processing the speaker's utterance, assuming a literal or a pragmatic interpretation.
The crucial difference here is that the listener does not make any further observations as in the previous examples. 
Yet, the speaker's utterance of the conditional has a strong effect on the the listener's beliefs concerning the joint event of the antecedent and the consequent, shown in the right panel of  \Cref{fig:results_sun_pragmatic}.
Contrary to her prior belief that $P(r,s)$ is almost a matter of chance, the listener judges it almost impossible that the two events both hold at the same time upon receiving the conditional information $R\rightarrow \neg S$: in $s_{dep}$, $R$ and $S$ are mutually exclusive, thus $P(r,s)=0$ and in $s_{ind\_low}$ where $R\rightarrow \neg S$ is assertable, $P(r,s)$ is close to 0.

In sum, the model is therefore able to account \dash by a single mechanism \dash for different interpretations with respect to the listener's degree of belief in the antecedent, namely by the interplay of pragmatic reasoning and an adequate, explicit representation of the interlocutors' prior probabilistic and in particular, \emph{causal} beliefs.

\section{On missing links \& biscuits}
\label{sec:missing-links-bcs}

Previous sections showed how the model of communication with conditionals presented here predicts that speakers use a conditional predominantly when there is a causal/inferential relation between antecedent and consequent and that therefore listeners will interpret conditionals accordingly.
This brings up the obvious question as to how the advocated approach positions itself with regard to cases, prominently discussed in the literature, in which antecedent and consequent are clearly \emph{not} causally or inferentially related.
The absence of a relation between antecedent and consequent can either result in infelicity, as is the case in what we here call \emph{missing-link conditionals} \citep[][]{Douven2017}, or trigger an altogether different kind of interpretation, as in what we here address as \emph{biscuit conditionals} \citep[e.g.,][]{Austin1956, geisLycanNonconditional93}.
This section deals with each case in turn.

\subsection{Missing-link conditionals}
\label{sec:miss-links}

Missing-link conditionals have been used in support of the inferentialist position that the requirement of a causal/inferential connection between antecendent and consequent is a necessary condition for assertability of a conditional, arguably situated in the semantics of conditionals because it is unclear how else a pragmatic account could explain the \emph{in}felicity of missing-link conditionals \citep[e.g.][]{Douven2008, Douven2017, Krzy2014, Skovgaard-Olsen2016a}.
This position is concretely exemplified by the contrast pair in (\ref{ex:douven-ml}), given by \citet{Douven2008}.

\begin{exe}
\ex \label{ex:douven-ml} There will be at least one heads in the first 1,000,000 tosses of this fair coin ($h_{10^{6}}$) \emph{if}
\begin{xlist}
\ex{\label{ex:douven-ml-a} there is a heads in the first ten tosses. ($h_{10}$)}
\ex[* ]{\label{ex:douven-ml-b} Chelsea wins the Champions League. ($c$)}
\end{xlist}
\end{exe}

\noindent What is remarkable about the contrast between the two sentences in (\ref{ex:douven-ml}) is this: on the one hand, both (\ref{ex:douven-ml-a}) and (\ref{ex:douven-ml-b}) arguably pass the minimal necessary requirement for assertability, namely that the probability of the consequent given the antecedent is high \dash in fact the difference between the relevant conditional probabilities is minute ($P(h_{10^6}\mid h_{10})=1$, $P(h_{10^6}\mid c)\approx 1$); on the other hand, while (\ref{ex:douven-ml-a}) is intuitively acceptable, for instance in a context where the speaker wants to highlight the entailment relation to a listener who might otherwise not attend to it sufficiently, (\ref{ex:douven-ml-b}) rather clearly is odd.

To account for the infelicity of \pref{ex:douven-ml-b} and assertability of \pref{ex:douven-ml-a}, inferentialism stipulates that a conceivable inferential relation between antecedent and consequent is a necessary requirement for assertability, howsoever the inferential link may exactly be defined.
It is not limited to deductive inferences, as modern inferentialism allows less strict inferential relations such as induction or abduction \citep[e.g.][]{Krzy2014, Douven2017}.\footnote{
   Several empirical studies \citep[e.g.,][]{Krzyzanowska2013,Skovgaard-Olsen2016, Skovgaard-Olsen2017} have shown that the link between antecedent and consequent has an influence on whether conditionals are accepted.
   Participants in a study from \citet{Douven2010} for instance interpreted conditionals differently depending on the type of the link (deductive, inductive or abductive).
   Yet, there is also empirical evidence supporting the view that the link between antecedent and consequent is attributable to discourse pragmatics rather than conventional semantics \citep[e.g. see][]{Cruz,lassiter_decomposing_nodate}.}
 Inferentialists have criticized pragmatic explanations of the perceived infelicity of missing-link conditionals for remaining too vague about how the pragmatic processes may concretely look like (e.g., \citealt{Douven2017}).
 
We argue here that \pref{ex:douven-ml-b} is similar in kind to the Sundowners Example,  yet more extreme, therefore leading to perceived infelicity.
The Sundowners Example from the previous section showcases that there are contexts in which the utterance of a conditional is not questioned per se, but will neither be accepted without further ado.
As there is no \emph{obviously} conceivable connection between `rain' and `not having sundowners at the hotel',  it seems quite natural for the listener to reply with a question asking for precisely this connection.
We consider infelicitous missing-link conditionals like \pref{ex:douven-ml-b} to be similar in kind but more extreme cases of the same variety: in \pref{ex:douven-ml-b} the speaker provides so little information that the listener does not have enough cues to make sense of the conditional utterance from world knowledge alone.
In other words, we maintain that the infelicity of \pref{ex:douven-ml-b} is a result of a failure of the listener to see a connection between antecedent and consequent that could rationalize the speaker's utterance choice.

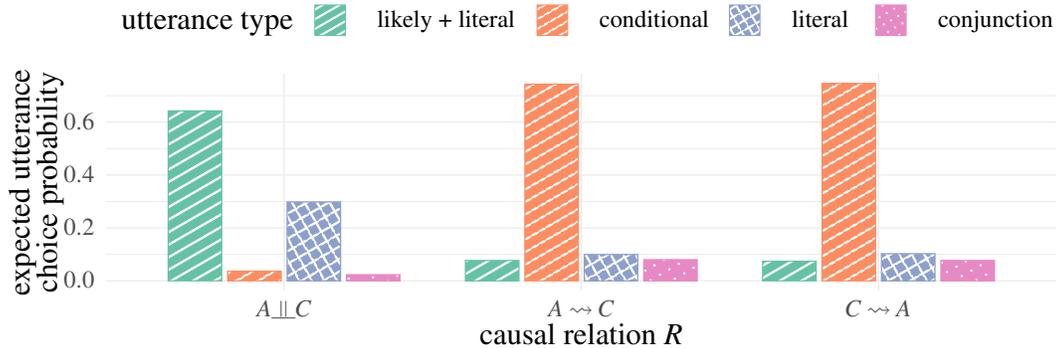
\begin{figure}
\scalebox{0.8}{\input{figs/speaker-evs.tex}}
\caption{Speaker's expected utterance choice probabilities computed on speaker's predictions, where $\alpha=3$, for a set of 10,000 states sampled from the default context prior, grouped by utterance type and causal relation.}
\label{fig:speaker-evs}
\end{figure}

To see how this is predicted by our model, we take the perspective of a listener, who tries to interpret a missing-link conditional and knows, from common sense world knowledge, that the antecedent and the consequent are two independent events.
Given the assumption of independence, our hyperrational speaker would never choose any of the conditional utterances (see \Cref{fig:speaker-uncertainty-best}),  and even for a speaker with a less optimal rationality parameter ($\alpha=3$), we observe that conditional utterances only have an expected utterance choice probability $<5\%$ when considering the speaker's predictions for states with $r=A\indep C$ from a set of 10,000 samples from the default context prior, as shown in \Cref{fig:speaker-evs}.

The reason for this low probability of choosing a conditional when $A$ and $C$ are assumed to be independent, is that very likely there will be an assertable,  more informative utterance available; for the independent Bayes net, the joint probability tables where $A\rightarrow C$ is assertable (i.e., $P^{(s)}(c\mid a) \geqslant \theta$) also satisfy the assertability condition for a literal or a conjunction.
Given the background knowledge of independence of the antecedent and the consequent and given that in these cases the speaker hardly ever chooses conditionals, the listener will naturally be highly surprised by the speaker's utterance of a conditional.
It is this notion of surprise on the part of the listener,  based on the speaker's expected utterance choice probabilities,  that allows us to infer the infelicity of missing-link conditionals: due to the listener's surprise, she will, arguably,  want to look for another way of rationalizing the assertion.
If no such option is forthcoming, the utterance feels infelicitous.
Missing-link conditionals are therefore \emph{not} accepted by the listener \dash contrary to the conditional in the Sundowners Example where the listener's initial surprise can be repaired at the content-level since the listener is, after all, able to accommodate a connection between antecedent and consequent.

The infelicity of missing-link conditionals as we explain it here, may further be couched within the resource-rational, sampling-based approach to explain causal reasoning.\footnote{Thanks to an anonymous reviewer for highlighting this interpretation.}
For example, \citet{dasgupta_where_2017} show that cognitive biases like \emph{subadditivity} or \emph{superadditivity} can be explained by a comparison of people's search for a plausible explanation of observed data with sampling algorithms, in particular MCMC sampling.
They argue that due to cognitive load and time constraints, the number of samples that people can draw is restricted,  such that the optimal answer is not found, giving rise to common cognitive biases.
As regards to missing-link conditionals,  
the listener's search for plausible Bayes nets with sufficient explanatory value to rationalize the speaker's utterance, seems to fail (e.g.,  in Example \ref{ex:douven-ml-b}).
Yet, when it comes to  the interpretation of Biscuit conditionals (see \Cref{sec:biscuits} below) or to conditionals like in the Sundowners example, it seems reasonable for a listener to be able to quickly come up with more or less satisfying candidates.

\subsection{Biscuit conditionals}
\label{sec:biscuits}

Similar to missing-link conditionals, biscuit conditionals (BCs) are commonly considered a special kind of conditional as both lack the probably most characteristic feature of conditionals,  the relation between antecedent and consequent.
Unlike missing-link conditionals,  biscuit conditionals are however felicitous, indeed quite common; see \pref{ex:bc} for the classical example from \citet{Austin1956}.

\begin{exe}
 \ex \label{ex:bc} If you're hungry, there are biscuits on the sideboard.
\end{exe}

\noindent
While Inferentialists have mostly excluded BCs from their analysis of conditionals altogether, we argue that the felicity of BCs can indeed be explained as a pragmatic phenomenon.\footnote{
   Recently, \citet{VanRooij2020, Rooij2021} proposed a generalization of their account for the assertability of conditionals which is able to account for BCs as well.}
 The model presented here helps explain why listeners, who will likely assume \textit{a priori} that the consequent and the antecedent in example (\ref{ex:bc}) are independent, would be surprised by an utterance of (\ref{ex:bc}) if it were to be interpreted like a normal conditional.
 It is plausible that for some conditionals, like BCs, the listener will see a repair strategy which enables a rationalization of the conditional after all.
 However, this rationalization does not take place at the content-level by an attempt to find a relation between antecedent and consequent as (unsuccessfully) in missing-link conditionals or (successfully) in the Sundowners Example, but at a different level, for instance at the level of speech-acts.
 Concretely, this idea could be integrated into our model by making the speaker's utterance choice additionally dependent on communicative goals, for example, whether the speaker wants to perform a speech-act (e.g.,  offer the listener some cookies),  or just wants to make a plain assertion (e.g., inform the listener about the existence of biscuits on the sideboard).

 Admittedly this sketch is just a first step towards a satisfying pragmatic account of BCs,  as there is certainly more needed to fully explain their use and interpretation.
 We also do not, with emphasis, suggest that this kind of ``surprise-repair'' interpretation is actively entertained during each reception of a BC:
 BCs arguably provide sufficient secondary cues for the listener to trigger a BC-like interpretation, including intonation or, in some languages, word order (e.g., in German, where BCs can occur with verb-third (V3) verb order and not with the usual verb-second (V2) as seen in simple indicative conditionals).
 
\subsection{Conditionals for the communication of \emph{in}dependence}
\label{sec:conditionals-independence}

We assumed that, first and foremost, the speaker's aim is to communicate (probabilistic) beliefs about certain events and saw that, based on this assumption, the speaker should only choose to utter a conditional if no other, more informative utterance is assertable.
However,  in some circumstances, conditionals may be used in order to put emphasis on the fact that the consequent is \emph{not} dependent on the antecedent.
Consider the example given in (\ref{itm:if-ind}):

\begin{exe}
	\ex \label{itm:if-ind} If you study, you will pass and if you don't study, you will pass, so don't worry, you will pass!
\end{exe}

Along the lines that we have been arguing,  a conditional alike should be avoided since evidently the speaker could say the same thing with a more informative, shorter alternative, e.g., ``\emph{You will pass (no matter what)!}''.
Notwithstanding, the conditional in (\ref{itm:if-ind}) seems a natural utterance to say, in particular when the speaker wants to emphasize the relation, here the \emph{independence} between studying and passing an exam. 
If we extended the set of alternative utterances available to our speaker by including this kind of combined conditionals ($A\rightarrow C \wedge \neg A\rightarrow C$), the listener would infer that the consequent holds true independent of the antecedent. 
We saw previously that our listener infers from the utterance of a conditional, that the speaker is likely uncertain about the truth or falsity of the antecedent.
For simplicity, let us assume the speaker refers to a state $s$ where $P^{(s)}(a)=P^{(s)}(\neg a)=0.5$.
Since the speaker uttered $A\rightarrow C$,  $P^{(s)}(c\mid a)\geqslant \theta$ must hold, and additionally, due to the speaker's utterance of the second conditional, $\neg A \rightarrow C$, $P^{(s)}(c\mid \neg a) \geqslant \theta$.
Then, $P^{(s)}(c) = 0.5 \cdot P^{(s)}(c\mid a) + 0.5 \cdot  P^{(s)}(c\mid \neg a)= 0.5\cdot (P^{(s)}(c\mid a) + P^{(s)}(c\mid \neg a))  \geqslant \theta $. 
Therefore, in this state, it is also possible to assert the consequent straightaway.
A speaker whose aim it is to communicate her uncertain beliefs should therefore prefer the simpler utterance, assuming higher cost for combined utterances like $A\rightarrow C \wedge \neg A \rightarrow C$. 
However, if the speaker's aim was two-minded, including the communication of her uncertain beliefs as well as highlighting the independence between the variables at hand, the combined conditional should become a likely utterance choice for the speaker:  for almost all states in which both conditionals are assertable,   $R=A\indep C$ whereas $C$ is also assertable in many states where $R=$ \ac{} or $R=$ \ca{}.

\section{Conclusion}
\label{sec:conclusion}

We have taken a probabilistic approach to pragmatic reasoning about conditionals here.
Our approach is a conservative one: we combined two independently motivated, well-established formalizations, namely the Rational-Speech-Act model from probabilistic pragmatics and causal Bayes nets, in order to predict a listener's posterior probabilistic beliefs about relevant events and how they are causally related.
Our contribution is a formal, computational model of the use and interpretation of conditionals that may be a starting point for other models in this vein to gain further insights on how rich prior beliefs are updated in the light of conditionals.

We showed that our model vindicates a number of different pragmatic inferences observed in communication with conditionals.
These include inferences such as the dependency between antecedent and consequent, the infelicity of missing-link conditionals or the tendency to interpret conditionals as biconditionals as in conditional perfection readings.
When the model is supplied with particular prior beliefs representing concrete utterance contexts, it also makes predictions corresponding to what is considered their intuitive interpretation; we showed that the model helps to explain the listener's varying inferences with respect to the probability of the antecedent in three concrete utterance contexts from \citet{Douven2012b}.

We explored the interpretation of conditionals under the assumption that conditionals are used to communicate stochastic information about co-occurrence,  while information about the underlying causal structure is communicated implicitly in our model.
In some circumstances, the speaker's communicative goal might however go further than this and include the communication about causal information as well, bringing into play explicit causal language. 
An interesting extension of the model may investigate the relationship between conditionals, causality and causal language further by considering different alternative utterances, including explicit causal expressions that we did not address here.

The flexibility of the model we presented moreover allows future extensions to systematically investigate the factors influencing the interpretation of special types of conditionals such as biscuit conditionals.
Other salient extensions of the present approach include applications to subjunctive conditionals, nested conditionals or conditionals with other logically complex antecedents or consequents.
Further, the quantitative predictions of the model offer the possibility to  test and improve it by a comparison with empirical data.

\clearpage
\bibliography{references.bib}

\begin{addresses}
  \begin{address}
    Britta Grusdt \\
    Wachsbleiche 27 \\
    49090 Osnabr\"uck, Germany \\
    \email{britta.grusdt@uni-osnabrueck.de}
  \end{address}
  \begin{address}
    Daniel Lassiter \\
    Dugald Stewart Building\\
    3 Charles Street, Edinburgh\\
    EH8 9AD, United Kingdom \\
    \email{danlassiter@stanford.edu}
  \end{address}
    \begin{address}
    Michael Franke \\
    Geschwister-Scholl-Platz \\
   72074 T\"ubingen, Germany \\
    \email{michael.franke@uni-tuebingen.de}
  \end{address}
\end{addresses}

\end{document}

%% file: figs/speaker-evs.tex
\begin{tikzpicture}[x=1pt,y=1pt]
\definecolor{fillColor}{RGB}{255,255,255}
\path[use as bounding box,fill=fillColor,fill opacity=0.00] (0,0) rectangle (505.89,180.67);
\begin{scope}
\path[clip] ( 50.22, 33.48) rectangle (499.89,136.22);
\definecolor{drawColor}{gray}{0.92}

\path[draw=drawColor,line width= 0.3pt,line join=round] ( 50.22, 50.66) --
	(499.89, 50.66);

\path[draw=drawColor,line width= 0.3pt,line join=round] ( 50.22, 75.69) --
	(499.89, 75.69);

\path[draw=drawColor,line width= 0.3pt,line join=round] ( 50.22,100.72) --
	(499.89,100.72);

\path[draw=drawColor,line width= 0.3pt,line join=round] ( 50.22,125.75) --
	(499.89,125.75);

\path[draw=drawColor,line width= 0.6pt,line join=round] ( 50.22, 38.15) --
	(499.89, 38.15);

\path[draw=drawColor,line width= 0.6pt,line join=round] ( 50.22, 63.18) --
	(499.89, 63.18);

\path[draw=drawColor,line width= 0.6pt,line join=round] ( 50.22, 88.21) --
	(499.89, 88.21);

\path[draw=drawColor,line width= 0.6pt,line join=round] ( 50.22,113.24) --
	(499.89,113.24);

\path[draw=drawColor,line width= 0.6pt,line join=round] (134.53, 33.48) --
	(134.53,136.22);

\path[draw=drawColor,line width= 0.6pt,line join=round] (275.06, 33.48) --
	(275.06,136.22);

\path[draw=drawColor,line width= 0.6pt,line join=round] (415.58, 33.48) --
	(415.58,136.22);
\definecolor{fillColor}{RGB}{102,194,165}

\path[fill=fillColor] ( 79.73, 38.15) rectangle (105.02,118.55);
\definecolor{fillColor}{RGB}{252,141,98}

\path[fill=fillColor] (107.83, 38.15) rectangle (133.13, 42.70);
\definecolor{fillColor}{RGB}{141,160,203}

\path[fill=fillColor] (135.94, 38.15) rectangle (161.23, 75.47);
\definecolor{fillColor}{RGB}{231,138,195}

\path[fill=fillColor] (164.04, 38.15) rectangle (189.34, 41.03);
\definecolor{fillColor}{RGB}{102,194,165}

\path[fill=fillColor] (220.25, 38.15) rectangle (245.55, 47.77);
\definecolor{fillColor}{RGB}{252,141,98}

\path[fill=fillColor] (248.36, 38.15) rectangle (273.65,131.20);
\definecolor{fillColor}{RGB}{141,160,203}

\path[fill=fillColor] (276.46, 38.15) rectangle (301.75, 50.61);
\definecolor{fillColor}{RGB}{231,138,195}

\path[fill=fillColor] (304.57, 38.15) rectangle (329.86, 48.16);
\definecolor{fillColor}{RGB}{102,194,165}

\path[fill=fillColor] (360.77, 38.15) rectangle (386.07, 47.42);
\definecolor{fillColor}{RGB}{252,141,98}

\path[fill=fillColor] (388.88, 38.15) rectangle (414.17,131.55);
\definecolor{fillColor}{RGB}{141,160,203}

\path[fill=fillColor] (416.98, 38.15) rectangle (442.28, 50.98);
\definecolor{fillColor}{RGB}{231,138,195}

\path[fill=fillColor] (445.09, 38.15) rectangle (470.38, 47.78);
\definecolor{drawColor}{RGB}{255,255,255}
\definecolor{fillColor}{RGB}{255,255,255}

\path[draw=drawColor,line width= 0.6pt,line join=round,line cap=rect,fill=fillColor] (101.18, 38.15) --
	(105.02, 40.37) --
	(105.02, 39.77) --
	(102.21, 38.15) --
	(101.18, 38.15) --
	cycle;

\path[draw=drawColor,line width= 0.6pt,line join=round,line cap=rect,fill=fillColor] ( 90.90, 38.15) --
	(105.02, 46.30) --
	(105.02, 45.70) --
	( 91.93, 38.15) --
	( 90.90, 38.15) --
	cycle;

\path[draw=drawColor,line width= 0.6pt,line join=round,line cap=rect,fill=fillColor] ( 80.63, 38.15) --
	(105.02, 52.23) --
	(105.02, 51.64) --
	( 81.66, 38.15) --
	( 80.63, 38.15) --
	cycle;

\path[draw=drawColor,line width= 0.6pt,line join=round,line cap=rect,fill=fillColor] ( 79.73, 43.56) --
	(105.02, 58.16) --
	(105.02, 57.57) --
	( 79.73, 42.96) --
	( 79.73, 43.56) --
	cycle;

\path[draw=drawColor,line width= 0.6pt,line join=round,line cap=rect,fill=fillColor] ( 79.73, 49.49) --
	(105.02, 64.09) --
	(105.02, 63.50) --
	( 79.73, 48.90) --
	( 79.73, 49.49) --
	cycle;

\path[draw=drawColor,line width= 0.6pt,line join=round,line cap=rect,fill=fillColor] ( 79.73, 55.42) --
	(105.02, 70.03) --
	(105.02, 69.43) --
	( 79.73, 54.83) --
	( 79.73, 55.42) --
	cycle;

\path[draw=drawColor,line width= 0.6pt,line join=round,line cap=rect,fill=fillColor] ( 79.73, 61.35) --
	(105.02, 75.96) --
	(105.02, 75.36) --
	( 79.73, 60.76) --
	( 79.73, 61.35) --
	cycle;

\path[draw=drawColor,line width= 0.6pt,line join=round,line cap=rect,fill=fillColor] ( 79.73, 67.29) --
	(105.02, 81.89) --
	(105.02, 81.30) --
	( 79.73, 66.69) --
	( 79.73, 67.29) --
	cycle;

\path[draw=drawColor,line width= 0.6pt,line join=round,line cap=rect,fill=fillColor] ( 79.73, 73.22) --
	(105.02, 87.82) --
	(105.02, 87.23) --
	( 79.73, 72.63) --
	( 79.73, 73.22) --
	cycle;

\path[draw=drawColor,line width= 0.6pt,line join=round,line cap=rect,fill=fillColor] ( 79.73, 79.15) --
	(105.02, 93.75) --
	(105.02, 93.16) --
	( 79.73, 78.56) --
	( 79.73, 79.15) --
	cycle;

\path[draw=drawColor,line width= 0.6pt,line join=round,line cap=rect,fill=fillColor] ( 79.73, 85.08) --
	(105.02, 99.69) --
	(105.02, 99.09) --
	( 79.73, 84.49) --
	( 79.73, 85.08) --
	cycle;

\path[draw=drawColor,line width= 0.6pt,line join=round,line cap=rect,fill=fillColor] ( 79.73, 91.01) --
	(105.02,105.62) --
	(105.02,105.02) --
	( 79.73, 90.42) --
	( 79.73, 91.01) --
	cycle;

\path[draw=drawColor,line width= 0.6pt,line join=round,line cap=rect,fill=fillColor] ( 79.73, 96.95) --
	(105.02,111.55) --
	(105.02,110.96) --
	( 79.73, 96.35) --
	( 79.73, 96.95) --
	cycle;

\path[draw=drawColor,line width= 0.6pt,line join=round,line cap=rect,fill=fillColor] ( 79.73,102.88) --
	(105.02,117.48) --
	(105.02,116.89) --
	( 79.73,102.29) --
	( 79.73,102.88) --
	cycle;

\path[draw=drawColor,line width= 0.6pt,line join=round,line cap=rect,fill=fillColor] ( 79.73,108.81) --
	( 96.59,118.55) --
	( 97.62,118.55) --
	( 79.73,108.22) --
	( 79.73,108.81) --
	cycle;

\path[draw=drawColor,line width= 0.6pt,line join=round,line cap=rect,fill=fillColor] ( 79.73,114.74) --
	( 86.32,118.55) --
	( 87.35,118.55) --
	( 79.73,114.15) --
	( 79.73,114.74) --
	cycle;

\path[draw=drawColor,line width= 0.6pt,dash pattern=on 2pt off 2pt ,line join=round,line cap=rect,fill=fillColor] (132.00, 38.15) --
	(133.13, 38.80) --
	(133.13, 38.20) --
	(133.03, 38.15) --
	(132.00, 38.15) --
	cycle;

\path[draw=drawColor,line width= 0.6pt,dash pattern=on 2pt off 2pt ,line join=round,line cap=rect,fill=fillColor] (121.73, 38.15) --
	(129.61, 42.70) --
	(130.64, 42.70) --
	(122.76, 38.15) --
	(121.73, 38.15) --
	cycle;

\path[draw=drawColor,line width= 0.6pt,dash pattern=on 2pt off 2pt ,line join=round,line cap=rect,fill=fillColor] (111.45, 38.15) --
	(119.34, 42.70) --
	(120.36, 42.70) --
	(112.48, 38.15) --
	(111.45, 38.15) --
	cycle;

\path[draw=drawColor,line width= 0.6pt,dash pattern=on 2pt off 2pt ,line join=round,line cap=rect,fill=fillColor] (107.83, 41.99) --
	(109.06, 42.70) --
	(110.09, 42.70) --
	(107.83, 41.39) --
	(107.83, 41.99) --
	cycle;

\path[draw=drawColor,line width= 0.6pt,dash pattern=on 4pt off 2pt ,line join=round,line cap=rect,fill=fillColor] (152.55, 38.15) --
	(161.23, 43.16) --
	(161.23, 42.56) --
	(153.58, 38.15) --
	(152.55, 38.15) --
	cycle;

\path[draw=drawColor,line width= 0.6pt,dash pattern=on 4pt off 2pt ,line join=round,line cap=rect,fill=fillColor] (142.28, 38.15) --
	(161.23, 49.09) --
	(161.23, 48.50) --
	(143.31, 38.15) --
	(142.28, 38.15) --
	cycle;

\path[draw=drawColor,line width= 0.6pt,dash pattern=on 4pt off 2pt ,line join=round,line cap=rect,fill=fillColor] (135.94, 40.42) --
	(161.23, 55.02) --
	(161.23, 54.43) --
	(135.94, 39.82) --
	(135.94, 40.42) --
	cycle;

\path[draw=drawColor,line width= 0.6pt,dash pattern=on 4pt off 2pt ,line join=round,line cap=rect,fill=fillColor] (135.94, 46.35) --
	(161.23, 60.95) --
	(161.23, 60.36) --
	(135.94, 45.76) --
	(135.94, 46.35) --
	cycle;

\path[draw=drawColor,line width= 0.6pt,dash pattern=on 4pt off 2pt ,line join=round,line cap=rect,fill=fillColor] (135.94, 52.28) --
	(161.23, 66.89) --
	(161.23, 66.29) --
	(135.94, 51.69) --
	(135.94, 52.28) --
	cycle;

\path[draw=drawColor,line width= 0.6pt,dash pattern=on 4pt off 2pt ,line join=round,line cap=rect,fill=fillColor] (135.94, 58.21) --
	(161.23, 72.82) --
	(161.23, 72.22) --
	(135.94, 57.62) --
	(135.94, 58.21) --
	cycle;

\path[draw=drawColor,line width= 0.6pt,dash pattern=on 4pt off 2pt ,line join=round,line cap=rect,fill=fillColor] (135.94, 64.15) --
	(155.54, 75.47) --
	(156.57, 75.47) --
	(135.94, 63.55) --
	(135.94, 64.15) --
	cycle;

\path[draw=drawColor,line width= 0.6pt,dash pattern=on 4pt off 2pt ,line join=round,line cap=rect,fill=fillColor] (135.94, 70.08) --
	(145.27, 75.47) --
	(146.30, 75.47) --
	(135.94, 69.48) --
	(135.94, 70.08) --
	cycle;

\path[draw=drawColor,line width= 0.6pt,dash pattern=on 4pt off 2pt ,line join=round,line cap=rect,fill=fillColor] (136.02, 75.47) --
	(135.94, 75.42) --
	(135.94, 75.47) --
	(136.02, 75.47) --
	cycle;

\path[draw=drawColor,line width= 0.6pt,dash pattern=on 4pt off 2pt ,line join=round,line cap=rect,fill=fillColor] (135.94, 38.63) --
	(136.22, 38.15) --
	(135.94, 38.15) --
	(135.94, 38.63) --
	cycle;

\path[draw=drawColor,line width= 0.6pt,dash pattern=on 4pt off 2pt ,line join=round,line cap=rect,fill=fillColor] (135.94, 48.91) --
	(142.15, 38.15) --
	(141.56, 38.15) --
	(135.94, 47.88) --
	(135.94, 48.91) --
	cycle;

\path[draw=drawColor,line width= 0.6pt,dash pattern=on 4pt off 2pt ,line join=round,line cap=rect,fill=fillColor] (135.94, 59.18) --
	(148.08, 38.15) --
	(147.49, 38.15) --
	(135.94, 58.15) --
	(135.94, 59.18) --
	cycle;

\path[draw=drawColor,line width= 0.6pt,dash pattern=on 4pt off 2pt ,line join=round,line cap=rect,fill=fillColor] (135.94, 69.45) --
	(154.02, 38.15) --
	(153.42, 38.15) --
	(135.94, 68.43) --
	(135.94, 69.45) --
	cycle;

\path[draw=drawColor,line width= 0.6pt,dash pattern=on 4pt off 2pt ,line join=round,line cap=rect,fill=fillColor] (138.40, 75.47) --
	(159.95, 38.15) --
	(159.35, 38.15) --
	(137.81, 75.47) --
	(138.40, 75.47) --
	cycle;

\path[draw=drawColor,line width= 0.6pt,dash pattern=on 4pt off 2pt ,line join=round,line cap=rect,fill=fillColor] (144.33, 75.47) --
	(161.23, 46.19) --
	(161.23, 45.17) --
	(143.74, 75.47) --
	(144.33, 75.47) --
	cycle;

\path[draw=drawColor,line width= 0.6pt,dash pattern=on 4pt off 2pt ,line join=round,line cap=rect,fill=fillColor] (150.26, 75.47) --
	(161.23, 56.47) --
	(161.23, 55.44) --
	(149.67, 75.47) --
	(150.26, 75.47) --
	cycle;

\path[draw=drawColor,line width= 0.6pt,dash pattern=on 4pt off 2pt ,line join=round,line cap=rect,fill=fillColor] (156.20, 75.47) --
	(161.23, 66.74) --
	(161.23, 65.71) --
	(155.60, 75.47) --
	(156.20, 75.47) --
	cycle;

\path[draw=drawColor,line width= 0.6pt,dash pattern=on 4pt off 4pt ,line join=round,line cap=round,fill=fillColor] (165.02, 39.12) circle (  0.26);

\path[draw=drawColor,line width= 0.6pt,dash pattern=on 4pt off 4pt ,line join=round,line cap=round,fill=fillColor] (176.49, 39.80) circle (  0.26);

\path[draw=drawColor,line width= 0.6pt,dash pattern=on 4pt off 4pt ,line join=round,line cap=round,fill=fillColor] (187.96, 40.49) circle (  0.26);

\path[draw=drawColor,line width= 0.6pt,dash pattern=on 4pt off 4pt ,line join=round,line cap=round,fill=fillColor] (183.38, 38.15) --
	(183.38, 38.15) --
	(183.39, 38.15) --
	(183.41, 38.16) --
	(183.42, 38.17) --
	(183.44, 38.17) --
	(183.45, 38.18) --
	(183.47, 38.18) --
	(183.49, 38.18) --
	(183.50, 38.18) --
	(183.52, 38.18) --
	(183.53, 38.18) --
	(183.55, 38.18) --
	(183.57, 38.17) --
	(183.58, 38.17) --
	(183.60, 38.17) --
	(183.61, 38.16) --
	(183.63, 38.15) --
	(183.64, 38.15) --
	(183.38, 38.15) --
	cycle;

\path[draw=drawColor,line width= 0.6pt,line join=round,line cap=rect,fill=fillColor] (245.02, 38.15) --
	(245.55, 38.45) --
	(245.55, 38.15) --
	(245.02, 38.15) --
	cycle;

\path[draw=drawColor,line width= 0.6pt,line join=round,line cap=rect,fill=fillColor] (234.75, 38.15) --
	(245.55, 44.38) --
	(245.55, 43.79) --
	(235.78, 38.15) --
	(234.75, 38.15) --
	cycle;

\path[draw=drawColor,line width= 0.6pt,line join=round,line cap=rect,fill=fillColor] (224.47, 38.15) --
	(241.14, 47.77) --
	(242.16, 47.77) --
	(225.50, 38.15) --
	(224.47, 38.15) --
	cycle;

\path[draw=drawColor,line width= 0.6pt,line join=round,line cap=rect,fill=fillColor] (220.25, 41.64) --
	(230.86, 47.77) --
	(231.89, 47.77) --
	(220.25, 41.05) --
	(220.25, 41.64) --
	cycle;

\path[draw=drawColor,line width= 0.6pt,line join=round,line cap=rect,fill=fillColor] (220.25, 47.57) --
	(220.59, 47.77) --
	(221.61, 47.77) --
	(220.25, 46.98) --
	(220.25, 47.57) --
	cycle;

\path[draw=drawColor,line width= 0.6pt,dash pattern=on 2pt off 2pt ,line join=round,line cap=rect,fill=fillColor] (265.57, 38.15) --
	(273.65, 42.81) --
	(273.65, 42.22) --
	(266.60, 38.15) --
	(265.57, 38.15) --
	cycle;

\path[draw=drawColor,line width= 0.6pt,dash pattern=on 2pt off 2pt ,line join=round,line cap=rect,fill=fillColor] (255.30, 38.15) --
	(273.65, 48.74) --
	(273.65, 48.15) --
	(256.33, 38.15) --
	(255.30, 38.15) --
	cycle;

\path[draw=drawColor,line width= 0.6pt,dash pattern=on 2pt off 2pt ,line join=round,line cap=rect,fill=fillColor] (248.36, 40.07) --
	(273.65, 54.67) --
	(273.65, 54.08) --
	(248.36, 39.48) --
	(248.36, 40.07) --
	cycle;

\path[draw=drawColor,line width= 0.6pt,dash pattern=on 2pt off 2pt ,line join=round,line cap=rect,fill=fillColor] (248.36, 46.00) --
	(273.65, 60.61) --
	(273.65, 60.01) --
	(248.36, 45.41) --
	(248.36, 46.00) --
	cycle;

\path[draw=drawColor,line width= 0.6pt,dash pattern=on 2pt off 2pt ,line join=round,line cap=rect,fill=fillColor] (248.36, 51.93) --
	(273.65, 66.54) --
	(273.65, 65.94) --
	(248.36, 51.34) --
	(248.36, 51.93) --
	cycle;

\path[draw=drawColor,line width= 0.6pt,dash pattern=on 2pt off 2pt ,line join=round,line cap=rect,fill=fillColor] (248.36, 57.87) --
	(273.65, 72.47) --
	(273.65, 71.88) --
	(248.36, 57.27) --
	(248.36, 57.87) --
	cycle;

\path[draw=drawColor,line width= 0.6pt,dash pattern=on 2pt off 2pt ,line join=round,line cap=rect,fill=fillColor] (248.36, 63.80) --
	(273.65, 78.40) --
	(273.65, 77.81) --
	(248.36, 63.20) --
	(248.36, 63.80) --
	cycle;

\path[draw=drawColor,line width= 0.6pt,dash pattern=on 2pt off 2pt ,line join=round,line cap=rect,fill=fillColor] (248.36, 69.73) --
	(273.65, 84.33) --
	(273.65, 83.74) --
	(248.36, 69.14) --
	(248.36, 69.73) --
	cycle;

\path[draw=drawColor,line width= 0.6pt,dash pattern=on 2pt off 2pt ,line join=round,line cap=rect,fill=fillColor] (248.36, 75.66) --
	(273.65, 90.27) --
	(273.65, 89.67) --
	(248.36, 75.07) --
	(248.36, 75.66) --
	cycle;

\path[draw=drawColor,line width= 0.6pt,dash pattern=on 2pt off 2pt ,line join=round,line cap=rect,fill=fillColor] (248.36, 81.59) --
	(273.65, 96.20) --
	(273.65, 95.60) --
	(248.36, 81.00) --
	(248.36, 81.59) --
	cycle;

\path[draw=drawColor,line width= 0.6pt,dash pattern=on 2pt off 2pt ,line join=round,line cap=rect,fill=fillColor] (248.36, 87.53) --
	(273.65,102.13) --
	(273.65,101.54) --
	(248.36, 86.93) --
	(248.36, 87.53) --
	cycle;

\path[draw=drawColor,line width= 0.6pt,dash pattern=on 2pt off 2pt ,line join=round,line cap=rect,fill=fillColor] (248.36, 93.46) --
	(273.65,108.06) --
	(273.65,107.47) --
	(248.36, 92.86) --
	(248.36, 93.46) --
	cycle;

\path[draw=drawColor,line width= 0.6pt,dash pattern=on 2pt off 2pt ,line join=round,line cap=rect,fill=fillColor] (248.36, 99.39) --
	(273.65,113.99) --
	(273.65,113.40) --
	(248.36, 98.80) --
	(248.36, 99.39) --
	cycle;

\path[draw=drawColor,line width= 0.6pt,dash pattern=on 2pt off 2pt ,line join=round,line cap=rect,fill=fillColor] (248.36,105.32) --
	(273.65,119.93) --
	(273.65,119.33) --
	(248.36,104.73) --
	(248.36,105.32) --
	cycle;

\path[draw=drawColor,line width= 0.6pt,dash pattern=on 2pt off 2pt ,line join=round,line cap=rect,fill=fillColor] (248.36,111.25) --
	(273.65,125.86) --
	(273.65,125.26) --
	(248.36,110.66) --
	(248.36,111.25) --
	cycle;

\path[draw=drawColor,line width= 0.6pt,dash pattern=on 2pt off 2pt ,line join=round,line cap=rect,fill=fillColor] (248.36,117.19) --
	(272.63,131.20) --
	(273.65,131.20) --
	(273.65,131.20) --
	(248.36,116.59) --
	(248.36,117.19) --
	cycle;

\path[draw=drawColor,line width= 0.6pt,dash pattern=on 2pt off 2pt ,line join=round,line cap=rect,fill=fillColor] (248.36,123.12) --
	(262.35,131.20) --
	(263.38,131.20) --
	(248.36,122.53) --
	(248.36,123.12) --
	cycle;

\path[draw=drawColor,line width= 0.6pt,dash pattern=on 2pt off 2pt ,line join=round,line cap=rect,fill=fillColor] (248.36,129.05) --
	(252.08,131.20) --
	(253.11,131.20) --
	(248.36,128.46) --
	(248.36,129.05) --
	cycle;

\path[draw=drawColor,line width= 0.6pt,dash pattern=on 4pt off 2pt ,line join=round,line cap=rect,fill=fillColor] (296.40, 38.15) --
	(301.75, 41.24) --
	(301.75, 40.65) --
	(297.42, 38.15) --
	(296.40, 38.15) --
	cycle;

\path[draw=drawColor,line width= 0.6pt,dash pattern=on 4pt off 2pt ,line join=round,line cap=rect,fill=fillColor] (286.12, 38.15) --
	(301.75, 47.17) --
	(301.75, 46.58) --
	(287.15, 38.15) --
	(286.12, 38.15) --
	cycle;

\path[draw=drawColor,line width= 0.6pt,dash pattern=on 4pt off 2pt ,line join=round,line cap=rect,fill=fillColor] (276.46, 38.50) --
	(297.43, 50.61) --
	(298.46, 50.61) --
	(276.87, 38.15) --
	(276.46, 38.15) --
	(276.46, 38.50) --
	cycle;

\path[draw=drawColor,line width= 0.6pt,dash pattern=on 4pt off 2pt ,line join=round,line cap=rect,fill=fillColor] (276.46, 44.43) --
	(287.15, 50.61) --
	(288.18, 50.61) --
	(276.46, 43.84) --
	(276.46, 44.43) --
	cycle;

\path[draw=drawColor,line width= 0.6pt,dash pattern=on 4pt off 2pt ,line join=round,line cap=rect,fill=fillColor] (276.46, 50.36) --
	(276.88, 50.61) --
	(277.91, 50.61) --
	(276.46, 49.77) --
	(276.46, 50.36) --
	cycle;

\path[draw=drawColor,line width= 0.6pt,dash pattern=on 4pt off 2pt ,line join=round,line cap=rect,fill=fillColor] (276.46, 41.83) --
	(278.59, 38.15) --
	(277.99, 38.15) --
	(276.46, 40.80) --
	(276.46, 41.83) --
	cycle;

\path[draw=drawColor,line width= 0.6pt,dash pattern=on 4pt off 2pt ,line join=round,line cap=rect,fill=fillColor] (277.33, 50.61) --
	(284.52, 38.15) --
	(283.93, 38.15) --
	(276.73, 50.61) --
	(277.33, 50.61) --
	cycle;

\path[draw=drawColor,line width= 0.6pt,dash pattern=on 4pt off 2pt ,line join=round,line cap=rect,fill=fillColor] (283.26, 50.61) --
	(290.45, 38.15) --
	(289.86, 38.15) --
	(282.66, 50.61) --
	(283.26, 50.61) --
	cycle;

\path[draw=drawColor,line width= 0.6pt,dash pattern=on 4pt off 2pt ,line join=round,line cap=rect,fill=fillColor] (289.19, 50.61) --
	(296.38, 38.15) --
	(295.79, 38.15) --
	(288.60, 50.61) --
	(289.19, 50.61) --
	cycle;

\path[draw=drawColor,line width= 0.6pt,dash pattern=on 4pt off 2pt ,line join=round,line cap=rect,fill=fillColor] (295.12, 50.61) --
	(301.75, 39.12) --
	(301.75, 38.15) --
	(301.72, 38.15) --
	(294.53, 50.61) --
	(295.12, 50.61) --
	cycle;

\path[draw=drawColor,line width= 0.6pt,dash pattern=on 4pt off 2pt ,line join=round,line cap=rect,fill=fillColor] (301.05, 50.61) --
	(301.75, 49.39) --
	(301.75, 48.36) --
	(300.46, 50.61) --
	(301.05, 50.61) --
	cycle;

\path[draw=drawColor,line width= 0.6pt,dash pattern=on 4pt off 4pt ,line join=round,line cap=round,fill=fillColor] (305.19, 42.93) circle (  0.26);

\path[draw=drawColor,line width= 0.6pt,dash pattern=on 4pt off 4pt ,line join=round,line cap=round,fill=fillColor] (307.76, 38.48) circle (  0.26);

\path[draw=drawColor,line width= 0.6pt,dash pattern=on 4pt off 4pt ,line join=round,line cap=round,fill=fillColor] (309.64, 45.50) circle (  0.26);

\path[draw=drawColor,line width= 0.6pt,dash pattern=on 4pt off 4pt ,line join=round,line cap=round,fill=fillColor] (312.21, 41.05) circle (  0.26);

\path[draw=drawColor,line width= 0.6pt,dash pattern=on 4pt off 4pt ,line join=round,line cap=round,fill=fillColor] (316.66, 43.61) circle (  0.26);

\path[draw=drawColor,line width= 0.6pt,dash pattern=on 4pt off 4pt ,line join=round,line cap=round,fill=fillColor] (319.23, 39.17) circle (  0.26);

\path[draw=drawColor,line width= 0.6pt,dash pattern=on 4pt off 4pt ,line join=round,line cap=round,fill=fillColor] (321.11, 46.18) circle (  0.26);

\path[draw=drawColor,line width= 0.6pt,dash pattern=on 4pt off 4pt ,line join=round,line cap=round,fill=fillColor] (323.68, 41.73) circle (  0.26);

\path[draw=drawColor,line width= 0.6pt,dash pattern=on 4pt off 4pt ,line join=round,line cap=round,fill=fillColor] (328.12, 44.30) circle (  0.26);

\path[draw=drawColor,line width= 0.6pt,dash pattern=on 4pt off 4pt ,line join=round,line cap=round,fill=fillColor] (314.33, 48.16) --
	(314.33, 48.15) --
	(314.34, 48.13) --
	(314.34, 48.12) --
	(314.34, 48.10) --
	(314.34, 48.09) --
	(314.35, 48.07) --
	(314.35, 48.05) --
	(314.34, 48.04) --
	(314.34, 48.02) --
	(314.34, 48.01) --
	(314.33, 47.99) --
	(314.33, 47.97) --
	(314.32, 47.96) --
	(314.32, 47.94) --
	(314.31, 47.93) --
	(314.30, 47.92) --
	(314.29, 47.90) --
	(314.28, 47.89) --
	(314.27, 47.88) --
	(314.26, 47.87) --
	(314.24, 47.86) --
	(314.23, 47.85) --
	(314.22, 47.84) --
	(314.20, 47.83) --
	(314.19, 47.83) --
	(314.17, 47.82) --
	(314.16, 47.82) --
	(314.14, 47.81) --
	(314.13, 47.81) --
	(314.11, 47.81) --
	(314.09, 47.81) --
	(314.08, 47.81) --
	(314.06, 47.81) --
	(314.05, 47.81) --
	(314.03, 47.81) --
	(314.01, 47.82) --
	(314.00, 47.82) --
	(313.98, 47.83) --
	(313.97, 47.84) --
	(313.96, 47.84) --
	(313.94, 47.85) --
	(313.93, 47.86) --
	(313.92, 47.87) --
	(313.91, 47.88) --
	(313.89, 47.90) --
	(313.88, 47.91) --
	(313.88, 47.92) --
	(313.87, 47.94) --
	(313.86, 47.95) --
	(313.85, 47.96) --
	(313.85, 47.98) --
	(313.84, 47.99) --
	(313.84, 48.01) --
	(313.83, 48.03) --
	(313.83, 48.04) --
	(313.83, 48.06) --
	(313.83, 48.07) --
	(313.83, 48.09) --
	(313.84, 48.11) --
	(313.84, 48.12) --
	(313.84, 48.14) --
	(313.85, 48.15) --
	(313.85, 48.16) --
	(314.33, 48.16) --
	cycle;

\path[draw=drawColor,line width= 0.6pt,line join=round,line cap=rect,fill=fillColor] (378.59, 38.15) --
	(386.07, 42.46) --
	(386.07, 41.87) --
	(379.62, 38.15) --
	(378.59, 38.15) --
	cycle;

\path[draw=drawColor,line width= 0.6pt,line join=round,line cap=rect,fill=fillColor] (368.32, 38.15) --
	(384.38, 47.42) --
	(385.41, 47.42) --
	(369.35, 38.15) --
	(368.32, 38.15) --
	cycle;

\path[draw=drawColor,line width= 0.6pt,line join=round,line cap=rect,fill=fillColor] (360.77, 39.72) --
	(374.11, 47.42) --
	(375.13, 47.42) --
	(360.77, 39.13) --
	(360.77, 39.72) --
	cycle;

\path[draw=drawColor,line width= 0.6pt,line join=round,line cap=rect,fill=fillColor] (360.77, 45.65) --
	(363.83, 47.42) --
	(364.86, 47.42) --
	(360.77, 45.06) --
	(360.77, 45.65) --
	cycle;

\path[draw=drawColor,line width= 0.6pt,dash pattern=on 2pt off 2pt ,line join=round,line cap=rect,fill=fillColor] (409.42, 38.15) --
	(414.17, 40.89) --
	(414.17, 40.30) --
	(410.44, 38.15) --
	(409.42, 38.15) --
	cycle;

\path[draw=drawColor,line width= 0.6pt,dash pattern=on 2pt off 2pt ,line join=round,line cap=rect,fill=fillColor] (399.14, 38.15) --
	(414.17, 46.82) --
	(414.17, 46.23) --
	(400.17, 38.15) --
	(399.14, 38.15) --
	cycle;

\path[draw=drawColor,line width= 0.6pt,dash pattern=on 2pt off 2pt ,line join=round,line cap=rect,fill=fillColor] (388.88, 38.15) --
	(414.17, 52.76) --
	(414.17, 52.16) --
	(389.89, 38.15) --
	(388.88, 38.15) --
	(388.88, 38.15) --
	cycle;

\path[draw=drawColor,line width= 0.6pt,dash pattern=on 2pt off 2pt ,line join=round,line cap=rect,fill=fillColor] (388.88, 44.08) --
	(414.17, 58.69) --
	(414.17, 58.09) --
	(388.88, 43.49) --
	(388.88, 44.08) --
	cycle;

\path[draw=drawColor,line width= 0.6pt,dash pattern=on 2pt off 2pt ,line join=round,line cap=rect,fill=fillColor] (388.88, 50.02) --
	(414.17, 64.62) --
	(414.17, 64.03) --
	(388.88, 49.42) --
	(388.88, 50.02) --
	cycle;

\path[draw=drawColor,line width= 0.6pt,dash pattern=on 2pt off 2pt ,line join=round,line cap=rect,fill=fillColor] (388.88, 55.95) --
	(414.17, 70.55) --
	(414.17, 69.96) --
	(388.88, 55.35) --
	(388.88, 55.95) --
	cycle;

\path[draw=drawColor,line width= 0.6pt,dash pattern=on 2pt off 2pt ,line join=round,line cap=rect,fill=fillColor] (388.88, 61.88) --
	(414.17, 76.48) --
	(414.17, 75.89) --
	(388.88, 61.29) --
	(388.88, 61.88) --
	cycle;

\path[draw=drawColor,line width= 0.6pt,dash pattern=on 2pt off 2pt ,line join=round,line cap=rect,fill=fillColor] (388.88, 67.81) --
	(414.17, 82.42) --
	(414.17, 81.82) --
	(388.88, 67.22) --
	(388.88, 67.81) --
	cycle;

\path[draw=drawColor,line width= 0.6pt,dash pattern=on 2pt off 2pt ,line join=round,line cap=rect,fill=fillColor] (388.88, 73.74) --
	(414.17, 88.35) --
	(414.17, 87.75) --
	(388.88, 73.15) --
	(388.88, 73.74) --
	cycle;

\path[draw=drawColor,line width= 0.6pt,dash pattern=on 2pt off 2pt ,line join=round,line cap=rect,fill=fillColor] (388.88, 79.68) --
	(414.17, 94.28) --
	(414.17, 93.69) --
	(388.88, 79.08) --
	(388.88, 79.68) --
	cycle;

\path[draw=drawColor,line width= 0.6pt,dash pattern=on 2pt off 2pt ,line join=round,line cap=rect,fill=fillColor] (388.88, 85.61) --
	(414.17,100.21) --
	(414.17, 99.62) --
	(388.88, 85.01) --
	(388.88, 85.61) --
	cycle;

\path[draw=drawColor,line width= 0.6pt,dash pattern=on 2pt off 2pt ,line join=round,line cap=rect,fill=fillColor] (388.88, 91.54) --
	(414.17,106.14) --
	(414.17,105.55) --
	(388.88, 90.95) --
	(388.88, 91.54) --
	cycle;

\path[draw=drawColor,line width= 0.6pt,dash pattern=on 2pt off 2pt ,line join=round,line cap=rect,fill=fillColor] (388.88, 97.47) --
	(414.17,112.08) --
	(414.17,111.48) --
	(388.88, 96.88) --
	(388.88, 97.47) --
	cycle;

\path[draw=drawColor,line width= 0.6pt,dash pattern=on 2pt off 2pt ,line join=round,line cap=rect,fill=fillColor] (388.88,103.40) --
	(414.17,118.01) --
	(414.17,117.41) --
	(388.88,102.81) --
	(388.88,103.40) --
	cycle;

\path[draw=drawColor,line width= 0.6pt,dash pattern=on 2pt off 2pt ,line join=round,line cap=rect,fill=fillColor] (388.88,109.34) --
	(414.17,123.94) --
	(414.17,123.35) --
	(388.88,108.74) --
	(388.88,109.34) --
	cycle;

\path[draw=drawColor,line width= 0.6pt,dash pattern=on 2pt off 2pt ,line join=round,line cap=rect,fill=fillColor] (388.88,115.27) --
	(414.17,129.87) --
	(414.17,129.28) --
	(388.88,114.67) --
	(388.88,115.27) --
	cycle;

\path[draw=drawColor,line width= 0.6pt,dash pattern=on 2pt off 2pt ,line join=round,line cap=rect,fill=fillColor] (388.88,121.20) --
	(406.81,131.55) --
	(407.83,131.55) --
	(388.88,120.61) --
	(388.88,121.20) --
	cycle;

\path[draw=drawColor,line width= 0.6pt,dash pattern=on 2pt off 2pt ,line join=round,line cap=rect,fill=fillColor] (388.88,127.13) --
	(396.53,131.55) --
	(397.56,131.55) --
	(388.88,126.54) --
	(388.88,127.13) --
	cycle;

\path[draw=drawColor,line width= 0.6pt,dash pattern=on 4pt off 2pt ,line join=round,line cap=rect,fill=fillColor] (440.24, 38.15) --
	(442.28, 39.32) --
	(442.28, 38.73) --
	(441.27, 38.15) --
	(440.24, 38.15) --
	cycle;

\path[draw=drawColor,line width= 0.6pt,dash pattern=on 4pt off 2pt ,line join=round,line cap=rect,fill=fillColor] (429.97, 38.15) --
	(442.28, 45.25) --
	(442.28, 44.66) --
	(430.99, 38.15) --
	(429.97, 38.15) --
	cycle;

\path[draw=drawColor,line width= 0.6pt,dash pattern=on 4pt off 2pt ,line join=round,line cap=rect,fill=fillColor] (419.69, 38.15) --
	(441.92, 50.98) --
	(442.28, 50.98) --
	(442.28, 50.59) --
	(420.72, 38.15) --
	(419.69, 38.15) --
	cycle;

\path[draw=drawColor,line width= 0.6pt,dash pattern=on 4pt off 2pt ,line join=round,line cap=rect,fill=fillColor] (416.98, 42.51) --
	(431.65, 50.98) --
	(432.67, 50.98) --
	(416.98, 41.92) --
	(416.98, 42.51) --
	cycle;

\path[draw=drawColor,line width= 0.6pt,dash pattern=on 4pt off 2pt ,line join=round,line cap=rect,fill=fillColor] (416.98, 48.45) --
	(421.37, 50.98) --
	(422.40, 50.98) --
	(416.98, 47.85) --
	(416.98, 48.45) --
	cycle;

\path[draw=drawColor,line width= 0.6pt,dash pattern=on 4pt off 2pt ,line join=round,line cap=rect,fill=fillColor] (416.98, 45.03) --
	(420.96, 38.15) --
	(420.36, 38.15) --
	(416.98, 44.00) --
	(416.98, 45.03) --
	cycle;

\path[draw=drawColor,line width= 0.6pt,dash pattern=on 4pt off 2pt ,line join=round,line cap=rect,fill=fillColor] (419.48, 50.98) --
	(426.89, 38.15) --
	(426.30, 38.15) --
	(418.89, 50.98) --
	(419.48, 50.98) --
	cycle;

\path[draw=drawColor,line width= 0.6pt,dash pattern=on 4pt off 2pt ,line join=round,line cap=rect,fill=fillColor] (425.41, 50.98) --
	(432.82, 38.15) --
	(432.23, 38.15) --
	(424.82, 50.98) --
	(425.41, 50.98) --
	cycle;

\path[draw=drawColor,line width= 0.6pt,dash pattern=on 4pt off 2pt ,line join=round,line cap=rect,fill=fillColor] (431.34, 50.98) --
	(438.75, 38.15) --
	(438.16, 38.15) --
	(430.75, 50.98) --
	(431.34, 50.98) --
	cycle;

\path[draw=drawColor,line width= 0.6pt,dash pattern=on 4pt off 2pt ,line join=round,line cap=rect,fill=fillColor] (437.27, 50.98) --
	(442.28, 42.32) --
	(442.28, 41.29) --
	(436.68, 50.98) --
	(437.27, 50.98) --
	cycle;

\path[draw=drawColor,line width= 0.6pt,dash pattern=on 4pt off 4pt ,line join=round,line cap=round,fill=fillColor] (445.36, 46.74) circle (  0.26);

\path[draw=drawColor,line width= 0.6pt,dash pattern=on 4pt off 4pt ,line join=round,line cap=round,fill=fillColor] (447.93, 42.29) circle (  0.26);

\path[draw=drawColor,line width= 0.6pt,dash pattern=on 4pt off 4pt ,line join=round,line cap=round,fill=fillColor] (452.38, 44.86) circle (  0.26);

\path[draw=drawColor,line width= 0.6pt,dash pattern=on 4pt off 4pt ,line join=round,line cap=round,fill=fillColor] (454.95, 40.41) circle (  0.26);

\path[draw=drawColor,line width= 0.6pt,dash pattern=on 4pt off 4pt ,line join=round,line cap=round,fill=fillColor] (456.83, 47.43) circle (  0.26);

\path[draw=drawColor,line width= 0.6pt,dash pattern=on 4pt off 4pt ,line join=round,line cap=round,fill=fillColor] (459.40, 42.98) circle (  0.26);

\path[draw=drawColor,line width= 0.6pt,dash pattern=on 4pt off 4pt ,line join=round,line cap=round,fill=fillColor] (461.96, 38.53) circle (  0.26);

\path[draw=drawColor,line width= 0.6pt,dash pattern=on 4pt off 4pt ,line join=round,line cap=round,fill=fillColor] (463.84, 45.54) circle (  0.26);

\path[draw=drawColor,line width= 0.6pt,dash pattern=on 4pt off 4pt ,line join=round,line cap=round,fill=fillColor] (466.41, 41.10) circle (  0.26);
\definecolor{drawColor}{RGB}{102,194,165}

\path[draw=drawColor,line width= 0.6pt,line cap=rect] ( 79.73, 38.15) rectangle (105.02,118.55);
\definecolor{drawColor}{RGB}{252,141,98}

\path[draw=drawColor,line width= 0.6pt,line cap=rect] (107.83, 38.15) rectangle (133.13, 42.70);
\definecolor{drawColor}{RGB}{141,160,203}

\path[draw=drawColor,line width= 0.6pt,line cap=rect] (135.94, 38.15) rectangle (161.23, 75.47);
\definecolor{drawColor}{RGB}{231,138,195}

\path[draw=drawColor,line width= 0.6pt,line cap=rect] (164.04, 38.15) rectangle (189.34, 41.03);
\definecolor{drawColor}{RGB}{102,194,165}

\path[draw=drawColor,line width= 0.6pt,line cap=rect] (220.25, 38.15) rectangle (245.55, 47.77);
\definecolor{drawColor}{RGB}{252,141,98}

\path[draw=drawColor,line width= 0.6pt,line cap=rect] (248.36, 38.15) rectangle (273.65,131.20);
\definecolor{drawColor}{RGB}{141,160,203}

\path[draw=drawColor,line width= 0.6pt,line cap=rect] (276.46, 38.15) rectangle (301.75, 50.61);
\definecolor{drawColor}{RGB}{231,138,195}

\path[draw=drawColor,line width= 0.6pt,line cap=rect] (304.57, 38.15) rectangle (329.86, 48.16);
\definecolor{drawColor}{RGB}{102,194,165}

\path[draw=drawColor,line width= 0.6pt,line cap=rect] (360.77, 38.15) rectangle (386.07, 47.42);
\definecolor{drawColor}{RGB}{252,141,98}

\path[draw=drawColor,line width= 0.6pt,line cap=rect] (388.88, 38.15) rectangle (414.17,131.55);
\definecolor{drawColor}{RGB}{141,160,203}

\path[draw=drawColor,line width= 0.6pt,line cap=rect] (416.98, 38.15) rectangle (442.28, 50.98);
\definecolor{drawColor}{RGB}{231,138,195}

\path[draw=drawColor,line width= 0.6pt,line cap=rect] (445.09, 38.15) rectangle (470.38, 47.78);
\end{scope}
\begin{scope}
\path[clip] (  0.00,  0.00) rectangle (505.89,180.67);
\definecolor{drawColor}{gray}{0.30}

\node[text=drawColor,anchor=base east,inner sep=0pt, outer sep=0pt, scale=  0.96] at ( 44.82, 34.84) {0.0};

\node[text=drawColor,anchor=base east,inner sep=0pt, outer sep=0pt, scale=  0.96] at ( 44.82, 59.87) {0.2};

\node[text=drawColor,anchor=base east,inner sep=0pt, outer sep=0pt, scale=  0.96] at ( 44.82, 84.90) {0.4};

\node[text=drawColor,anchor=base east,inner sep=0pt, outer sep=0pt, scale=  0.96] at ( 44.82,109.93) {0.6};
\end{scope}
\begin{scope}
\path[clip] (  0.00,  0.00) rectangle (505.89,180.67);
\definecolor{drawColor}{gray}{0.30}

\node[text=drawColor,anchor=base,inner sep=0pt, outer sep=0pt, scale=  0.96] at (134.53, 21.46) {$A\indep C$};

\node[text=drawColor,anchor=base,inner sep=0pt, outer sep=0pt, scale=  0.96] at (275.06, 21.46) {$A\rightsquigarrow C$};

\node[text=drawColor,anchor=base,inner sep=0pt, outer sep=0pt, scale=  0.96] at (415.58, 21.46) {$C\rightsquigarrow A$};
\end{scope}
\begin{scope}
\path[clip] (  0.00,  0.00) rectangle (505.89,180.67);
\definecolor{drawColor}{RGB}{0,0,0}

\node[text=drawColor,anchor=base,inner sep=0pt, outer sep=0pt, scale=  1.20] at (275.06,  8.33) {causal relation $R$};
\end{scope}
\begin{scope}
\path[clip] (  0.00,  0.00) rectangle (505.89,180.67);
\definecolor{drawColor}{RGB}{0,0,0}

\node[text=drawColor,rotate= 90.00,anchor=base,inner sep=0pt, outer sep=0pt, scale=  1.20] at ( 14.26, 84.85) {expected utterance};

\node[text=drawColor,rotate= 90.00,anchor=base,inner sep=0pt, outer sep=0pt, scale=  1.20] at ( 27.22, 84.85) {choice probability};
\end{scope}
\begin{scope}
\path[clip] (  0.00,  0.00) rectangle (505.89,180.67);
\definecolor{drawColor}{RGB}{0,0,0}

\node[text=drawColor,anchor=base west,inner sep=0pt, outer sep=0pt, scale=  1.20] at ( 57.61,157.32) {utterance type};
\end{scope}
\begin{scope}
\path[clip] (  0.00,  0.00) rectangle (505.89,180.67);
\definecolor{fillColor}{RGB}{102,194,165}

\path[fill=fillColor] (149.01,154.22) rectangle (163.47,168.67);
\definecolor{drawColor}{RGB}{255,255,255}
\definecolor{fillColor}{RGB}{255,255,255}

\path[draw=drawColor,line width= 0.6pt,line join=round,line cap=rect,fill=fillColor] (159.42,154.33) --
	(163.35,156.61) --
	(163.35,156.15) --
	(160.22,154.33) --
	(159.42,154.33) --
	cycle;

\path[draw=drawColor,line width= 0.6pt,line join=round,line cap=rect,fill=fillColor] (151.47,154.33) --
	(163.35,161.19) --
	(163.35,160.74) --
	(152.27,154.33) --
	(151.47,154.33) --
	cycle;

\path[draw=drawColor,line width= 0.6pt,line join=round,line cap=rect,fill=fillColor] (149.13,157.57) --
	(163.35,165.78) --
	(163.35,165.33) --
	(149.13,157.11) --
	(149.13,157.57) --
	cycle;

\path[draw=drawColor,line width= 0.6pt,line join=round,line cap=rect,fill=fillColor] (149.13,162.16) --
	(160.21,168.56) --
	(161.01,168.56) --
	(149.13,161.70) --
	(149.13,162.16) --
	cycle;

\path[draw=drawColor,line width= 0.6pt,line join=round,line cap=rect,fill=fillColor] (149.13,166.75) --
	(152.27,168.56) --
	(153.06,168.56) --
	(149.13,166.29) --
	(149.13,166.75) --
	cycle;
\definecolor{drawColor}{RGB}{102,194,165}

\path[draw=drawColor,line width= 0.2pt,line cap=rect] (149.01,154.22) rectangle (163.47,168.67);
\end{scope}
\begin{scope}
\path[clip] (  0.00,  0.00) rectangle (505.89,180.67);
\definecolor{fillColor}{RGB}{252,141,98}

\path[fill=fillColor] (254.13,154.22) rectangle (268.58,168.67);
\definecolor{drawColor}{RGB}{255,255,255}
\definecolor{fillColor}{RGB}{255,255,255}

\path[draw=drawColor,line width= 0.6pt,dash pattern=on 2pt off 2pt ,line join=round,line cap=rect,fill=fillColor] (264.53,154.33) --
	(268.47,156.61) --
	(268.47,156.15) --
	(265.33,154.33) --
	(264.53,154.33) --
	cycle;

\path[draw=drawColor,line width= 0.6pt,dash pattern=on 2pt off 2pt ,line join=round,line cap=rect,fill=fillColor] (256.58,154.33) --
	(268.47,161.19) --
	(268.47,160.74) --
	(257.38,154.33) --
	(256.58,154.33) --
	cycle;

\path[draw=drawColor,line width= 0.6pt,dash pattern=on 2pt off 2pt ,line join=round,line cap=rect,fill=fillColor] (254.24,157.57) --
	(268.47,165.78) --
	(268.47,165.33) --
	(254.24,157.11) --
	(254.24,157.57) --
	cycle;

\path[draw=drawColor,line width= 0.6pt,dash pattern=on 2pt off 2pt ,line join=round,line cap=rect,fill=fillColor] (254.24,162.16) --
	(265.33,168.56) --
	(266.12,168.56) --
	(254.24,161.70) --
	(254.24,162.16) --
	cycle;

\path[draw=drawColor,line width= 0.6pt,dash pattern=on 2pt off 2pt ,line join=round,line cap=rect,fill=fillColor] (254.24,166.75) --
	(257.38,168.56) --
	(258.17,168.56) --
	(254.24,166.29) --
	(254.24,166.75) --
	cycle;
\definecolor{drawColor}{RGB}{252,141,98}

\path[draw=drawColor,line width= 0.2pt,line cap=rect] (254.13,154.22) rectangle (268.58,168.67);
\end{scope}
\begin{scope}
\path[clip] (  0.00,  0.00) rectangle (505.89,180.67);
\definecolor{fillColor}{RGB}{141,160,203}

\path[fill=fillColor] (345.08,154.22) rectangle (359.54,168.67);
\definecolor{drawColor}{RGB}{255,255,255}
\definecolor{fillColor}{RGB}{255,255,255}

\path[draw=drawColor,line width= 0.6pt,dash pattern=on 4pt off 2pt ,line join=round,line cap=rect,fill=fillColor] (355.49,154.33) --
	(359.42,156.61) --
	(359.42,156.15) --
	(356.28,154.33) --
	(355.49,154.33) --
	cycle;

\path[draw=drawColor,line width= 0.6pt,dash pattern=on 4pt off 2pt ,line join=round,line cap=rect,fill=fillColor] (347.54,154.33) --
	(359.42,161.19) --
	(359.42,160.74) --
	(348.34,154.33) --
	(347.54,154.33) --
	cycle;

\path[draw=drawColor,line width= 0.6pt,dash pattern=on 4pt off 2pt ,line join=round,line cap=rect,fill=fillColor] (345.20,157.57) --
	(359.42,165.78) --
	(359.42,165.33) --
	(345.20,157.11) --
	(345.20,157.57) --
	cycle;

\path[draw=drawColor,line width= 0.6pt,dash pattern=on 4pt off 2pt ,line join=round,line cap=rect,fill=fillColor] (345.20,162.16) --
	(356.28,168.56) --
	(357.08,168.56) --
	(345.20,161.70) --
	(345.20,162.16) --
	cycle;

\path[draw=drawColor,line width= 0.6pt,dash pattern=on 4pt off 2pt ,line join=round,line cap=rect,fill=fillColor] (345.20,166.75) --
	(348.33,168.56) --
	(349.13,168.56) --
	(345.20,166.29) --
	(345.20,166.75) --
	cycle;

\path[draw=drawColor,line width= 0.6pt,dash pattern=on 4pt off 2pt ,line join=round,line cap=rect,fill=fillColor] (345.20,158.27) --
	(347.47,154.33) --
	(347.01,154.33) --
	(345.20,157.47) --
	(345.20,158.27) --
	cycle;

\path[draw=drawColor,line width= 0.6pt,dash pattern=on 4pt off 2pt ,line join=round,line cap=rect,fill=fillColor] (345.20,166.22) --
	(352.06,154.33) --
	(351.60,154.33) --
	(345.20,165.42) --
	(345.20,166.22) --
	cycle;

\path[draw=drawColor,line width= 0.6pt,dash pattern=on 4pt off 2pt ,line join=round,line cap=rect,fill=fillColor] (348.43,168.56) --
	(356.65,154.33) --
	(356.19,154.33) --
	(347.97,168.56) --
	(348.43,168.56) --
	cycle;

\path[draw=drawColor,line width= 0.6pt,dash pattern=on 4pt off 2pt ,line join=round,line cap=rect,fill=fillColor] (353.02,168.56) --
	(359.42,157.47) --
	(359.42,156.68) --
	(352.56,168.56) --
	(353.02,168.56) --
	cycle;

\path[draw=drawColor,line width= 0.6pt,dash pattern=on 4pt off 2pt ,line join=round,line cap=rect,fill=fillColor] (357.61,168.56) --
	(359.42,165.42) --
	(359.42,164.63) --
	(357.15,168.56) --
	(357.61,168.56) --
	cycle;
\definecolor{drawColor}{RGB}{141,160,203}

\path[draw=drawColor,line width= 0.2pt,line cap=rect] (345.08,154.22) rectangle (359.54,168.67);
\end{scope}
\begin{scope}
\path[clip] (  0.00,  0.00) rectangle (505.89,180.67);
\definecolor{fillColor}{RGB}{231,138,195}

\path[fill=fillColor] (414.20,154.22) rectangle (428.66,168.67);
\definecolor{drawColor}{RGB}{255,255,255}
\definecolor{fillColor}{RGB}{255,255,255}

\path[draw=drawColor,line width= 0.6pt,dash pattern=on 4pt off 4pt ,line join=round,line cap=round,fill=fillColor] (414.55,157.47) circle (  0.20);

\path[draw=drawColor,line width= 0.6pt,dash pattern=on 4pt off 4pt ,line join=round,line cap=round,fill=fillColor] (416.00,162.90) circle (  0.20);

\path[draw=drawColor,line width= 0.6pt,dash pattern=on 4pt off 4pt ,line join=round,line cap=round,fill=fillColor] (417.45,168.33) circle (  0.20);

\path[draw=drawColor,line width= 0.6pt,dash pattern=on 4pt off 4pt ,line join=round,line cap=round,fill=fillColor] (417.99,159.46) circle (  0.20);

\path[draw=drawColor,line width= 0.6pt,dash pattern=on 4pt off 4pt ,line join=round,line cap=round,fill=fillColor] (419.44,164.89) circle (  0.20);

\path[draw=drawColor,line width= 0.6pt,dash pattern=on 4pt off 4pt ,line join=round,line cap=round,fill=fillColor] (419.97,156.02) circle (  0.20);

\path[draw=drawColor,line width= 0.6pt,dash pattern=on 4pt off 4pt ,line join=round,line cap=round,fill=fillColor] (421.43,161.45) circle (  0.20);

\path[draw=drawColor,line width= 0.6pt,dash pattern=on 4pt off 4pt ,line join=round,line cap=round,fill=fillColor] (422.88,166.88) circle (  0.20);

\path[draw=drawColor,line width= 0.6pt,dash pattern=on 4pt off 4pt ,line join=round,line cap=round,fill=fillColor] (423.42,158.01) circle (  0.20);

\path[draw=drawColor,line width= 0.6pt,dash pattern=on 4pt off 4pt ,line join=round,line cap=round,fill=fillColor] (424.87,163.44) circle (  0.20);

\path[draw=drawColor,line width= 0.6pt,dash pattern=on 4pt off 4pt ,line join=round,line cap=round,fill=fillColor] (425.40,154.56) circle (  0.20);

\path[draw=drawColor,line width= 0.6pt,dash pattern=on 4pt off 4pt ,line join=round,line cap=round,fill=fillColor] (426.86,159.99) circle (  0.20);

\path[draw=drawColor,line width= 0.6pt,dash pattern=on 4pt off 4pt ,line join=round,line cap=round,fill=fillColor] (428.31,165.42) circle (  0.20);
\definecolor{drawColor}{RGB}{231,138,195}

\path[draw=drawColor,line width= 0.2pt,line cap=rect] (414.20,154.22) rectangle (428.66,168.67);
\end{scope}
\begin{scope}
\path[clip] (  0.00,  0.00) rectangle (505.89,180.67);
\definecolor{drawColor}{RGB}{0,0,0}

\node[text=drawColor,anchor=base west,inner sep=0pt, outer sep=0pt, scale=  0.96] at (178.52,158.14) {likely + literal};
\end{scope}
\begin{scope}
\path[clip] (  0.00,  0.00) rectangle (505.89,180.67);
\definecolor{drawColor}{RGB}{0,0,0}

\node[text=drawColor,anchor=base west,inner sep=0pt, outer sep=0pt, scale=  0.96] at (283.64,158.14) {conditional};
\end{scope}
\begin{scope}
\path[clip] (  0.00,  0.00) rectangle (505.89,180.67);
\definecolor{drawColor}{RGB}{0,0,0}

\node[text=drawColor,anchor=base west,inner sep=0pt, outer sep=0pt, scale=  0.96] at (374.59,158.14) {literal};
\end{scope}
\begin{scope}
\path[clip] (  0.00,  0.00) rectangle (505.89,180.67);
\definecolor{drawColor}{RGB}{0,0,0}

\node[text=drawColor,anchor=base west,inner sep=0pt, outer sep=0pt, scale=  0.96] at (443.71,158.14) {conjunction};
\end{scope}
\end{tikzpicture}

%% file: manuscript-grusdt-lassiter-franke.bbl
\newcommand{\SortNoop}[1]{} \newcommand{\noop}[1]{}
\begin{thebibliography}{114}
\providecommand{\natexlab}[1]{#1}
\providecommand{\url}[1]{\texttt{#1}}
\providecommand{\urlprefix}{}
\expandafter\ifx\csname urlstyle\endcsname\relax
  \providecommand{\doi}[1]{doi:\discretionary{}{}{}#1}\else
  \providecommand{\doi}{doi:\discretionary{}{}{}\begingroup
  \urlstyle{rm}\Url}\fi

\bibitem[{Aloni(2007)}]{Aloni_2007:FC_BiOT}
Aloni, Maria. 2007.
\newblock Expressing ignorance or indifference. {M}odal implicatures in
  bi-directional optimality theory.
\newblock In Balder ten Cate \& Henk Zeevat (eds.), \emph{Logic, language and
  computation: Papers from the 6th international tbilisi symposium}, vol. 4363,
  1--20. Berlin: Springer Verlag.

\bibitem[{Atlas \& Levinson(1981)}]{Atlas1981a}
Atlas, Jay~David \& Stephen~C Levinson. 1981.
\newblock {It-Clefts, Informativesness, and Logical Form}.
\newblock In \emph{Radical pragmatics}, 1--61. Academic Press.

\bibitem[{Austin(1956)}]{Austin1956}
Austin, J.L. 1956.
\newblock {Ifs and cans}.
\newblock In \emph{{Proceedings of the British Academy}}, vol.~42, 109--132.
  Cambridge University Press.
\newblock \doi{10.2307/2964530}.

\bibitem[{van~der Auwera(1997)}]{VanDerAuwera1997}
van~der Auwera, Johan. 1997.
\newblock {Pragmatics in the last quarter century: The case of conditional
  perfection}.
\newblock \emph{Journal of Pragmatics} 27(3). 261--274.
\newblock \doi{10.1016/s0378-2166(96)00058-6}.

\bibitem[{Bennett(2003)}]{bennett03}
Bennett, Jonathan~F. 2003.
\newblock \emph{{A Philosophical Guide to Conditionals}}.
\newblock Oxford University Press.

\bibitem[{Briggs(2012)}]{briggs12}
Briggs, Rachael. 2012.
\newblock Interventionist counterfactuals.
\newblock \emph{Philosophical studies} 160(1). 139--166.

\bibitem[{Brochhagen et~al.(2018)Brochhagen, Franke \& van
  Rooij}]{Brochhagen2018}
Brochhagen, Thomas, Michael Franke \& Robert van Rooij. 2018.
\newblock {Coevolution of Lexical Meaning and Pragmatic Use}.
\newblock \emph{Cognitive Science} 42(8). 2757--2789.
\newblock \doi{10.1111/cogs.12681}.

\bibitem[{Burnett(2019)}]{Burnett2019}
Burnett, Heather. 2019.
\newblock {Signalling games, sociolinguistic variation and the construction of
  style}.
\newblock \emph{Linguistics and Philosophy} 42(5). 419--450.
\newblock \doi{10.1007/s10988-018-9254-y}.

\bibitem[{Cheng(1997)}]{Cheng1997}
Cheng, Patricia~W. 1997.
\newblock {From covariation to causation: A causal power theory.}
\newblock \emph{Psychological Review} 104(2). 367--405.
\newblock \doi{10.1037//0033-295x.104.2.367}.

\bibitem[{Cruz et~al.(2016)Cruz, Over, Oaksford \& Baratgin}]{Cruz}
Cruz, Nicole, David Over, Mike Oaksford \& Jean Baratgin. 2016.
\newblock {Centering and the meaning of conditionals}.
\newblock In \emph{Cogsci 2016}, 1104--1110.

\bibitem[{Dasgupta et~al.(2017)Dasgupta, Schulz \&
  Gershman}]{dasgupta_where_2017}
Dasgupta, Ishita, Eric Schulz \& Samuel~J. Gershman. 2017.
\newblock Where do hypotheses come from?
\newblock \emph{Cognitive Psychology} 96. 1--25.
\newblock \doi{10.1016/j.cogpsych.2017.05.001}.
\newblock
  \urlprefix\url{https://www.sciencedirect.com/science/article/pii/S0010028516302766}.

\bibitem[{Degen et~al.(2020)Degen, Hawkins, Graf, Kreiss \&
  Goodman}]{Degen2020}
Degen, Judith, Robert~D. Hawkins, Caroline Graf, Elisa Kreiss \& Noah~D.
  Goodman. 2020.
\newblock {When Redundancy Is Useful: A Bayesian Approach to "Overinformative"
  Referring Expressions}.
\newblock \emph{Psychological Review} 127(4). 591--621.
\newblock \doi{10.1037/rev0000186}.

\bibitem[{D{\'i}ez(1993)}]{diezParameterAdjustmentBayes1993a}
D{\'i}ez, F.~J. 1993.
\newblock Parameter adjustment in {{Bayes}} networks. {{The}} generalized noisy
  {{OR}}\textendash gate.
\newblock In David Heckerman \& Abe Mamdani (eds.), \emph{Uncertainty in
  {{Artificial Intelligence}}}, 99--105. {Morgan Kaufmann}.
\newblock \doi{10.1016/B978-1-4832-1451-1.50016-0}.

\bibitem[{Douven(2008)}]{Douven2008}
Douven, Igor. 2008.
\newblock {The evidential support theory of conditionals}.
\newblock \emph{Synthese} 164(1). 19--44.
\newblock \doi{10.1007/s11229-007-9214-5}.

\bibitem[{Douven(2012)}]{Douven2012b}
Douven, Igor. 2012.
\newblock {Learning Conditional Information}.
\newblock \emph{Mind and Language} 27(3). 239--263.
\newblock \doi{10.1111/j.1468-0017.2012.01443.x}.

\bibitem[{Douven(2017)}]{Douven2017}
Douven, Igor. 2017.
\newblock {How to account for the oddness of missing-link conditionals}.
\newblock \emph{Synthese} 194. 1541--1554.
\newblock \doi{10.1007/s11229-015-0756-7}.

\bibitem[{Douven et~al.(2018)Douven, Elqayam, Singmann \& van
  Wijnbergen-Huitink}]{Douven2018}
Douven, Igor, Shira Elqayam, Henrik Singmann \& Janneke van Wijnbergen-Huitink.
  2018.
\newblock {Conditionals and inferential connections: A hypothetical inferential
  theory}.
\newblock \emph{Cognitive Psychology} 101. 50--81.
\newblock \doi{10.1016/j.cogpsych.2017.09.002}.

\bibitem[{Douven \& Romeijn(2011)}]{Douven2011a}
Douven, Igor \& Jan~Willem Romeijn. 2011.
\newblock {A new resolution of the Judy Benjamin problem}.
\newblock \emph{Mind} 120(479). 637--670.
\newblock \doi{10.1093/mind/fzr051}.

\bibitem[{Douven \& Verbrugge(2010)}]{Douven2010}
Douven, Igor \& Sara Verbrugge. 2010.
\newblock {The Adams family}.
\newblock \emph{Cognition} 117(3). 302--318.
\newblock \doi{10.1016/j.cognition.2010.08.015}.

\bibitem[{Douven \& Verbrugge(2012)}]{douven_indicatives_2012}
Douven, Igor \& Sara Verbrugge. 2012.
\newblock Indicatives, concessives, and evidential support.
\newblock \emph{Thinking \& Reasoning} 18(4). 480--499.
\newblock \doi{10.1080/13546783.2012.716009}.
\newblock
  \urlprefix\url{http://www.tandfonline.com/doi/abs/10.1080/13546783.2012.716009}.

\bibitem[{Earman(1992)}]{earman92}
Earman, John. 1992.
\newblock \emph{Bayes or bust? a critical examination of {B}ayesian
  confirmation theory}.
\newblock MIT Press.

\bibitem[{Edgington(1995)}]{edgington95}
Edgington, Dorothy. 1995.
\newblock On conditionals.
\newblock \emph{Mind} 104(414). 235--329.

\bibitem[{Eva et~al.(2019)Eva, Hartmann \& Rad}]{Eva2019}
Eva, Benjamin, Stephan Hartmann \& Soroush~Rafiee Rad. 2019.
\newblock {Learning from Conditionals}.
\newblock \emph{Mind} \doi{10.1093/mind/fzz025}.

\bibitem[{Evans \& Over(2004)}]{evansover04}
Evans, Johathan St. B.~T. \& David~E. Over. 2004.
\newblock \emph{If}.
\newblock Oxford University Press.

\bibitem[{Evans et~al.(2007)Evans, Handley, Neilens \&
  Over}]{evansThinkingConditionalsStudy2007}
Evans, Jonathan St~BT, Simon~J Handley, Helen Neilens \& David~E Over. 2007.
\newblock Thinking about conditionals: A study of individual differences.
\newblock \emph{Memory \& cognition} 35(7). 1772--1784.

\bibitem[{Fernbach \& Darlow(2010)}]{Fernbach2010}
Fernbach, Philip~M \& Adam Darlow. 2010.
\newblock {Neglect of alternative causes in predictive but not diagnostic
  reasoning}.
\newblock \emph{Psychological science} 21(3). 329--336.
\newblock \doi{10.1177/0956797610361430}.

\bibitem[{Fernbach et~al.(2011)Fernbach, Darlow \& Sloman}]{Fernbach2011}
Fernbach, Philip~M, Adam Darlow \& Steven~A Sloman. 2011.
\newblock {Asymmetries in predictive and diagnostic reasoning.}
\newblock \emph{Journal of experimental psychology: General} 140(2). 168--185.
\newblock \doi{10.1037/a0022100}.

\bibitem[{Fernbach \& Rehder(2013)}]{Fernbach2013}
Fernbach, Philip~M \& Bob Rehder. 2013.
\newblock {Cognitive shortcuts in causal inference}.
\newblock \emph{Argument and Computation} 4(1). 64--88.
\newblock \doi{10.1080/19462166.2012.682655}.

\bibitem[{de~Finetti(1936)}]{definetti36}
de~Finetti, Bruno. 1936.
\newblock La logique de la probabilit{\'e}.
\newblock In \emph{Actes du congr{\`e}s international de philosophie
  scientifique}, vol.~4, 1--9. Hermann Editeurs Paris.

\bibitem[{von Fintel(2001)}]{VonFintel2001}
von Fintel, Kai. 2001.
\newblock Conditional strengthening: A study in implicature.
\newblock Https://web.mit.edu/fintel/fintel-2001-condstrength.pdf.
\newblock \urlprefix\url{http://files/86/Von Fintel - 2001 - Conditional
  strengthening.pdf}.

\bibitem[{van Fraassen(1976)}]{vanfraassen76}
van Fraassen, Bas~C. 1976.
\newblock Probabilities of conditionals.
\newblock In W.L. Harper \& C.A. Hooker (eds.), \emph{Foundations of
  probability theory, statistical inference, and statistical theories of
  science}, vol.~1, 261--308. Reidel.

\bibitem[{Frank \& Goodman(2012)}]{Frank2012a}
Frank, Michael~C. \& Noah~D. Goodman. 2012.
\newblock {Predicting pragmatic reasoning in language games}.
\newblock \emph{Science} 336(6084). 998.
\newblock \doi{10.1126/science.1218633}.

\bibitem[{Franke(2011)}]{Franke2011:Quantity-Implic}
Franke, Michael. 2011.
\newblock Quantity implicatures, exhaustive interpretation, and rational
  conversation.
\newblock \emph{Semantics \& Pragmatics} 4(1). 1--82.

\bibitem[{Franke(2014)}]{Franke2012:Pragmatic-Reaso}
Franke, Michael. 2014.
\newblock Pragmatic reasoning about unawareness.
\newblock \emph{Erkenntnis} 79(4). 729--767.

\bibitem[{Franke \& Bergen(2020)}]{Franke2020}
Franke, Michael \& Leon Bergen. 2020.
\newblock {Theory-driven statistical modeling for semantics and pragmatics: A
  case study on grammatically generated implicature readings}.
\newblock \emph{Language} 96(2). e77--e96.
\newblock \doi{10.1353/lan.2020.0034}.

\bibitem[{Franke \& J{\"{a}}ger(2016)}]{Franke2016}
Franke, Michael \& Gerhard J{\"{a}}ger. 2016.
\newblock {Probabilistic pragmatics, or why Bayes' rule is probably important
  for pragmatics}.
\newblock \emph{Zeitschrift f{\"{u}}r Sprachwissenschaft} 35(1). 3--44.

\bibitem[{Franke \&
  {\SortNoop{Jager}}de~Jager(2011)}]{FrankeJagerde-Jager2010:Now-that-you-me}
Franke, Michael \& Tikitu {\SortNoop{Jager}}de~Jager. 2011.
\newblock Now that you mention it: {A}wareness dynamics in discourse and
  decisions.
\newblock In Anton Benz, Christian Ebert, Gerhard J{\"{a}}ger \& Robert
  {\SortNoop{Rooij}}van~Rooij (eds.), \emph{Language, games, and evolution}
  LNAI 6207, 60--91. Heidelberg: Springer.

\bibitem[{Gates et~al.(2018)Gates, Veuthey, Tessler, Smith, Gerstenberg, Bayet
  \& Tenenbaum}]{Gates2018}
Gates, Monica~A, Tess Veuthey, Michael~Henry Tessler, Kevin~A Smith, Tobias
  Gerstenberg, Laurie Bayet \& Josh Tenenbaum. 2018.
\newblock Tiptoeing around it: Inference from absence in potentially offensive
  speech.
\newblock In Chuck Kalish, Martina Rau, Jerry Zhu \& Timothy Rogers (eds.),
  \emph{Cogsci 2018}, 1693--1698.

\bibitem[{Geis \& Lycan(1993)}]{geisLycanNonconditional93}
Geis, Michael~L. \& William~G. Lycan. 1993.
\newblock Nonconditional conditionals.
\newblock \emph{Philosophical Topics} 21(2). 35--56.
\newblock \urlprefix\url{http://www.jstor.org/stable/43154153}.

\bibitem[{Geis \& Zwicky(1971)}]{Geis1971}
Geis, Michael~L \& Arnold~M Zwicky. 1971.
\newblock {On invited inferences}.
\newblock \emph{Linguistic inquiry} 2(4). 561--566.

\bibitem[{Geurts(2010)}]{Geurts2010:Quantity-Implic}
Geurts, Bart. 2010.
\newblock \emph{Quantity implicatures}.
\newblock Cambridge, UK: Cambridge University Press.

\bibitem[{Goodman \& Stuhlm{\"{u}}ller(2014)}]{Goodman2014a}
Goodman, ND \& A~Stuhlm{\"{u}}ller. 2014.
\newblock {The design and implementation of probabilistic programming
  languages}.
\newblock \urlprefix\url{http://dippl.org}.

\bibitem[{Goodman \& Frank(2016)}]{Goodman2016}
Goodman, Noah~D \& Michael~C Frank. 2016.
\newblock {Pragmatic language interpretation as probabilistic inference}.
\newblock \emph{Trends in cognitive sciences} 20(11). 818--829.

\bibitem[{Goodman \& Stuhlm{\"{u}}ller(2013)}]{Goodman2013}
Goodman, Noah~D \& Andreas Stuhlm{\"{u}}ller. 2013.
\newblock {Knowledge and implicature: Modeling language understanding as social
  cognition}.
\newblock \emph{Topics in cognitive science} 5(1). 173--184.

\bibitem[{Grice(1989)}]{griceIC}
Grice, H.~Paul. 1989.
\newblock Indicative conditionals.
\newblock In \emph{{Studies in the Way of Words}}, Harvard University Press.

\bibitem[{Grice(1975)}]{Grice1975}
Grice, HP. 1975.
\newblock {Logic and Conversation}.
\newblock In Maite Ezcurdia \& Robert~J. Stainton (eds.), \emph{The
  semantics-pragmatics boundary in philosophy}, 47. Broadview Press.

\bibitem[{Griffiths \& Tenenbaum(2005)}]{griffiths_structure_2005}
Griffiths, Thomas~L. \& Joshua~B. Tenenbaum. 2005.
\newblock Structure and strength in causal induction.
\newblock \emph{Cognitive psychology} 51(4). 334--384.
\newblock Publisher: Elsevier.

\bibitem[{Grove \& Halpern(1997)}]{Grove1997}
Grove, Adam~J. \& Joseph~Y. Halpern. 1997.
\newblock {Probability Update: Conditioning vs. Cross-Entropy}.
\newblock In \emph{{UAI} 1997}, 208--214.

\bibitem[{G{\"{u}}nther(2018)}]{Gunther2018}
G{\"{u}}nther, Mario. 2018.
\newblock {Learning conditional and causal information by Jeffrey imaging on
  Stalnaker conditionals}.
\newblock \emph{Journal of Philosophical Logic} 47(4). 851--876.
\newblock \doi{10.1007/s10992-017-9452-z}.

\bibitem[{Hadjichristidis et~al.(2001)Hadjichristidis, Stevenson, Over, Sloman,
  Evans \& Feeney}]{hadj}
Hadjichristidis, Constantinos, Rosemary~J Stevenson, David~E Over, Steven~A
  Sloman, JSBT Evans \& Aidan Feeney. 2001.
\newblock On the evaluation of ‘if p then q’ conditionals.
\newblock In \emph{Proceedings of the twenty-third annual conference of the
  {Cognitive Science Society}}, 409--414. Lawrence Erlbaum Associates.

\bibitem[{Hagmayer \& Waldmann(2007)}]{Hagmayer}
Hagmayer, York \& Michael~R Waldmann. 2007.
\newblock {Inferences about unobserved causes in human contingency learning}.
\newblock \emph{The Quarterly Journal of Experimental Psychology} 60(3).
  330--355.
\newblock \doi{10.1080/17470210601002470}.

\bibitem[{H{\'a}jek(1989)}]{hajek89}
H{\'a}jek, Alan. 1989.
\newblock Probabilities of conditionals -- revisited.
\newblock \emph{Journal of Philosophical Logic} 18(4). 423--428.

\bibitem[{Heifetz et~al.(2006)Heifetz, Meier \&
  Schipper}]{HeifetzMeier2006:Interactive-Una}
Heifetz, Aviad, Martin Meier \& Burkhard~C. Schipper. 2006.
\newblock Interactive unawareness.
\newblock \emph{Journal of Economic Theory} 130. 78--94.

\bibitem[{Herbstritt \& Franke(2019)}]{Herbstritt2019}
Herbstritt, Michele \& Michael Franke. 2019.
\newblock {Complex probability expressions {\&} higher-order uncertainty:
  Compositional semantics, probabilistic pragmatics {\&} experimental data}.
\newblock \emph{Cognition} 186. 50--71.
\newblock \doi{10.1016/j.cognition.2018.11.013}.

\bibitem[{Hiddleston(2005)}]{hiddleston05}
Hiddleston, Eric. 2005.
\newblock A causal theory of counterfactuals.
\newblock \emph{No{\^u}s} 39(4). 632--657.

\bibitem[{Horn(1984)}]{Horn1984}
Horn, Laurence~R. 1984.
\newblock {Toward a new taxonomy for pragmatic inference}.
\newblock \emph{Meaning, form, and use in context: linguistic applications} 11.
  11--42.

\bibitem[{Horn(2000)}]{Horn2000}
Horn, Laurence~R. 2000.
\newblock From if to iff: Conditional perfection as pragmatic strengthening.
\newblock \emph{Journal of Pragmatics} 32(3). 289--326.
\newblock \doi{https://doi.org/10.1016/S0378-2166(99)00053-3}.

\bibitem[{Hyttinen et~al.(2011)Hyttinen, Eberhardt \&
  Hoyer}]{hyttinenNoisyORModelsLatent2011}
Hyttinen, Antti, Frederick Eberhardt \& Patrik~O Hoyer. 2011.
\newblock Noisy-{{OR Models}} with {{Latent Confounding}}.
\newblock In \emph{Proceedings of the {{Twenty-Seventh Conference}} on
  {{Uncertainty}} in {{Artificial Intelligence}}}, 363--372.

\bibitem[{Icard \& Goodman(2015)}]{Icard2015}
Icard, Thomas~F. \& Noah~D Goodman. 2015.
\newblock {A Resource-Rational Approach to the Causal Frame Problem}.
\newblock In \emph{Cogsci 2015}, 962--967.
\newblock \doi{10.1111/1462-2920.14386}.

\bibitem[{Kao et~al.(2014{\natexlab{a}})Kao, Bergen \& Goodman}]{Kao2014}
Kao, Justine~T, Leon Bergen \& Noah~D Goodman. 2014{\natexlab{a}}.
\newblock {Formalizing the Pragmatics of Metaphor Understanding}.
\newblock \emph{CogSci 2014} 1(36). 719--724.
\newblock
  \urlprefix\url{https://escholarship.org/content/qt09h3p4cz/qt09h3p4cz.pdf}.

\bibitem[{Kao et~al.(2014{\natexlab{b}})Kao, Wu, Bergen \& Goodman}]{Kao2014a}
Kao, Justine~T, Jean~Y Wu, Leon Bergen \& Noah~D Goodman. 2014{\natexlab{b}}.
\newblock {Nonliteral understanding of number words}.
\newblock \emph{Proceedings of the National Academy of Sciences} 111(33).
  12002--12007.
\newblock \doi{10.1073/pnas.1407479111}.

\bibitem[{Katzir(2007)}]{Katzir2007}
Katzir, Roni. 2007.
\newblock {Structurally-defined alternatives}.
\newblock \emph{Linguistics and philosophy} 30. 669--690.
\newblock \doi{10.1007/s10988-008-9029-y}.

\bibitem[{Kaufmann(2004)}]{kaufmann04}
Kaufmann, Stefan. 2004.
\newblock Conditioning against the grain.
\newblock \emph{Journal of Philosophical Logic} 33(6). 583--606.

\bibitem[{Kaufmann(2013)}]{kaufmann13}
Kaufmann, Stefan. 2013.
\newblock Causal premise semantics.
\newblock \emph{Cognitive science} 37(6). 1136--1170.

\bibitem[{Khoo(2016)}]{khoo16b}
Khoo, Justin. 2016.
\newblock Probabilities of conditionals in context.
\newblock \emph{Linguistics and philosophy} 39(1). 1--43.

\bibitem[{Kratzer(1991)}]{kratzer91}
Kratzer, Angelika. 1991.
\newblock Modality.
\newblock In Arnim von Stechow \& Dieter Wunderlich (eds.), \emph{Semantik: Ein
  internationales {H}andbuch der zeitgen{\"o}ssischen {F}orschung}, 639--650.
  Walter de Gruyter.

\bibitem[{Krynski \& Tenenbaum(2007)}]{Krynski2007}
Krynski, Tevye~R \& Joshua~B Tenenbaum. 2007.
\newblock {The Role of Causality in Judgment Under Uncertainty}.
\newblock \emph{Journal of Experimental Psychology: General} 136(3). 430--450.
\newblock \doi{10.1037/0096-3445.136.3.430}.

\bibitem[{Krzy{\.{z}}anowska et~al.(2013)Krzy{\.{z}}anowska, Wenmackers \&
  Douven}]{Krzyzanowska2013}
Krzy{\.{z}}anowska, Karolina, S.~Wenmackers \& Igor Douven. 2013.
\newblock {Inferential Conditionals and Evidentiality}.
\newblock \emph{Journal of Logic, Language and Information} 22(3). 315--334.
\newblock \doi{10.1007/s10849-013-9178-4}.

\bibitem[{Krzy{\.{z}}anowska et~al.(2014)Krzy{\.{z}}anowska, Wenmackers \&
  Douven}]{Krzy2014}
Krzy{\.{z}}anowska, Karolina, Sylvia Wenmackers \& Igor Douven. 2014.
\newblock {Rethinking Gibbard's Riverboat Argument}.
\newblock \emph{Springer} 102(4). 771--792.
\newblock \doi{10.1007/s11225-013-9507-2}.

\bibitem[{Lassiter(2017)}]{Lassiter2017}
Lassiter, Daniel. 2017.
\newblock {Probabilistic language in indicative and counterfactual
  conditionals}.
\newblock \emph{Semantics and Linguistic Theory} 27(0). 525.
\newblock \doi{10.3765/salt.v27i0.4188}.

\bibitem[{Lassiter(2018)}]{lassiter17}
Lassiter, Daniel. 2018.
\newblock Talking about (quasi-)higher-order uncertainty.
\newblock In Cleo Condoravdi \& Tracy~Holloway King (eds.), \emph{Tokens of
  meaning: Papers in honor of {L}auri {K}arttunen}, CSLI Publications.

\bibitem[{Lassiter(2020)}]{lassiter19}
Lassiter, Daniel. 2020.
\newblock What we can learn from how trivalent conditionals avoid triviality.
\newblock \emph{Inquiry} 63(9-10). 1087--1114.

\bibitem[{Lassiter(\noop{3001}in press)}]{lassiter_decomposing_nodate}
Lassiter, Daniel. \noop{3001}in press.
\newblock Decomposing relevance in conditionals.
\newblock \emph{Mind and Language} .

\bibitem[{Lassiter \& Goodman(2017)}]{lassitergoodman17}
Lassiter, Daniel \& Noah~D.\ Goodman. 2017.
\newblock Adjectival vagueness in a {B}ayesian model of interpretation.
\newblock \emph{Synthese} 194(10). 3801--3836.

\bibitem[{Levinson(2000)}]{Levinson2000}
Levinson, SC. 2000.
\newblock \emph{{Presumptive meanings: The theory of generalized conversational
  implicature}}.
\newblock MIT press.

\bibitem[{Lewis(1976)}]{lewis76}
Lewis, David. 1976.
\newblock {Probabilities of conditionals and conditional probabilities}.
\newblock \emph{Philosophical Review} 85(3). 297--315.
\newblock \doi{10.2307/2184045}.

\bibitem[{Lucas \& Kemp(2015)}]{Lucas2015}
Lucas, Christopher~G. \& Charles Kemp. 2015.
\newblock {An improved probabilistic account of counterfactual reasoning.}
\newblock \emph{Psychological Review} 122(4). 700--734.
\newblock \doi{10.1037/a0039655}.

\bibitem[{Matsumoto(1995)}]{Matsumoto1995}
Matsumoto, Yo. 1995.
\newblock {The conversational condition on horn scales}.
\newblock \emph{Linguistics and Philosophy} 18(1). 21--60.
\newblock \doi{10.1007/BF00984960}.

\bibitem[{Milne(1997)}]{milne97}
Milne, Peter. 1997.
\newblock Bruno de {F}inetti and the logic of conditional events.
\newblock \emph{The British Journal for the Philosophy of Science} 48(2).
  195--232.

\bibitem[{Moldovan(2013)}]{moldovanDenyingAntecedentConditional2013}
Moldovan, Andrei. 2013.
\newblock Denying the antecedent and conditional perfection again .

\bibitem[{Moss(2015)}]{moss15}
Moss, Sarah. 2015.
\newblock On the semantics and pragmatics of epistemic vocabulary.
\newblock \emph{Semantics and Pragmatics} 8. 1--81.

\bibitem[{Newstead(1997{\natexlab{a}})}]{newstead_conditional_1997}
Newstead, Stephen~E. 1997{\natexlab{a}}.
\newblock Conditional {Reasoning} with {Realistic} {Material}.
\newblock \emph{Thinking \& Reasoning} 3(1). 49--76.
\newblock \doi{10.1080/135467897394428}.
\newblock
  \urlprefix\url{https://www.tandfonline.com/doi/full/10.1080/135467897394428}.

\bibitem[{Newstead(1997{\natexlab{b}})}]{newstead1997conditional}
Newstead, Stephen~E. 1997{\natexlab{b}}.
\newblock Conditional reasoning with realistic material.
\newblock \emph{Thinking \& Reasoning} 3(1). 49--76.

\bibitem[{Oberauer et~al.(2007)Oberauer, Weidenfeld \&
  Fischer}]{oberauer_what_2007}
Oberauer, Klaus, Andrea Weidenfeld \& Katrin Fischer. 2007.
\newblock What makes us believe a conditional? {The} roles of covariation and
  causality.
\newblock \emph{Thinking \& Reasoning} 13(4). 340--369.
\newblock Publisher: Taylor \& Francis.

\bibitem[{Oberauer \& Wilhelm(2003)}]{oberauer_meanings_2003}
Oberauer, Klaus \& Oliver Wilhelm. 2003.
\newblock The meaning(s) of conditionals: {Conditional} probabilities, mental
  models, and personal utilities.
\newblock \emph{Journal of Experimental Psychology: Learning, Memory, and
  Cognition} 29(4). 680--693.
\newblock \doi{10.1037/0278-7393.29.4.680}.
\newblock
  \urlprefix\url{http://doi.apa.org/getdoi.cfm?doi=10.1037/0278-7393.29.4.680}.

\bibitem[{Pearl(1988)}]{pearl88}
Pearl, Judea. 1988.
\newblock \emph{{Probabilistic Reasoning in Intelligent Systems: Networks of
  Plausible Inference}}.
\newblock Morgan Kaufmann.

\bibitem[{Pearl(2009)}]{Pearl2009}
Pearl, Judea. 2009.
\newblock \emph{{Causality}}.
\newblock Cambridge University Press.
\newblock \doi{10.1017/CBO9780511803161}.

\bibitem[{Pearl(2013)}]{Pearl2013}
Pearl, Judea. 2013.
\newblock {Structural counterfactuals: A brief introduction}.
\newblock \emph{Cognitive Science} 37(6). 977--985.
\newblock \doi{10.1111/cogs.12065}.

\bibitem[{Pearl(2014)}]{Pearl2014}
Pearl, Judea. 2014.
\newblock \emph{{Probabilistic Reasoning in Intelligent Systems: Networks of
  Plausible Inference by Judea Pearl}}.
\newblock Elsevier.

\bibitem[{Qing \& Franke(2015)}]{Qing2015}
Qing, Ciyang \& Michael Franke. 2015.
\newblock {Variations on a bayesian theme: Comparing bayesian models of
  referential reasoning}.
\newblock In \emph{Bayesian natural language semantics and pragmatics},
  201--220. Springer.

\bibitem[{Qing et~al.(2016)Qing, Goodman \& Lassiter}]{Qing2016}
Qing, Ciyang, Noah~D Goodman \& Daniel Lassiter. 2016.
\newblock {A rational speech-act model of projective content}.
\newblock \emph{CogSci} 1110--1115.

\bibitem[{Quine(1965)}]{quine65}
Quine, Willard Van~Orman. 1965.
\newblock \emph{Elementary logic: Revised edition}.
\newblock Harper.

\bibitem[{Rips(2010)}]{Rips2010}
Rips, Lance~J. 2010.
\newblock {Two causal theories of counterfactual conditionals}.
\newblock \emph{Cognitive Science} 34(2). 175--221.
\newblock \doi{10.1111/j.1551-6709.2009.01080.x}.

\bibitem[{Roberts(2012)}]{Roberts2012:Information-Str}
Roberts, Craige. 2012.
\newblock Information structure in discourse: {T}owards an integrated theory of
  pragmatics.
\newblock \emph{Semantics \& Pragmatics} 5(6). 1--69.

\bibitem[{van Rooij \& Schulz(2019)}]{VanRooij2019}
van Rooij, Robert \& Katrin Schulz. 2019.
\newblock {Conditionals, Causality and Conditional Probability}.
\newblock \emph{Journal of Logic, Language and Information} 28(1). 55--71.
\newblock \doi{10.1007/s10849-018-9275-5}.

\bibitem[{van Rooij \& Schulz(2020)}]{VanRooij2020}
van Rooij, Robert \& Katrin Schulz. 2020.
\newblock {Conditionals As Representative Inferences}.
\newblock \emph{Axiomathes} 0123456789.
\newblock \doi{10.1007/s10516-020-09477-9}.

\bibitem[{van Rooij \& Schulz(2021)}]{Rooij2021}
van Rooij, Robert \& Katrin Schulz. 2021.
\newblock Why those biscuits are relevant and on the sideboard.
\newblock \emph{Theoria} theo.12309.
\newblock \doi{10.1111/theo.12309}.
\newblock
  \urlprefix\url{https://onlinelibrary.wiley.com/doi/10.1111/theo.12309}.

\bibitem[{Rothschild(2015)}]{Rothschild2015}
Rothschild, Daniel. 2015.
\newblock {Conditionals and propositions in semantics}.
\newblock \emph{Journal of Philosophical Logic} 44(6). 781--791.
\newblock \doi{10.1007/s10992-015-9359-5}.

\bibitem[{Santorio(2016)}]{santorio16}
Santorio, Paolo. 2016.
\newblock Interventions in premise semantics.
\newblock \emph{Philosophers' Imprint} .

\bibitem[{Schuster \& Degen(2019)}]{Schuster}
Schuster, Sebastian \& Judith Degen. 2019.
\newblock {Speaker-specific adaptation to variable use of uncertainty
  expressions}.
\newblock In \emph{Cogsci 2019}, 2769--2775.
\newblock \urlprefix\url{https://osf.io/qnspg}.

\bibitem[{Scontras et~al.(2017)Scontras, Tessler \& Franke}]{Scontras2017}
Scontras, G, MH~Tessler \& Michael Franke. 2017.
\newblock {Probabilistic language understanding: An introduction to the
  Rational Speech Act framework}.
\newblock Retrieved 2021-3-24 from https://www.problang.org.
\newblock \urlprefix\url{https://www.problang.org}.

\bibitem[{Singmann et~al.(2014)Singmann, Klauer \& Over}]{singmann_new_2014}
Singmann, Henrik, Karl~Christoph Klauer \& David Over. 2014.
\newblock New normative standards of conditional reasoning and the dual-source
  model.
\newblock \emph{Frontiers in Psychology} 5.
\newblock
  \urlprefix\url{https://www.frontiersin.org/article/10.3389/fpsyg.2014.00316}.

\bibitem[{Skovgaard-Olsen(2016)}]{Skovgaard-Olsen2016a}
Skovgaard-Olsen, Niels. 2016.
\newblock {Motivating the Relevance Approach to Conditionals}.
\newblock \emph{Mind and Language} 31(5). 555--579.
\newblock \doi{10.1111/mila.12120}.

\bibitem[{Skovgaard-Olsen et~al.(2016)Skovgaard-Olsen, Singmann \&
  Klauer}]{Skovgaard-Olsen2016}
Skovgaard-Olsen, Niels, Henrik Singmann \& Karl~Christoph Klauer. 2016.
\newblock {The relevance effect and conditionals}.
\newblock \emph{Cognition} 150. 26--36.

\bibitem[{Skovgaard-Olsen et~al.(2017)Skovgaard-Olsen, Singmann \&
  Klauer}]{Skovgaard-Olsen2017}
Skovgaard-Olsen, Niels, Henrik Singmann \& Karl~Christoph Klauer. 2017.
\newblock {Relevance and Reason Relations}.
\newblock \emph{Cognitive Science} 41. 1202--1215.
\newblock \doi{10.1111/cogs.12462}.

\bibitem[{Stalnaker(1968)}]{stalnaker68}
Stalnaker, Robert. 1968.
\newblock A theory of conditionals.
\newblock In Nicholas Rescher (ed.), \emph{Studies in logical theory}, 98--112.
  Blackwell.

\bibitem[{Stalnaker \& Jeffrey(1994)}]{stalnakerjeffrey94}
Stalnaker, Robert \& Richard Jeffrey. 1994.
\newblock Conditionals as random variables.
\newblock In \emph{Probability and conditionals: Belief revision and rational
  decision}, 31--46. Cambridge University Press.

\bibitem[{Swanson(2015)}]{swanson15}
Swanson, Eric. 2015.
\newblock The application of constraint semantics to the language of subjective
  uncertainty.
\newblock \emph{Journal of Philosophical Logic} 1--26.

\bibitem[{Tenenbaum \& Griffiths(2003)}]{tenenbaum_theory-based_2003}
Tenenbaum, Joshua~B. \& Thomas~L. Griffiths. 2003.
\newblock Theory-based causal inference.
\newblock \emph{Advances in neural information processing systems} 43--50.
\newblock Publisher: MIT; 1998.

\bibitem[{Tenenbaum et~al.(2011)Tenenbaum, Kemp, Griffiths \&
  Goodman}]{tenenbaumetal11}
Tenenbaum, Joshua~B., Charles Kemp, Tom~L. Griffiths \& Noah~D. Goodman. 2011.
\newblock How to grow a mind: Statistics, structure, and abstraction.
\newblock \emph{Science} 331(6022). 1279--1285.

\bibitem[{{\"U}lk{\"u}men et~al.(2016){\"U}lk{\"u}men, Fox \& Malle}]{fox15}
{\"U}lk{\"u}men, G{\"u}lden, Craig~R Fox \& Bertram~F Malle. 2016.
\newblock Two dimensions of subjective uncertainty: Clues from natural
  language.

\bibitem[{Vandenburgh(2020)}]{Vandenburgh2020}
Vandenburgh, Jonathan. 2020.
\newblock Conditional learning through causal models.
\newblock \emph{Synthese} \doi{10.1007/s11229-020-02891-x}.

\bibitem[{Yalcin(2012)}]{yalcin12a}
Yalcin, Seth. 2012.
\newblock Bayesian expressivism.
\newblock \emph{Proceedings of the {A}ristotelian {S}ociety} 112. 123--160.

\bibitem[{Yoon et~al.(2016)Yoon, Tessler, Goodman \& Frank}]{Yoon2016}
Yoon, Erica~J, Michael~Henry Tessler, Noah~D Goodman \& Michael~C Frank. 2016.
\newblock {Talking with tact: Polite language as a balance between kindness and
  informativity}.
\newblock In \emph{Cogsci 2016}, 2771--2776. Cognitive Science Society.

\end{thebibliography}
